\documentclass[11pt, a4paper]{deepmind}

\usepackage{times}

\usepackage{xspace}
\usepackage{siunitx}
\usepackage{amsmath}
\usepackage{amssymb}
\usepackage{enumitem}
\usepackage[T1]{fontenc}
\usepackage{multicol}
\usepackage{caption}
\usepackage{subcaption}
\usepackage{sidecap}
\usepackage{multirow}
\usepackage{placeins}
\usepackage{titlesec}
\usepackage[capitalise,noabbrev]{cleveref}
\usepackage{graphicx}
\usepackage{url}
\usepackage{lineno}
\usepackage{titletoc}
\usepackage{bibunits}
\usepackage{bm}

\usepackage{placeins}

\makeatletter
\usepackage{etoolbox}
\patchcmd\H@refstepcounter{\protected@edef}{\protected@xdef}{}{}
\makeatother

\ifdefined\unit\else
  \ifdefined\NewCommandCopy
    \NewCommandCopy\unit\si
  \else
    \NewDocumentCommand\unit{O{}m}{\si[#1]{#2}}
  \fi
\fi

\crefformat{footnote}{#2\footnotemark[#1]#3}

\crefname{appsec}{Appendix Section}{Appendix Sections}
\crefname{appfig}{Appendix Figure}{Appendix Figures}
\crefname{apptab}{Appendix Table}{Appendix Tables}
\crefname{appeq}{Appendix Equation}{Appendix Equations}

\newcommand{\datetime}[2]{\texttt{#1\_#2\xspace}}
\newcommand{\quarterdegree}{\ang{0.25}\xspace}
\newcommand{\pointonedegree}{\ang{0.1}\xspace}

\newcommand{\hresfczero}{HRES-fc0\xspace}

\newcommand{\ourmodel}{GraphCast\xspace}

\newcommand{\pangumodel}{Pangu-Weather\xspace}

\newcommand{\varweight}{w}
\newcommand{\latitudeweight}{a}
\newcommand{\stdweight}{s}
\newcommand{\sll}{i}

\newcommand{\Z}{Z}
\newcommand{\x}{x}
\newcommand{\xvec}{\mathbf{\x}}
\newcommand{\xpred}{\hat{\x}}

\newcommand{\Xpred}{\hat{\X}}
\newcommand{\X}{X}
\newcommand{\y}{y}
\newcommand{\ypred}{\hat{\y}}
\newcommand{\Y}{Y}
\newcommand{\Ypred}{\hat{\Y}}

\newcommand{\G}{G}
\newcommand{\Gquarterdegree}{\G_{\quarterdegree}}
\newcommand{\Gsuper}{\text{\G}}

\newcommand{\dd}{d}
\newcommand{\dinit}{\dd_0}
\newcommand{\dsize}{\Delta \dd}
\newcommand{\ttt}{t}
\newcommand{\T}{T}
\newcommand{\lt}{\tau}

\newcommand{\rr}{r}
\newcommand{\R}{R}

\newcommand{\climatology}{C}

\newcommand{\fwdtrue}{\Phi}
\newcommand{\fwdmodel}{\phi}

\newcommand{\forcing}{\mathbf{f}}
\newcommand{\constant}{\mathbf{c}}

\newcommand{\edge}{\mathbf{e}}

\makeatletter
\def\varlevel#1{\def\tempa{\lowercase{#1}}\futurelet\next\varlevel@i}%
\def\varlevel@i{\ifx\next\bgroup\expandafter\varlevel@ii\else\expandafter\varlevel@end\fi}%
\def\varlevel@ii#1{\textsc{\tempa#1}}%
\def\varlevel@end{\textsc{\tempa}}%
\makeatother

\makeatletter
\def\enotation#1{\def\tempa{#1}\futurelet\next\enotation@i}%
\def\enotation@i{\ifx\next\bgroup\expandafter\enotation@ii\else\expandafter\enotation@end\fi}%
\def\enotation@ii#1{$\tempa\mathrm{e}{#1}$}%
\def\enotation@end{$\tempa$}%
\makeatother

\newcommand{\numtargets}{1380}
\newcommand{\numtargetswithoutfifty}{1280}
\newcommand{\numtargetswithoutfiftyandhundred}{1180}
\newcommand{\numtargetspangu}{252}

\newcommand{\percentbetterthanhres}{90.3}
\newcommand{\percentsignificantlybetterthanhrestwosigfig}{90}
\newcommand{\percentsignificantlybetterthanhres}{89.9}

\newcommand{\percentbetterthanhresoptimalblurring}{88.0}

\newcommand{\percentsignificantlybetterthanhresx}{96.9} 

\newcommand{\percentsignificantlybetterthanhresxx}{99.7}

\newcommand{\percentbetterthanpangu}{99.2}

\newcommand{\vll}[2]{\textsc{#1}#2}

\title{GraphCast: Learning skillful medium-range global weather forecasting}

\author[*,1]{Remi~Lam}
\author[*,1]{Alvaro~Sanchez-Gonzalez}
\author[*,1]{Matthew~Willson}
\author[*,1]{Peter~Wirnsberger}
\author[*,1]{Meire~Fortunato}
\author[*,1]{Ferran~Alet} 
\author[*,1]{Suman~Ravuri}
\author[1]{Timo~Ewalds}
\author[1]{Zach~Eaton-Rosen}
\author[1]{Weihua~Hu}
\author[2]{Alexander~Merose}
\author[2]{Stephan~Hoyer}
\author[1]{George~Holland}
\author[1]{Oriol~Vinyals}
\author[1]{Jacklynn~Stott} 
\author[1]{Alexander~Pritzel} 
\author[1]{Shakir~Mohamed}
\author[1]{Peter~Battaglia}

\affil[*]{equal contribution}
\affil[1]{Google DeepMind}
\affil[2]{Google Research}

\correspondingauthor{remilam@google.com, alvarosg@google.com, matthjw@google.com, peterbattaglia@google.com}


\date{}

\DeclareUnicodeCharacter{2212}{-}

\begin{abstract}
Global medium-range weather forecasting is critical to decision-making across many social and economic domains. 
Traditional numerical weather prediction uses increased compute resources to improve forecast accuracy, but cannot directly use historical weather data to improve the underlying model. We introduce a machine learning-based method called ``\ourmodel'', which can be trained directly from reanalysis data. It predicts hundreds of weather variables, over 10 days at \quarterdegree resolution globally, in under one minute.
We show that \ourmodel significantly outperforms the most accurate operational deterministic systems on \percentsignificantlybetterthanhrestwosigfig\% of \numtargets{} verification targets, and its forecasts support better severe event prediction, including tropical cyclones, atmospheric rivers, and extreme temperatures.
\ourmodel{} is a key advance in accurate and efficient weather forecasting, and helps realize the promise of machine learning for modeling complex dynamical systems.
\end{abstract}

\begin{document}

\begin{bibunit}

\maketitle

\section*{Introduction}

It is 05:45 UTC in mid-October, 2022, in Bologna, Italy, and the European Centre for Medium-Range Weather Forecasts (ECMWF)'s new High-Performance Computing Facility has just started operation. For the past several hours the Integrated Forecasting System (IFS) has been running sophisticated calculations to forecast Earth's weather over the next days and weeks, and its first predictions have just begun to be disseminated to users. This process repeats every six hours, every day, to supply the world with the most accurate weather forecasts available.

The IFS, and modern weather forecasting more generally, are triumphs of science and engineering. The dynamics of weather systems are among the most complex physical phenomena on Earth, and each day, countless decisions made by individuals, industries, and policymakers depend on accurate weather forecasts, from deciding whether to wear a jacket or to flee a dangerous storm.
The dominant approach for weather forecasting today is ``numerical weather prediction'' (NWP), which involves solving the governing equations of weather using supercomputers. 
The success of NWP lies in the rigorous and ongoing research practices that provide increasingly detailed descriptions of weather phenomena, and how well NWP scales to greater accuracy with greater computational resources~\cite{benjamin2019100,bauerQuietRevolution}.
As a result, the accuracy of weather forecasts have increased year after year, to the point where the surface temperature, or the path of a hurricane, can be predicted many days ahead---a possibility that was unthinkable even a few decades ago. 

But while traditional NWP scales well with compute, its accuracy does not improve with increasing amounts of historical data. There are vast archives of weather and climatological data, e.g. ECMWF's MARS \cite{mars_userdoc}, but until recently there have been few practical means for using such data to directly improve the quality of forecast models. Rather, NWP methods are improved by highly trained experts innovating better models, algorithms, and approximations, which can be a time-consuming and costly process.

Machine learning-based weather prediction (MLWP) offers an alternative to traditional NWP, where forecast models are trained directly from historical data. This has potential to improve forecast accuracy by capturing patterns and scales in the data which are not easily represented in explicit equations. MLWP also offers opportunities for greater efficiency by exploiting modern deep learning hardware, rather than supercomputers, and striking more favorable speed-accuracy trade-offs. 
Recently MLWP has helped improve on NWP-based forecasting in regimes where traditional NWP is relatively weak, for example sub-seasonal heat wave prediction~\cite{lopez2022global} and precipitation nowcasting from radar images~\cite{shi2017deep,sonderby2020metnet,ravuri2021skilful,espeholt2022deep}, where accurate equations and robust numerical methods are not as available.

In medium-range weather forecasting, i.e., predicting atmospheric variables up to 10 days ahead, NWP-based systems like the IFS are still most accurate. The top deterministic operational system in the world is ECMWF's High RESolution forecast (HRES), a component of IFS which produces global 10-day forecasts at \pointonedegree latitude/longitude resolution, in around an hour~\cite{rasp2020weatherbench}. However, over the past several years, MLWP methods for medium-range forecasting have been steadily advancing, facilitated by benchmarks such as WeatherBench~\cite{rasp2020weatherbench}. Deep learning architectures based on convolutional neural networks~\cite{weyn2019can,weyn2020improving,rasp2021data} and Transformers~\cite{nguyen2023climax} have shown promising results at latitude/longitude resolutions coarser than \ang{1.0}, and recent works---which use graph neural networks (GNN)~\cite{keisler2022forecasting}, Fourier neural operators~\cite{pathak2022fourcastnet,kurth2022fourcastnet}, and Transformers~\cite{bi2022pangu}---have reported performance that begins to approach IFS's at \ang{1.0} and \quarterdegree for a handful of variables, and lead times up to seven days.

\begin{figure}
  \centering
  \includegraphics[width=0.95\textwidth]{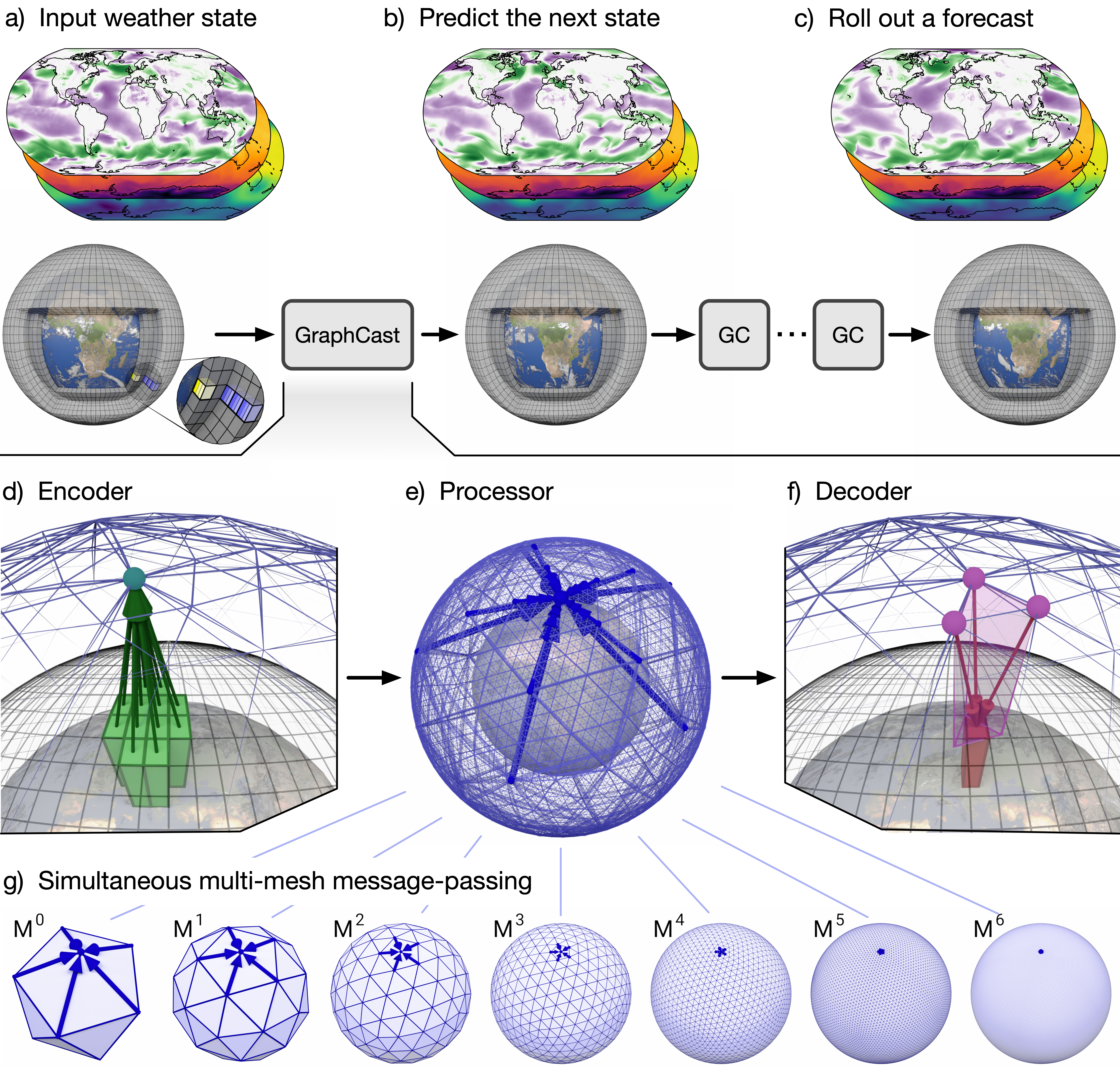}
  \caption{\small\textbf{Model schematic.}
  (a) The input weather state(s) are defined on a \quarterdegree{} latitude-longitude grid comprising a total of $721\times 1440 = 1,038,240$ points. Yellow layers in the closeup pop-out window represent the 5 surface variables, and blue layers represent the 6 atmospheric variables that are repeated at 37 pressure levels ($5 + 6\times 37 = 227$ variables per point in total), resulting in a state representation of $235,680,480$ values.
  (b) \ourmodel{} predicts the next state of the weather on the grid.
  (c) A forecast is made by iteratively applying \ourmodel{} to each previous predicted state, to produce a sequence of states which represent the weather at successive lead times.
  (d) The Encoder component of the \ourmodel{} architecture maps local regions of the input (green boxes) into nodes of the multi-mesh graph representation (green, upward arrows which terminate in the green-blue node).
  (e) The Processor component updates each multi-mesh node using learned message-passing (heavy blue arrows that terminate at a node).
  (f) The Decoder component maps the processed multi-mesh features (purple nodes) back onto the grid representation (red, downward arrows which terminate at a red box).
  (g) The multi-mesh is derived from icosahedral meshes of increasing resolution, from the base mesh ($M^0$, $12$ nodes) to the finest resolution ($M^6$, $40,962$ nodes), which has uniform resolution across the globe.
  It contains the set of nodes from $M^6$, and all the edges from $M^0$ to $M^6$.
  The learned message-passing over the different meshes' edges happens simultaneously, so that each node is updated by all of its incoming edges.}
  \label{fig:schematic}
\end{figure}

\begin{center}
\begin{table}[htbp]
\small
\centering
\begin{tabular}{| l | l | l |} 
\hline
Surface variables (5) & Atmospheric variables (6) & Pressure levels (37) \\
 \hline
\textbf{2-meter temperature} (\varlevel{2t}) & \textbf{Temperature} (\varlevel{t}) & 1, 2, 3, 5, 7, 10, 20, 30, \textbf{50}, 70, \\
\textbf{10 metre u wind component} (\varlevel{10u}) & \textbf{U component of wind}  (\varlevel{u}) & \textbf{100}, 125, \textbf{150}, 175, \textbf{200}, 225, \\
\textbf{10 metre v wind component} (\varlevel{10v}) & \textbf{V component of wind} (\varlevel{v}) & \textbf{250}, \textbf{300}, 350, \textbf{400}, 450, \textbf{500}, \\
\textbf{Mean sea-level pressure} (\varlevel{msl}) & \textbf{Geopotential} (\varlevel{z}) & 550, \textbf{600}, 650, \textbf{700}, 750, 775, \\
Total precipitation (\varlevel{tp}) & \textbf{Specific humidity} (\varlevel{q}) & 800, 825, \textbf{850}, 875, 900, \textbf{925}, \\
\ & Vertical wind speed (\varlevel{w}) & 950, 975, \textbf{1000} \\
\hline
\end{tabular}
\caption{\small\textbf{Weather variables and levels modeled by \ourmodel{}.} The numbers in parentheses in the column headings are the number of entries in the column. Boldfaced variables and levels indicates those which were included in the scorecard evaluation.}
\label{tab:variables}
\end{table}
\end{center}

\section*{\ourmodel{}}

Here we introduce a new MLWP approach for global medium-range weather forecasting called ``\ourmodel{}'', which produces an accurate 10-day forecast in under a minute on a single Google Cloud TPU v4 device, and supports applications including predicting tropical cyclone tracks, atmospheric rivers, and extreme temperatures. 

\ourmodel takes as input the two most recent states of Earth's weather---the current time and six hours earlier---and predicts the next state of the weather six hours ahead. A single weather state is represented by a \quarterdegree latitude/longitude grid ($721 \times 1440$), which corresponds to roughly $28\times 28$ kilometer resolution at the equator (\cref{fig:schematic}a), where each grid point represents a set of surface and atmospheric variables (listed in~\cref{tab:variables}). Like traditional NWP systems, \ourmodel is autoregressive: it can be ``rolled out'' by feeding its own predictions back in as input, to generate an arbitrarily long trajectory of weather states (\cref{fig:schematic}b--c).

\ourmodel is implemented as a neural network architecture, based on GNNs in an ``encode-process-decode'' configuration \cite{battaglia2018relational}, with a total of 36.7 million parameters. Previous GNN-based learned simulators~\cite{sanchez2020learning,pfaff2021learning} have been very effective at learning the complex dynamics of fluid and other systems modeled by partial differential equations, which supports their suitability for modeling weather dynamics. 

The encoder (\cref{fig:schematic}d) uses a single GNN layer to map variables (normalized to zero-mean unit-variance) represented as node attributes on the input grid to learned node attributes on an internal ``multi-mesh'' representation.

The multi-mesh (\cref{fig:schematic}g) is a graph which is spatially homogeneous, with high spatial resolution over the globe. It is defined by refining a regular icosahedron (12 nodes, 20 faces, 30 edges) iteratively six times, where each refinement divides each triangle into four smaller ones (leading to four times more faces and edges), and reprojecting the nodes onto the sphere. The multi-mesh contains the 40,962 nodes from the highest resolution mesh, and the union of all the edges created in the intermediate graphs, forming a flat hierarchy of edges with varying lengths.

The processor (\cref{fig:schematic}e) uses 16 unshared GNN layers to perform learned message-passing on the multi-mesh, enabling efficient local and long-range information propagation with few message-passing steps. 

The decoder (\cref{fig:schematic}f) maps the final processor layer's learned features from the multi-mesh representation back to the latitude-longitude grid. It uses a single GNN layer, and predicts the output as a residual update to the most recent input state (with output normalization to achieve unit-variance on the target residual). See Supplements~\cref{sec:app:ourmodel} for further architectural details.

During model development, we used 39 years (1979--2017) of historical data from ECMWF's ERA5~\cite{hersbach2020era5} reanalysis archive. As a training objective, we averaged the mean squared error (MSE) weighted by vertical level. Error was computed between \ourmodel's predicted state and the corresponding ERA5 state over $N$ autoregressive steps. The value of $N$ was increased incrementally from 1 to 12 (i.e., six hours to three days) over the course of training. \ourmodel{} was trained to minimize the training objective using gradient descent and backpropagation. Training \ourmodel{} took roughly four weeks on 32 Cloud TPU v4 devices using batch parallelism. See Supplements~\cref{sec:app:trainingdetails} for further training details.

Consistent with real deployment scenarios, where future information is not available for model development, we evaluated \ourmodel on the held out data from the years 2018 onward (see Supplements~\cref{sec:app:trainvaltest}).

\section*{Verification methods}
We verify \ourmodel's forecast skill comprehensively by comparing its accuracy to HRES's on a large number of variables, levels, and lead times. We quantify the respective skills of \ourmodel, HRES, and ML baselines with two skill metrics: the root mean square error~(RMSE) and the anomaly correlation coefficient~(ACC).

Of the 227 variable and level combinations predicted by \ourmodel{} at each grid point, we evaluated its skill versus HRES on 69 of them, corresponding to the 13 levels of WeatherBench\cite{rasp2020weatherbench} and variables from the ECMWF Scorecard~\cite{haiden2018evaluation}; see boldface variables and levels in \cref{tab:variables} and Supplements~\cref{sec:app:hres} for which HRES cycle was operational during the evaluation period. Note, we exclude total precipitation from the evaluation because ERA5 precipitation data has known biases \cite{lavers2022evaluation}. In addition to the aggregate performance reported in the main text, Supplements~\cref{sec:app:additional_results} provides further detailed evaluations, including other variables, regional performance, latitude and pressure level effects, spectral properties, blurring, comparisons to other ML-based forecasts, and effects of model design choices. 

In making these comparisons, two key choices underlie how skill is established: (1) the selection of the ground truth for comparison, and (2) a careful accounting of the data assimilation windows used to ground data with observations. 
We use ERA5 as the ground truth for evaluating \ourmodel, since it was trained to take ERA5 data as input and predict ERA5 data as outputs.
However, evaluating HRES forecasts against ERA5 would result in non-zero error on the initial forecast step. Instead, we constructed an ``HRES forecast at step 0'' (HRES-fc0) dataset to use as ground truth for HRES. HRES-fc0 contains the inputs to HRES forecasts at future initializations (see Supplements~\cref{sec:app:hres}), ensuring that each data point is grounded by recent observations and that the zeroth step of HRES forecasts will have zero error.

Fair comparisons between methods require that no method should have privileged information not available to the other. Because of the nature of weather forecast data, this requires careful control of the differences between the ERA5 and HRES data assimilation windows.
Each day, HRES assimilates observations using four +/-3h windows centered on 00z, 06z, 12z and 18z (where 18z means 18:00 UTC), while ERA5 uses two +9h/-3h windows centered on 00z and 12z, or equivalently two +3h/-9h windows centered on 06z and 18z.
We chose to evaluate \ourmodel's forecasts from the 06z and 18z initializations, ensuring its inputs carry information from +3h of future observations, matching HRES's inputs. We did not evaluate \ourmodel from 00z and 12z initializations, avoiding a mismatch between a +9h lookahead in ERA5 inputs versus +3h lookahead for HRES inputs.
We applied the same logic when choosing target lead times and evaluate targets only every 12h to ensure that the ground truth ERA5 and HRES have the same +3h lookahead (see Supplements~\cref{sec:app:comparingGraphcastHRES}).

HRES's forecasts initialized at 06z and 18z are only run for a horizon of 3.75 days (HRES's 00z and 12z initializations are run for 10 days). Therefore, our figures will indicate a transition with dashed line, where the 3.5 days before the line are comparisons with HRES initialized at 06z and 18z, and after the line are comparisons with initializations at 00z and 12z.
Supplements~\cref{sec:app:evaluationdetails} contains further verification details.

\begin{figure}
  \centering
  \includegraphics[width=\textwidth]{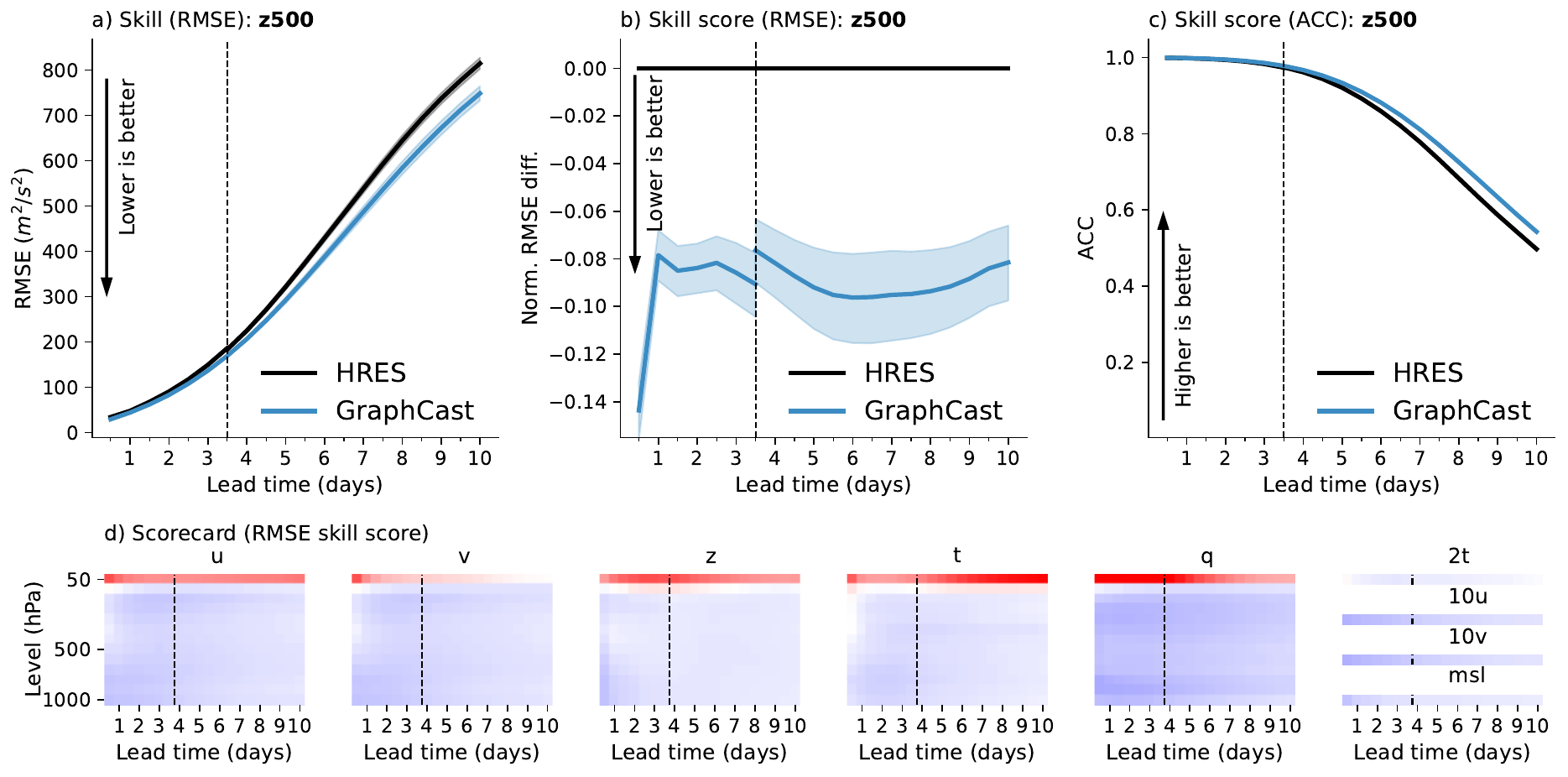}
  \caption{\small\textbf{Skill and skill scores for \ourmodel{} and HRES in 2018}. (a) RMSE skill (y-axis) for \ourmodel{} (blue lines) and HRES (black lines), on \vll{z}{500}, as a function of lead time (x-axis). Error bars represent 95\% confidence intervals. The vertical dashed line represents 3.5 days, which is the last 12 hour increment of the HRES 06z/18z forecasts. The black line represents HRES, where lead times earlier and later than 3.5 days are from the 06z/18z and 00z/12z initializations, respectively. (b) RMSE skill score (y-axis) for \ourmodel{} versus HRES, on \vll{z}{500}, as a function of lead time (x-axis). Error bars represent 95\% confidence intervals for the skill score. We observe a discontinuity in \ourmodel's curve because skill scores up to 3.5 days are computed between \ourmodel{} (initialized at 06z/18z) and HRES's 06z/18z initialization, while after 3.5 days skill scores are computed with respect to HRES's 00z/12z initializations. (c) ACC skill (y-axis) for \ourmodel{} (blue lines) and HRES (black lines), on \vll{z}{500}, as a function of lead time (x-axis). (d) Scorecard of RMSE skill scores for \ourmodel{}, with respect to HRES. Each subplot corresponds to one variable: \vll{u}{}, \vll{v}{}, \vll{z}{}, \vll{t}{}, \vll{q}{}, \vll{2t}{},  \vll{10u}{}, \vll{10v}{}, \vll{msl}{}, respectively. The rows of each heatmap correspond to the 13 pressure levels (for the atmospheric variables),  from 50 \unit{hPa} at the top to 1000 \unit{hPa} at the bottom. The columns of each heatmap correspond to the 20 lead times at 12 hour intervals, from 12 hours on the left to 10 days on the right. Each cell's color represents the skill score, as shown in (b), where blue represents negative values (\ourmodel{} has better skill) and red represents positive values (HRES has better skill).}
  \label{fig:resultshres}
\end{figure}

\section*{Forecast verification results}

We find that \ourmodel{} has greater weather forecasting skill than HRES when evaluated on 10-day forecasts at a horizontal resolution of \quarterdegree{} for latitude/longitude and at 13 vertical levels.

\cref{fig:resultshres}a--c show how \ourmodel (blue lines) outperforms HRES (black lines) on the \varlevel{z}{500} (geopotential at 500 \unit{hPa}) ``headline'' field in terms of RMSE skill, RMSE skill score (i.e., the normalized RMSE difference between model $A$ and baseline $B$ defined as $(\mathrm{RMSE}_A - \mathrm{RMSE}_B) / \mathrm{RMSE}_B$), and ACC skill. Using \varlevel{z}{500}, which encodes the synoptic-scale pressure distribution, is common in the literature, as it has strong meteorological importance~\cite{rasp2020weatherbench}.
The plots show \ourmodel{} has better skill scores across all lead times, with a skill score improvement around $7\%$--$14\%$. Plots for additional headline variables are in Supplements~\cref{sec:app:additional_variables}.

\cref{fig:resultshres}d summarizes the RMSE skill scores for all \numtargets{} evaluated variables and pressure levels, across the 10 day forecasts, in a format analogous to the ECMWF Scorecard. The cell colors are proportional to the skill score, where blue indicates \ourmodel{} had better skill and red indicates HRES had higher skill. \ourmodel{} outperformed HRES on $\percentbetterthanhres\%$ of the \numtargets{} targets, and significantly ($p \le 0.05$, nominal sample size $n \in \{729, 730\}$) outperformed HRES on $\percentsignificantlybetterthanhres\%$ of targets.  See Supplements~\cref{sec:app:statistical_testing} for methodology and Supplements~\cref{tab:app:significance_test_details} for $p$-values, test statistics and effective sample sizes.

The regions of the atmosphere in which HRES had better performance than \ourmodel{} (top rows in red in the scorecards), were disproportionately localized in the stratosphere, and had the lowest training loss weight (see Supplements~\cref{app:sec:resultslatlevel}).
When excluding the 50 \unit{hPa} level, \ourmodel{} significantly outperforms HRES on $\percentsignificantlybetterthanhresx\%$ of the remaining \numtargetswithoutfifty{} targets.
When excluding levels 50 and 100 \unit{hPa}, \ourmodel{} significantly outperforms HRES on $\percentsignificantlybetterthanhresxx\%$ of the \numtargetswithoutfiftyandhundred{} remaining targets.
When conducting per region evaluations, we found the previous results to generally hold across the globe, as detailed in Supplements~\cref{fig:app:region_diagram,fig:app:regional_results_1,fig:app:regional_results_2}.

We found that increasing the number of auto-regressive steps in the MSE loss improves \ourmodel performance at longer lead time (see Supplements~\cref{sec:app:AR_ablation}) and encourages it to express its uncertainty by predicting spatially smoothed outputs, leading to blurrier forecasts at longer lead times (see Supplements~\cref{sec:app:spectra}).
HRES's underlying physical equations, however, do not lead to blurred predictions.
To assess whether \ourmodel's relative advantage over HRES on RMSE skill is maintained if HRES is also allowed to blur its forecasts, we fit blurring filters to \ourmodel{} and to HRES, by minimizing the RMSE with respect to the models' respective ground truths. We found that optimally blurred \ourmodel{} has greater skill than optimally blurred HRES on $\percentbetterthanhresoptimalblurring\%$ of our \numtargets{} verification targets which is generally consistent with our above conclusions (see Supplements~\cref{sec:app:optimalfiltering}).

We also compared \ourmodel's performance to the top competing ML-based weather model, \pangumodel{}~\cite{bi2022pangu}, and found \ourmodel{} outperformed it on $\percentbetterthanpangu\%$ of the \numtargetspangu{} targets they presented (see Supplements~\cref{sec:app:baselinedetails} for details).

\begin{figure}%
  \centering
  \includegraphics[width=\textwidth]{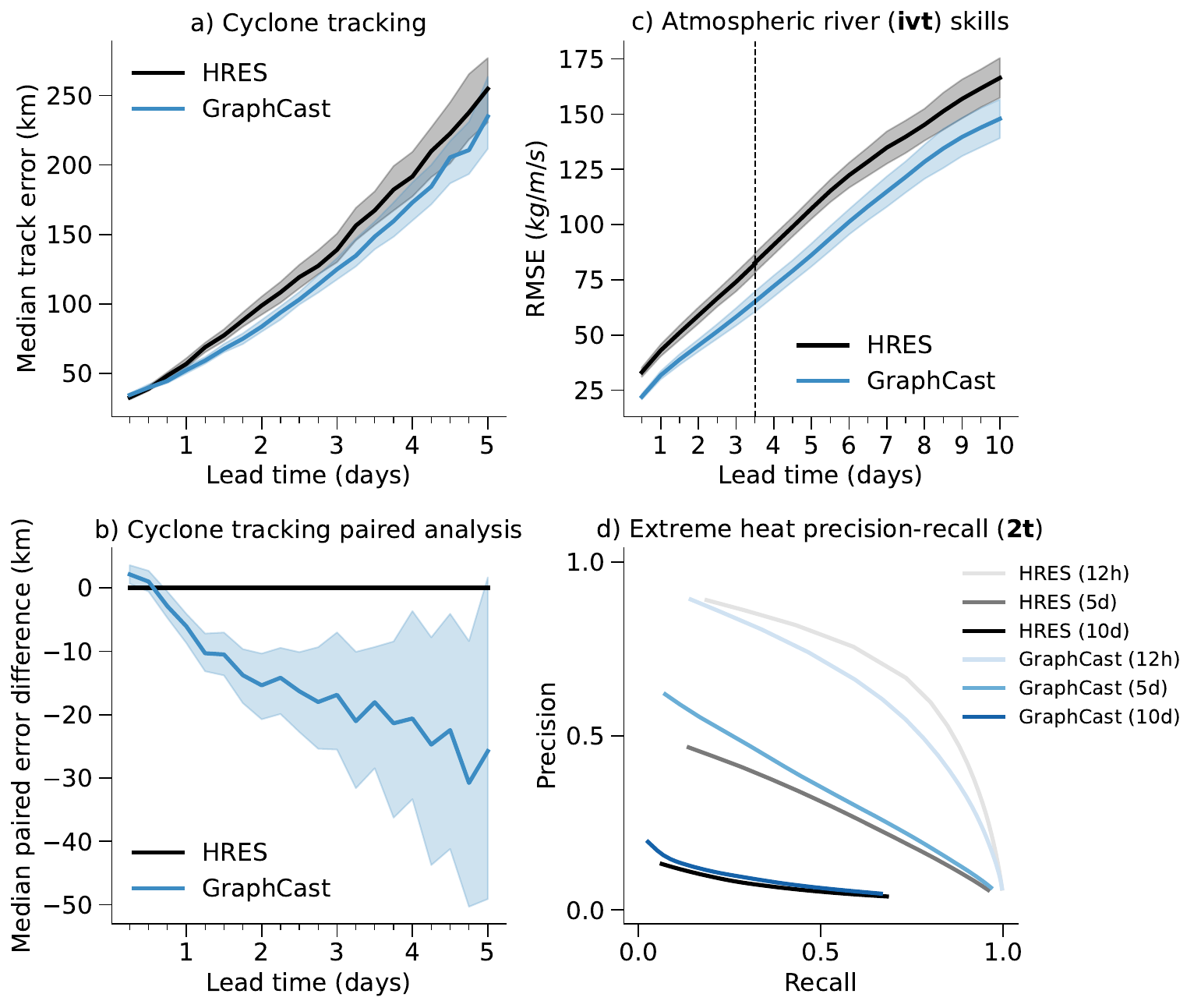}
  \caption{\small\textbf{Severe-event prediction.} (a) Cyclone forecasting performances for \ourmodel{} and HRES. The x-axis represents lead times (in days), and the y-axis represents median track error (in km). Error bars represent  bootstrapped 95\% confidence intervals for the median. (b) Cyclone forecasting paired error difference between \ourmodel{} and HRES. The x-axis represents lead times (in days), and the y-axis represents median paired error difference (in km). Error bars represent bootstrapped 95\% confidence intervals for the median difference (see Supplements~\cref{app:cyclones}). (c) Atmospheric river prediction (\varlevel{ivt}) skills for \ourmodel{} and HRES. The x-axis represents lead times (in days), and the y-axis represents RMSE. Error bars are 95\% confidence intervals. (d) Extreme heat prediction precision-recall for \ourmodel{} and HRES. The x-axis represents recall, and the y-axis represents precision. The curves represent different precision-recall trade-offs when sweeping over gain applied to forecast signals (see Supplements~\cref{sec:app:extremetemperature}).}
  \label{fig:resultsextremes}
\end{figure}

\section*{Severe event forecasting results}
Beyond evaluating \ourmodel's forecast skill against HRES's on a wide range of variables and lead times, we also evaluate how its forecasts support predicting severe events, including tropical cyclones, atmospheric rivers, and extreme temperature. These are key downstream applications for which \ourmodel is not specifically trained, but which are very important for human activity.

\subsection*{Tropical cyclone tracks}
Improving the accuracy of tropical cyclone forecasts can help avoid injury and loss of life, as well as reducing economic harm \cite{martinez2020forecast}.
A cyclone's existence, strength, and trajectory is predicted by applying a tracking algorithm to forecasts of geopotential (\varlevel{z}), horizontal wind (\varlevel{10u}/\varlevel{10v}, \varlevel{U}/\varlevel{V}), and mean sea-level pressure (\varlevel{msl}).
We implemented a tracking algorithm based on ECMWF's published protocols \cite{magnusson2021tropical} and applied it to \ourmodel{}'s forecasts, to produce cyclone track predictions~(see Supplements~\cref{app:cyclones}).
As a baseline for comparison, we used the operational tracks obtained from HRES's \pointonedegree forecasts, stored in the TIGGE archive~\cite{bougeault2010thorpex, swinbank2016},
and measured errors for both models against the tracks from IBTrACS~\cite{knapp2010international,ibtracs-dataset-2018}, a separate reanalysis dataset of cyclone tracks aggregated from various analysis and observational sources.
Consistent with established evaluation of tropical cyclone prediction \cite{magnusson2021tropical}, we evaluate all tracks when both \ourmodel{} and HRES detect a cyclone, ensuring that both models are evaluated on the same events, and verify that each model's true-positive rates are similar.

\cref{fig:resultsextremes}a shows \ourmodel{} has lower median track error than HRES over 2018--2021.
As per-track errors for HRES and \ourmodel are correlated, we also measured the per-track paired error difference between the two models and found that \ourmodel is significantly better than HRES for lead time 18 hours to 4.75 days, as shown in~\cref{fig:resultsextremes}b. The error bars show the bootstrapped 95\% confidence intervals for the median (see Supplements \cref{app:cyclones} for details).

\subsection*{Atmospheric rivers}

Atmospheric rivers are narrow regions of the atmosphere which are responsible for the majority of the poleward water vapor transport across the mid-latitudes, and generate 30\%-65\% of annual precipitation on the U.S. West Coast \cite{chapman2019improving}. Their strength can be characterized by the vertically integrated water vapor transport \varlevel{ivt} \cite{neiman2008meteorological, moore2012physical}, indicating whether an event will provide beneficial precipitation or be associated with catastrophic damage~\cite{corringham2019atmospheric}.
\varlevel{ivt} can be computed from the non-linear combination of the horizontal wind speed (\varlevel{u} and \varlevel{v}) and specific humidity (\varlevel{q}), which \ourmodel{} predicts.
We evaluate \ourmodel{} forecasts over coastal North America and the Eastern Pacific during cold months (Oct--Apr), when atmospheric rivers are most frequent.
Despite not being specifically trained to characterize atmospheric rivers, \cref{fig:resultsextremes}c shows that \ourmodel{} improves the prediction of \varlevel{ivt} compared to HRES, from 25\% at short lead time, to 10\% at longer horizons (see Supplements \cref{sec:app:atmosphericriver} for details).

\subsection*{Extreme heat and cold}
Extreme heat and cold are characterized by large anomalies with respect to typical climatology \cite{magnusson2014verification, lopez2022global, ecmwf2022heatwaveuk}, which can be dangerous and disrupt human activities.
We evaluate the skill of HRES and \ourmodel{} in predicting events above the top 2\% climatology across location, time of day, and month of the year, for \varlevel{2t} at 12-hour, 5-day, and 10-day lead times, for land regions across northern and southern hemisphere over summer months. We plot precision-recall curves
\cite{saito2015precision} to reflect different possible trade-offs between reducing false positives (high precision) and reducing false negatives (high recall). For each forecast, we obtain the curve by varying a ``gain'' parameter that scales the \varlevel{2t} forecast's deviations with respect to the median climatology.

\cref{fig:resultsextremes}d shows \ourmodel{}'s precision-recall curves are above HRES's for 5- and 10-day lead times, suggesting \ourmodel{}'s forecasts are generally superior than HRES at extreme classification over longer horizons. By contrast, HRES has better precision-recall at the 12-hour lead time, which is consistent with the \varlevel{2t} skill score of \ourmodel{} over HRES being near zero, as shown in~\cref{fig:resultshres}d.
We generally find these results to be consistent across other variables relevant to extreme heat, such as \varlevel{t850} and \varlevel{z500}~\cite{ecmwf2022heatwaveuk}, other extreme thresholds (5\%, 2\% and 0.5\%), and extreme cold forecasting in winter. See Supplements~\cref{sec:app:extremetemperature} for details.

\section*{Effect of training data recency}

\ourmodel{} can be re-trained periodically with recent data, which in principle allows it to capture weather patterns that change over time, such as the ENSO cycle and other oscillations, as well as effects of climate change.
We trained four variants of \ourmodel{} with data that always began in 1979, but ended in 2017, 2018, 2019, and 2020, respectively (we label the variant ending in 2017 as ``\ourmodel:$<$2018'', etc). We compared their performances to HRES on 2021 test data.

\begin{figure}%
  \centering
  \includegraphics[width=0.5\textwidth]{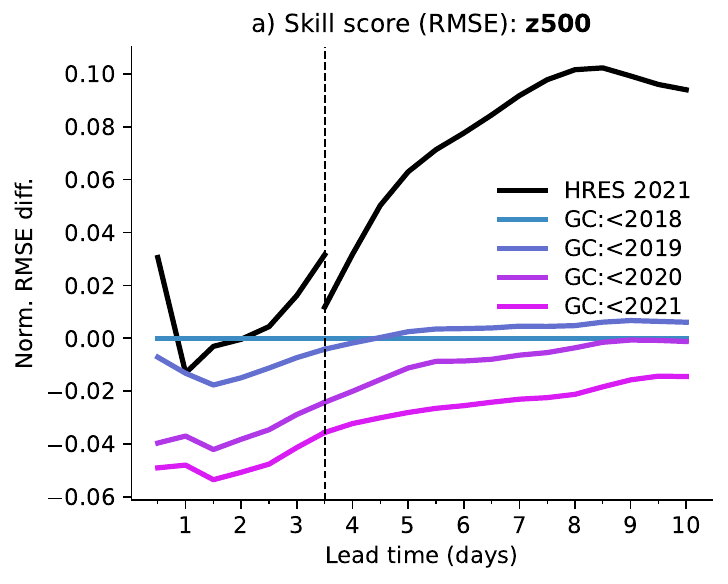}
  \caption{\small\textbf{Training \ourmodel{} on more recent data.} Each colored line represents \ourmodel{} trained with data ending before a different year, from 2018 (blue) to 2021 (purple). The y-axis represents RMSE skill scores on 2021 test data, for \varlevel{z}{500}, with respect to \ourmodel{} trained up to before 2018, over lead times (x-axis). The vertical dashed line represents 3.5 days, where the HRES 06z/18z forecasts end. The black line represents HRES, where lead times earlier and later than 3.5 days are from the 06z/18z and 00z/12z initializations, respectively.}
  \label{fig:resultsablations}
\end{figure}

\cref{fig:resultsablations} shows the skill scores (normalized by \ourmodel:$<$2018) of the four variants and HRES, for \varlevel{z}{500}. We found that while \ourmodel{}'s performance when trained up to before 2018 is still competitive with HRES in 2021, training it up to before 2021 further improves its skill scores (see Supplements~\cref{sec:app:datarecency}). We speculate this recency effect allows recent weather trends to be captured to improve accuracy. 
This shows that \ourmodel's performance can be improved by re-training on more recent data.

\section*{Conclusions}

\ourmodel{}'s forecast skill and efficiency compared to HRES shows MLWP methods are now competitive with traditional weather forecasting methods. Additionally, \ourmodel{}'s performance on severe event forecasting, which it was not directly trained for, demonstrates its robustness and potential for downstream value. We believe this marks a turning point in weather forecasting, which helps open new avenues to strengthen the breadth of weather-dependent decision-making by individuals and industries, by making cheap prediction more accurate, more accessible, and suitable for specific applications.

With 36.7 million parameters, \ourmodel{} is a relatively small model by modern ML standards, chosen to keep the memory footprint tractable.
And while HRES is released on \ang{0.1} resolution, 137 levels, and up to 1 hour time steps, \ourmodel{} operated on \quarterdegree{} latitude-longitude resolution, 37 vertical levels, and 6 hour time steps, because of the ERA5 training data's native \quarterdegree{} resolution, and engineering challenges in fitting higher resolution data on hardware.
Generally \ourmodel{} should be viewed as a family of models, with the current version being the largest we can practically fit under current engineering constraints, but which have potential to scale much further in the future with greater compute resources and higher resolution data.

One key limitation of our approach is in how uncertainty is handled.
We focused on deterministic forecasts and compared against HRES, but the other pillar of ECMWF's IFS, the ensemble forecasting system, ENS, is especially important for 10+ day forecasts. The non-linearity of weather dynamics means there is increasing uncertainty at longer lead times, which is not well-captured by a single deterministic forecast. ENS addresses this by generating multiple, stochastic forecasts, which model the empirical distribution of future weather, however generating multiple forecasts is expensive. By contrast, \ourmodel{}'s MSE training objective encourages it to express its uncertainty by spatially blurring its predictions, which may limit its value for some applications.
Building systems that model uncertainty more explicitly is a crucial next step.

It is important to emphasize that data-driven MLWP depends critically on large quantities of high-quality data, assimilated via NWP, and that rich data sources like ECMWF's MARS archive are invaluable. Therefore, our approach should not be regarded as a replacement for traditional weather forecasting methods, which have been developed for decades, rigorously tested in many real-world contexts, and offer many features we have not yet explored.
Rather our work should be interpreted as evidence that MLWP is able to meet the challenges of real-world forecasting problems, and has potential to complement and improve the current best methods.

Beyond weather forecasting, \ourmodel{} can open new directions for other important geo-spatiotemporal forecasting problems, including climate and ecology, energy, agriculture, and human and biological activity, as well as other complex dynamical systems. We believe that learned simulators, trained on rich, real-world data, will be crucial in advancing the role of machine learning in the physical sciences.

\section*{Data and Materials Availability}
GraphCast's code and trained weights are publicly available on github \url{https://github.com/deepmind/graphcast}. This work used publicly available data from the European Centre for Medium Range Forecasting (ECMWF). We use the  ECMWF archive (expired real-time) products for ERA5, HRES and TIGGE products, whose use is governed by the Creative Commons Attribution 4.0 International (CC BY 4.0). We use IBTrACS Version 4 from \url{https://www.ncei.noaa.gov/products/international-best-track-archive} and reference~\cite{knapp2010international,ibtracs-dataset-2018} as required.
The Earth texture in figure 1 is used under CC~BY~4.0 from \protect\url{https://www.solarsystemscope.com/textures/}.

\section*{Acknowledgments}
In alphabetical order, we thank Kelsey Allen, Charles Blundell, Matt Botvinick, Zied Ben Bouallegue, Michael Brenner, Rob Carver, Matthew Chantry, Marc Deisenroth, Peter Deuben, Marta Garnelo, Ryan Keisler, Dmitrii Kochkov, Christopher Mattern, Piotr Mirowski, Peter Norgaard, Ilan Price, Chongli Qin, Sébastien Racanière, Stephan Rasp, Yulia Rubanova, Kunal Shah, Jamie Smith, Daniel Worrall, and countless others at Alphabet and ECMWF for advice and feedback on our work. We also thank ECMWF for providing invaluable datasets to the research community. The style of the opening paragraph was inspired by D. Fan et al., Science Robotics, 4 (36), (2019).

\putbib[references]
\end{bibunit}

\newpage

\begin{bibunit}

\section*{Supplementary materials}

Supplements S1-S9
Figures 5-53
Tables 3-5

\startcontents[sections]
\printcontents[sections]{}{1}{}

\newpage 
\section{Datasets}\label{sec:app:datasets}

In this section, we give an overview of the data we used to train and evaluate \ourmodel (Supplements~\cref{sec:app:era5}), the data defining the forecasts of the NWP baseline HRES, as well as HRES-fc0, which we use as ground truth for HRES (Supplements~\cref{sec:app:hres}). Finally, we describe the data used in the tropical cyclone analysis (\cref{sec:app:tropicalcyclonedataset}).

We constructed multiple datasets for training and evaluation, comprised of subsets of ECMWF's data archives and IBTrACS~\cite{knapp2010international,ibtracs-dataset-2018}. We generally distinguish between the source data, which we refer to as ``archive'' or ``archived data'', versus the datasets we have built from these archives, which we refer to as ``datasets''.

\subsection{ERA5}\label{sec:app:era5}

For training and evaluating \ourmodel, we built our datasets from a subset of ECMWF's ERA5~\cite{hersbach2020era5}\footnote{See ERA5 documentation: \url{https://confluence.ecmwf.int/display/CKB/ERA5}.} archive, which is a large corpus of data that represents the global weather from 1959 to the present, at \quarterdegree{} latitude/longitude resolution, and 1 hour increments, for hundreds of static, surface, and atmospheric variables. The ERA5 archive is based on \textit{reanalysis}, which uses ECMWF's HRES model (cycle 42r1) that was operational for most of 2016 (see~\cref{app:tab:hrescycles}), within ECMWF's 4D-Var data assimilation system. ERA5 assimilated 12-hour windows of observations, from 21z-09z and 09z-21z, as well as previous forecasts, into a dense representation of the weather's state, for each historical date and time.

Our ERA5 dataset contains a subset of available variables in ECMWF's ERA5 archive (\cref{tab:app:variables}), on 37 pressure levels\footnote{We follow common practice of using pressure as our vertical coordinate, instead of altitude. A ``pressure level'' is a field of altitudes with equal pressure. E.g., ``pressure level 500 \unit{hPa}'' corresponds to the field of altitudes for which the pressure is 500 \unit{hPa}. The relationship between pressure and altitude is determined by the geopotential variable.}:
1, 2, 3, 5, 7, 10, 20, 30, 50, 70, 100, 125, 150, 175, 200, 225, 250, 300, 350, 400, 450,
500, 550, 600, 650, 700, 750, 775, 800, 825, 850, 875, 900, 925, 950, 975, 1000 \unit{hPa}.
The range of years included was 1979-01-01 to 2022-01-10, which were downsampled to 6 hour time intervals (corresponding to 00z, 06z, 12z and 18z each day). The downsampling is performed by subsampling, except for the total precipitation, which is accumulated for the 6 hours leading up to the corresponding downsampled time.

\begin{center}
\begin{table}[htbp]
\centering
\begin{tabular}{||c | c | c | c | c||} 
 \hline
 Type & Variable name & Short & ECMWF & Role (accumulation \\ [0.5ex] 
  &  &  name & Parameter ID & period, if applicable) \\ [0.5ex] 
 \hline\hline
 Atmospheric & Geopotential & z & 129 & Input/Predicted \\
\hline
Atmospheric & Specific humidity & q & 133 & Input/Predicted \\
\hline
Atmospheric & Temperature & t & 130 & Input/Predicted \\
\hline
Atmospheric & U component of wind & u & 131 & Input/Predicted \\
\hline
Atmospheric & V component of wind & v & 132 & Input/Predicted \\
\hline
Atmospheric & Vertical velocity & w & 135 & Input/Predicted \\
\hline
Single & 2 metre temperature & 2t & 167 & Input/Predicted 
\\
 \hline
Single & 10 metre u wind component & 10u & 165 & Input/Predicted \\
\hline
Single & 10 metre v wind component & 10v & 166 & Input/Predicted \\
\hline
Single & Mean sea level pressure & msl & 151 & Input/Predicted \\
 \hline
Single & Total precipitation & tp & 228 & Input/Predicted (6h) \\
 \hline
 \hline
Single & TOA incident solar radiation & tisr & 212 & Input (1h) \\
\hline
Static & Geopotential at surface & z & 129 & Input \\
\hline
Static & Land-sea mask & lsm & 172 & Input \\
\hline
Static & Latitude & n/a & n/a & Input \\
\hline
Static & Longitude & n/a & n/a & Input \\
\hline
Clock & Local time of day & n/a & n/a & Input \\
\hline
Clock & Elapsed year progress & n/a & n/a & Input \\
 \hline
\end{tabular}
\caption{\small\textbf{ECMWF variables used in our datasets.} The ``Type'' column indicates whether the variable represents a \textit{static} property, a time-varying \textit{single}-level property (e.g., surface variables are included), or a time-varying \textit{atmospheric} property. The ``Variable name'' and ``Short name'' columns are ECMWF's labels. The ``ECMWF Parameter ID'' column is a ECMWF's numeric label, and can be used to construct the URL for ECMWF's description of the variable, by appending it as suffix to the following prefix, replacing ``ID'' with the numeric code: \texttt{https://apps.ecmwf.int/codes/grib/param-db/?id=ID}. The ``Role'' column indicates whether the variable is something our model takes as input and predicts, or only uses as input context (the double horizontal line separates predicted from input-only variables, to make the partitioning more visible).}
\label{tab:app:variables}
\end{table}
\end{center}

\subsection{HRES}\label{sec:app:hres}

Evaluating the HRES model baseline requires two separate sets of data, namely the forecast data and the ground truth data, which are summarized in the subsequent sub-sections. The HRES versions which were operational during our test years are shown in~\cref{app:tab:hrescycles}.

\begin{center}
\begin{table}[htbp]
\centering
\begin{tabular}{||c | l | c | l||} 
 \hline
 IFS cycle & Dates of operation & Used in ERA5 & HRES evaluation year(s) \\ [0.5ex] 
 \hline\hline
 42r1 & 2016-03-08 -- 2016-11-21 & $\checkmark$ & --\\
 \hline
 43r1 & 2016-11-22 -- 2017-07-10 & \ & -- \\
 \hline
 43r3 & 2017-07-11 -- 2018-06-04 & \ & 2018 \\
 \hline
 45r1 & 2018-06-05 -- 2019-06-10 & \ & 2018, 2019 \\
 \hline
 46r1 & 2019-06-11 -- 2020-06-29 & \ & 2019, 2020 \\
 \hline
 47r1 & 2020-06-30 -- 2021-05-10 & \ & 2020, 2021 \\
 \hline
 47r2 & 2021-05-11 -- 2021-10-11 & \ & 2021 \\
 \hline
 47r3 & 2021-10-12 -- present & \ & 2021, 2022 \\
 \hline
\end{tabular}
\caption{\small\textbf{\pointonedegree resolution IFS cycles since 2016.}  The table shows every IFS cycle that operated at \pointonedegree latitude/longitude resolution. The columns represent the IFS cycle version, its dates of operation, whether it was used for data assimilation for ERA5, and the years it was used as a baseline for comparing to \ourmodel{} in our results evaluation. See 
\protect\url{https://www.ecmwf.int/en/forecasts/documentation-and-support/changes-ecmwf-model} for the full cycle release schedule.}
\label{app:tab:hrescycles}
\end{table}
\end{center}

\paragraph{HRES operational forecasts}\label{sec:app:hresfc}
HRES is generally considered to be the most accurate deterministic NWP-based weather model in the world, so to evaluate the HRES baseline, we built a dataset of HRES's archived historical forecasts. HRES is regularly updated by ECMWF, so these forecasts represent the latest HRES model at the time the forecasts were made.  The forecasts were downloaded at their native representation (which uses spherical harmonics and an octahedral reduced Gaussian grid, TCo1279~\cite{a-new-grid-for-the-ifs}), and roughly corresponds to \pointonedegree latitude/longitude resolution. We then spatially downsampled the forecasts to a \quarterdegree{} latitude/longitude grid (to match ERA5's resolution) using ECMWF's Metview library, with default 
$\mathtt{regrid}$ parameters. We temporally downsampled them to 6 hour intervals. There are two groups of HRES forecasts: those initialized at 00z/12z which are released for 10 day horizons, and those initialized at 06z/18z which are released for 3.75 day horizons.

\begin{figure}
  \centering
  \includegraphics[width=\textwidth]{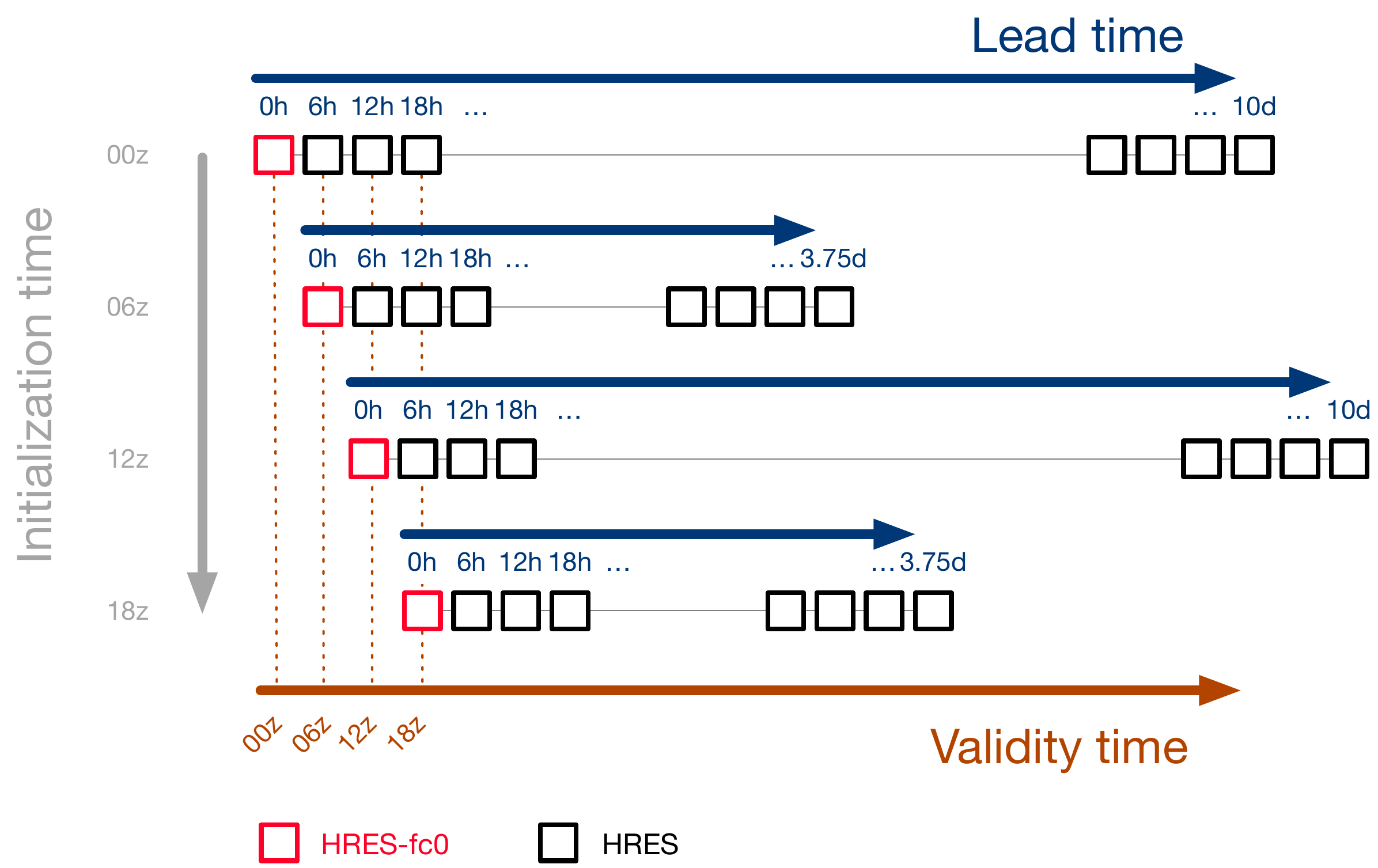}
  \caption{\small\textbf{Schematic of HRES-fc0.} Each horizontal line represent a forecast made by HRES, initialized at a different time (grey axis). HRES forecasts initialized from 00z and 12z make predictions up to 10 days lead time (blue axis), while HRES forecasts initialized from 06z and 18z make predictions up to 3.75 days. Each square represent a state predicted by HRES, by 6 hours increments (smaller time steps are omitted from the schematic, as well as states in the middle of a forecast trajectory). Red squares represent the forecast at time 0 for each HRES forecast, and defines the data points included in \hresfczero{}. The brown axis represents the validity time and allows visualizing the alignment of predictions from different initialization time. For instance, the error of the prediction made by HRES, initialized at 06z (second row of squares from the top), at 12h lead time, i.e., 18z validity time (3rd square from the left) would be measured against the first step of the HRES forecast initialized at 18z (red square from the last row of square).}
  \label{fig:app:HRES-fc0_schematic}
\end{figure}

\paragraph{\hresfczero{}}\label{sec:app:hresfczero}
For evaluating the skill of the HRES operational forecasts, we constructed a ground truth dataset, ``HRES-fc0'', based on ECMWF's HRES operational forecast archive.
This dataset comprises the initial time step of each HRES forecast, at initialization times 00z, 06z, 12z, and 18z (see \cref{fig:app:HRES-fc0_schematic}). The \hresfczero{} data is similar to the ERA5 data, but it is assimilated using the latest ECMWF NWP model at the forecast time, and assimilates observations from $\pm$3 hours around the corresponding date and time. Note, ECMWF also provides an archive of ``HRES Analysis'' data, which is distinct from our \hresfczero{} dataset. The HRES Analysis dataset includes both atmospheric and land surface analyses, but is not the input which is provided to the HRES forecasts, therefore we do not use it as ground truth because it would introduce discrepancies between HRES forecasts and ground truth, simply due to HRES using different inputs, which would be especially prominent at short lead times.

\paragraph{HRES NaN handling}\label{sec:app:hresnan}
A very small subset of the values from the ECMWF HRES archive for the variable geopotential at 850\unit{hPa} (\varlevel{z}{850}) and 925\unit{hPa} (\varlevel{z}{925}) are not numbers (NaN). These NaN's seem to be distributed uniformly across the 2016-2021 range and across forecast times.
This represents about 0.00001$\%$ of the pixels for \varlevel{z}{850} (1 pixel every ten 1440 x 721 latitude-longitude frames), 0.00000001$\%$ of the pixels for \varlevel{z}{925} (1 pixel every ten thousand 1440 x 721 latitude-longitude frames) and has no measurable impact on performance. For easier comparison, we filled these rare missing values with the weighted average of the immediate neighboring pixels. We used a weight of 1 for side-to-side neighbors and 0.5 weights for diagonal neighbors\footnote{In the extremely rare cases that the neighbors were also NaN's, we dropped both the NaN neighbor and the opposed neighbor from the weighted average.}.

\subsection{Tropical cyclone datasets}\label{sec:app:tropicalcyclonedataset}
For our analysis of tropical cyclone forecasting, we used the IBTrACS~\cite{ibtracs-dataset-2018,knapp2010international,levinson2010toward,kruk2010technique} archive to construct the ground truth dataset. This includes historical cyclone tracks from around a dozen authoritative sources.
Each track is a time series, at 6-hour intervals (00z, 06z, 12z, 18z), where each timestep represents the eye of the cyclone in latitude/longitude coordinates, along with the corresponding Saffir-Simpson category and other relevant meteorological features at that point in time.

For the HRES baseline, we used the TIGGE archive, which provides cyclone tracks estimated with the operational tracker, from HRES's forecasts at \pointonedegree resolution~\cite{bougeault2010thorpex,swinbank2016}.
The data is stored as XML files available for download under~\url{https://confluence.ecmwf.int/display/TIGGE/Tools}.
To convert the data into a format suitable for further post-processing and analysis, we implemented a parser that extracts cyclone tracks for the years of interest. The relevant sections (tags) in the XML files are those of type ``forecast'', which typically contain multiple tracks corresponding to different initial forecast times. Within these tags, we then extract the cyclone name (tag ``cycloneName''), the latitude (tag ``latitude'') and the longitude (tag ``longitude'') values, and the valid time (tag ``validTime'').

See \cref{app:cyclones} for details of the tracker algorithm and results.

\newpage
\section{Notation and problem statement}\label{sec:app:notationsandproblemstatement}

In this section, we define useful time notations use throughout the paper (\cref{sec:app:timenotation}), formalize the general forecasting problem we tackle (\cref{sec:app:problemstatement}), and detail how we model the state of the weather (\cref{sec:app:modelingecmwf}).

\subsection{Time notation}\label{sec:app:timenotation}
The time notation used in forecasting can be confusing, involving a number of different time symbols, e.g., to denote the initial forecast time, validity time, forecast horizon, etc. We therefore introduce some standardized terms and notation for clarity and simplicity. We refer to a particular point in time as ``date-time'', indicated by calendar date and UTC time. For example, \datetime{2018-06-21}{18:00:00} means June 21, 2018, at 18:00 UTC. For shorthand, we also sometimes use the Zulu convention, i.e., 00z, 06z, 12z, 18z mean \texttt{00:00}, \texttt{06:00}, \texttt{12:00}, \texttt{18:00} UTC, respectively. We further define the following symbols:
\begin{itemize}
    \item $\ttt$: Forecast time step index, which indexes the number of steps since the forecast was initialized.
    \item $\T$: Forecast horizon, which represents the total number of steps in a forecast.
    \item $\dd$: Validity time, which indicates the date-time of a particular weather state.
    \item $\dinit$: Forecast initialization time, indicating the validity time of a forecast's initial inputs.
    \item $\dsize$: Forecast step duration, indicating how much time elapses during one forecast step.
    \item $\lt$: Forecast lead time, which represents the elapsed time in the forecast (i.e., $\lt=\ttt\dsize$).
\end{itemize}

\subsection{General forecasting problem statement}\label{sec:app:problemstatement}
Let $\Z^{\dd}$ denote the true state of the global weather at time $\dd$. The time evolution of the true weather can be represented by an underlying discrete-time dynamics function, $\fwdtrue$, which generates the state at the next time step ($\dsize$ in the future) based on the current one, i.e., $\Z^{\dd+\dsize} = \fwdtrue(\Z^{\dd})$. We then obtain a trajectory of $T$ future weather states by applying $\fwdtrue$ autoregressively $T$ times,
\begin{align}
    \Z^{\dd+\dsize:\dd+\T\dsize} &= (\underbrace{\fwdtrue(\Z^{\dd}), \fwdtrue(\Z^{\dd+\dsize}), \dots, \fwdtrue(\Z^{\dd+(\T-1)\dsize})}_{1 \dots \T \text{ autoregressive iterations}}).
\label{eq:app:fwdtrue}
\end{align}

Our goal is to find an accurate and efficient model, $\fwdmodel$, of the true dynamics function, $\fwdtrue$, that can efficiently forecast the state of the weather over some forecast horizon, $\T\dsize$. We assume that we cannot observe $\Z^{\dd}$ directly, but instead only have some partial observation $\X^{\dd}$, which is an incomplete representation of the state information required to predict the weather perfectly. Because $\X^{\dd}$ is only an approximation of the instantaneous state $\Z^{\dd}$, we also provide $\fwdmodel$ with one or more past states, $\X^{\dd-\dsize}, \X^{\dd-2\dsize}, ...$, in addition to $\X^{\dd}$. The model can then, in principle, leverage this additional context information to approximate $\Z^{\dd}$ more accurately. Thus $\fwdmodel$ predicts a future weather state as,
\begin{align}
\Xpred^{\dd+\dsize} &= \fwdmodel(\X^{\dd}, \X^{\dd-\dsize}, ...).
\label{eq:app:fwdmodel}
\end{align}
Analogous to \cref{eq:app:fwdtrue}, the prediction $\Xpred^{\dd+\dsize}$ can be fed back into $\fwdmodel$ to autoregressively produce a full forecast,
\begin{align}
\Xpred^{\dd+\dsize:\dd+\T\dsize} &= (\underbrace{\fwdmodel(\X^{\dd}, \X^{\dd-\dsize}, ...), \fwdmodel(\Xpred^{\dd+\dsize}, \X^{\dd}, ...), \dots, \fwdmodel(\Xpred^{\dd+(\T-1)\dsize}, \Xpred^{\dd+(\T-2)\dsize}, ...)}_{1 \dots \T \text{ autoregressive iterations}}).
\label{eq:app:fwdmodelforecast}
\end{align}

We assess the forecast quality, or skill, of $\fwdmodel$ by quantifying how well the predicted trajectory, $\Xpred^{\dd+\dsize:\dd+\T\dsize}$, matches the ground-truth trajectory, $\X^{\dd+\dsize:\dd+\T\dsize}$. However, it is important to highlight again that $\X^{\dd+\dsize:\dd+\T\dsize}$ only comprises our observations of $\Z^{\dd+\dsize:\dd+\T\dsize}$, which itself is unobserved. We measure the consistency between forecasts and ground truth with an objective function, 
\begin{align*}
    \mathcal{L}\left(\Xpred^{\dd+\dsize:\dd+\T\dsize}, \X^{\dd+\dsize:\dd+\T\dsize}\right),
\end{align*}
which is described explicitly in~\cref{sec:app:evaluationdetails}.

In our work, the temporal resolution of data and forecasts was always $\dsize=6$ hours with a maximum forecast horizon of 10 days, corresponding to a total of $T=40$ steps. Because $\dsize$ is a constant throughout this paper, we can simplify the notation using $(\X^{\ttt}, \X^{\ttt+1}, \dots, \X^{\ttt+\T})$ instead of $(\X^{\dd}, \X^{\dd+\dsize}, \dots, \X^{\dd+\T\dsize})$, to index time with an integer instead of a specific date-time.

\subsection{Modeling ECMWF weather data}\label{sec:app:modelingecmwf}
For training and evaluating models, we treat our ERA5 dataset as the ground truth representation of the surface and atmospheric weather state. As described in \cref{sec:app:hres}, we used the \hresfczero{} dataset as ground truth for evaluating the skill of HRES.

In our dataset, an ERA5 weather state $\X^{\ttt}$ comprises all variables in \cref{tab:app:variables}, at a \quarterdegree{} horizontal latitude-longitude resolution with a total of $721 \times 1440 = 1,038,240$ grid points and 37 vertical pressure levels. The atmospheric variables are defined at all pressure levels and the set of (horizontal) grid points is given by $\Gquarterdegree = \{-90.0, -89.75, \dots, 90.0\} \times \{-179.75, -179.5, \dots, 180.0\}$.
These variables are uniquely identified by their short name (and the pressure level, for atmospheric variables). For example, the surface variable ``2 metre temperature'' is denoted \varlevel{2t}; the atmospheric variable ``Geopotential'' at pressure level 500 \unit{hPa} is denoted \varlevel{z}{500}. Note, only the ``predicted'' variables are output by our model, because the ``input''-only variables are forcings that are known apriori, and simply appended to the state on each time-step. We ignore them in the description for simplicity, so in total there are 5 surface variables and 6 atmospheric variables.

From all these variables, our model predicts 5 surface variables and 6 atmospheric variables for a total of 227 target variables.
Several other static and/or external variables were also provided as input context for our model. These variables are shown in~\cref{tab:variables} and \cref{tab:app:variables}.
The static/external variables include information such as the geometry of the grid/mesh, orography (surface geopotential), land-sea mask and radiation at the top of the atmosphere.

We refer to the subset of variables in~$\X^{\ttt}$ that correspond to a particular grid point~$i$ (1,038,240 in total) as~$\xvec^{\ttt}_i$, and to each variable~$j$ of the 227 target variables as~$\x^{\ttt}_{i,j}$. The full state representation $\X^{\ttt}$ therefore contains a total of $721 \times 1440 \times (5 + 6 \times 37) = 235,680,480$ values. Note, at the poles, the $1440$ longitude points are equal, so the actual number of distinct grid points is slightly smaller.

\newpage

\section{\ourmodel{} model}\label{sec:app:ourmodel}

This section provides a detailed description of \ourmodel{}, starting with the autoregressive generation of a forecast (\cref{sec:app:generatingaforecast}), an overview of the architecture in plain language (\cref{sec:app:architectureoverview}), followed by a technical description the all the graphs defining \ourmodel (\cref{sec:app:modeldetails}),  its encoder (\cref{sec:app:encoder}), processor (\cref{sec:app:processor}), and decoder (\cref{sec:app:decoder}), as well as all the normalization and parameterization details (\cref{sec:app:othermodeldetails}).

\subsection{Generating a forecast}\label{sec:app:generatingaforecast}

Our \ourmodel{} model is defined as a one-step learned simulator that takes the role of $\fwdmodel$ in \cref{eq:app:fwdmodel} and predicts the next step based on two consecutive input states,
\begin{align}
    \Xpred^{\ttt+1} = \text{\ourmodel{}}(\X^{\ttt}, \X^{\ttt-1}).
\end{align}
As in \cref{eq:app:fwdmodelforecast}, we can apply \ourmodel{} iteratively to produce a forecast
\begin{align}
    \Xpred^{\ttt+1:\ttt+\T} = (\underbrace{\text{\ourmodel}(\X^{\ttt}, \X^{\ttt-1}), \text{\ourmodel}(\Xpred^{\ttt+1}, \X^{\ttt}), \dots, \text{\ourmodel}(\Xpred^{\ttt+\T-1}, \Xpred^{\ttt+\T-2})}_{1 \dots \T \text{ autoregressive iterations}})
    \label{eq:app:ourmodelforecast}
\end{align}
of arbitrary length, $\T$. This is illustrated in ~\cref{fig:schematic}b,c. We found, in early experiments, that two input states yielded better performance than one, and that three did not help enough to justify the increased memory footprint.

\subsection{Architecture overview} \label{sec:app:architectureoverview}

The core architecture of \ourmodel{} uses GNNs in an ``encode-process-decode'' configuration \cite{battaglia2018relational}, as depicted in~\cref{fig:schematic}d,e,f. GNN-based learned simulators are very effective at learning complex physical dynamics of fluids and other materials \cite{sanchez2020learning,pfaff2021learning}, as the structure of their representations and computations are analogous to learned finite element solvers \cite{alet2019graph}. A key advantage of GNNs is that the input graph's structure determines what parts of the representation interact with one another via learned message-passing, allowing arbitrary patterns of spatial interactions over any range. By contrast, a convolutional neural network (CNN) is restricted to computing interactions within local patches (or, in the case of dilated convolution, over regularly strided longer ranges). And while Transformers~\cite{vaswani2017attention} can also compute arbitrarily long-range computations, they do not scale well with very large inputs (e.g., the 1 million-plus grid points in \ourmodel's global inputs) because of the quadratic memory complexity induced by computing all-to-all interactions. Contemporary extensions of Transformers often sparsify possible interactions to reduce the complexity, which in effect makes them analogous to GNNs (e.g., graph attention networks~\cite{velivckovic2017graph}).

The way we capitalize on the GNN's ability to model arbitrary sparse interactions is by introducing \ourmodel's internal ``multi-mesh'' representation, which allows long-range interactions within few message-passing steps and has generally homogeneous spatial resolution over the globe. This is in contrast with a latitude-longitude grid which induce a non-uniform distribution of grid points. Using the latitude-longitude grid is not an advisable representation due to its spatial inhomogeneity, and high resolution at the poles which demands disproportionate compute resources.

Our multi-mesh is constructed by first dividing a regular icosahedron (12 nodes and 20 faces) iteratively 6 times to obtain a hierarchy of icosahedral meshes with a total of 40,962 nodes and 81,920 faces on the highest resolution. We leveraged the fact that the coarse-mesh nodes are subsets of the fine-mesh nodes, which allowed us to superimpose edges from all levels of the mesh hierarchy onto the finest-resolution mesh. This procedure yields a multi-scale set of meshes, with coarse edges bridging long distances at multiple scales, and fine edges capturing local interactions. \cref{fig:schematic}g shows each individual refined mesh, and \cref{fig:schematic}e shows the full multi-mesh. 

\ourmodel's encoder (\cref{fig:schematic}d) first maps the input data, from the original latitude-longitude grid, into learned features on the multi-mesh, using a GNN with directed edges from the grid points to the multi-mesh. The processor (\cref{fig:schematic}e) then uses a 16-layer deep GNN to perform learned message-passing on the multi-mesh, allowing efficient propagation of information across space due to the long-range edges. The decoder (\cref{fig:schematic}f) then maps the final multi-mesh representation back to the latitude-longitude grid using a GNN with directed edges, and combines this grid representation, $\Ypred^{\ttt+k}$, with the input state, $\Xpred^{\ttt+k}$, to form the output prediction, $\Xpred^{\ttt+k+1}=\Xpred^{\ttt+k} + \Ypred^{\ttt+k}$. 

The encoder and decoder do not require the raw data to be arranged in a regular rectilinear grid, and can also be applied to arbitrary mesh-like state discretizations~\cite{alet2019graph}. The general architecture builds on various GNN-based learned simulators which have been successful in many complex fluid systems and other physical domains~\cite{sanchez2020learning,pfaff2021learning,fortunato2022multiscale}. Similar approaches were used in weather forecasting \cite{keisler2022forecasting}, with promising results.

On a single Cloud TPU v4 device, \ourmodel{} can generate a \quarterdegree resolution, 10-day forecast (at 6-hour steps) in under 60 seconds. For comparison, ECMWF's IFS system runs on a 11,664-core cluster, and generates a \ang{0.1} resolution, 10-day forecast (released at 1-hour steps for the first 90 hours, 3-hour steps for hours 93-144, and 6-hour steps from 150-240 hours, in about an hour of compute time~\cite{rasp2020weatherbench}. See the HRES release details here: \url{https://www.ecmwf.int/en/forecasts/datasets/set-i.}.

\subsection{\ourmodel{}'s graph} \label{sec:app:modeldetails}

\ourmodel{} is implemented using GNNs in an ``encode-process-decode'' configuration, where the encoder maps (surface and atmospheric) features on the input latitude-longitude grid to a multi-mesh, the processor performs many rounds of message-passing on the multi-mesh, and the decoder maps the multi-mesh features back to the output latitude-longitude grid (see \cref{fig:schematic}).

The model operates on a graph $\mathcal{G} (\mathcal{V}^\text{G}, \mathcal{V}^\text{M}, \mathcal{E}^\text{M},  \mathcal{E}^\text{G2M}, \mathcal{E}^\text{M2G})$, defined in detail in the subsequent paragraphs.

\paragraph{Grid nodes}
$\mathcal{V^\Gsuper}$ represents the set containing each of the grid nodes $v^\Gsuper_i$.
Each grid node represents a vertical slice of the atmosphere at a given latitude-longitude point, $i$.
The features associated with each grid node $v^\Gsuper_i$ are $\mathbf{v}^\text{\G,features}_i = [\xvec^{\ttt-1}_i, \xvec^\ttt_i, \forcing^{\ttt-1}_i,  \forcing^\ttt_i, \forcing^{\ttt+1}_i, \constant_i]$, where $\xvec^\ttt_i$ is the time-dependent weather state $X^t$ corresponding to grid node $v^\Gsuper_i$ and includes all the predicted data variables for all 37 atmospheric levels as well as surface variables.
The forcing terms $\forcing^\ttt$ consist of time-dependent features that can be computed analytically, and do not need to be predicted by \ourmodel{}.
They include the total incident solar radiation at the top of the atmosphere, accumulated over 1 hour, the sine and cosine of the local time of day (normalized to [0, 1)), and the sine and cosine of the of year progress (normalized to [0, 1)).
The constants $\constant_i$ are static features: the binary land-sea mask, the geopotential at the surface, the cosine of the latitude, and the sine and cosine of the longitude.
At \quarterdegree{} resolution, there is a total of $721 \times 1440 = 1,038,240$ grid nodes, each with (5 \emph{surface variables} + 6 \emph{atmospheric variables} $\times$ 37 \emph{levels}) $\times$ 2 \emph{steps} + 5 \emph{forcings} $\times$ 3 \emph{steps} + 5 \emph{constant} = 474 input features.

\paragraph{Mesh nodes}
$\mathcal{V^\text{M}}$ represents the set containing each of the mesh nodes $v^\text{M}_i$.
Mesh nodes are placed uniformly around the globe in a \R-refined icosahedral mesh $M^\R$.
$M^0$ corresponds to a unit-radius icosahedron (12 nodes and 20 triangular faces) with faces parallel to the poles (see \cref{fig:schematic}g).
The mesh is iteratively refined $M^\rr \rightarrow M^{\rr+1}$ by splitting each triangular face into 4 smaller faces, resulting in an extra node in the middle of each edge, and re-projecting the new nodes back onto the unit sphere.\footnote{Note this split and re-project mechanism leads to a maximum difference of 16.4\% and standard deviation of 6.5\% in triangle edge lengths across the mesh.}
Features $\mathbf{v}^\text{M,features}_i$ associated with each mesh node $v^\text{M}_i$ include the cosine of the latitude, and the sine and cosine of the longitude.
\ourmodel{} works with a mesh that has been refined $\R=6$ times, $M^6$, resulting in 40,962 mesh nodes (see Supplementary \cref{tab:app:icosahedral-mesh}), each with the 3 input features.

\begin{center}
\begin{table}[htbp]
\centering
\begin{tabular}{|c |  c  c  c  c c c c|} 
 \hline
 Refinement & 0 & 1 & 2 & 3 & 4 & 5 & 6 \\ [0.5ex] 
 \hline
 Num Nodes & 12 & 42 & 162 & 642 & 2,562 & 10,242 & 40,962 \\ 
 \hline
 Num Faces & 20 & 80 & 320 & 1,280 & 5,120 & 20,480 & 81,920 \\ 
 \hline
 Num Edges & 60 & 240 & 960 & 3,840 & 15,360 & 61,440 & 245,760\\ 
 \hline
 Num Multilevel Edges & 60 & 300 & 1,260 & 5,100 & 20,460 & 81,900 & 327,660\\
 \hline
\end{tabular}
\caption{\small\textbf{Multi-mesh statistics.} Statistics of the multilevel refined icosahedral mesh as function of the refinement level $R$. Edges are considered to be bi-directional and therefore we count each edge in the mesh twice (once for each direction).}
\label{tab:app:icosahedral-mesh}
\end{table}
\end{center}

\paragraph{Mesh edges}
$\mathcal{E}^\text{M}$ are bidirectional edges added between mesh nodes that are connected in the mesh.
Crucially, mesh edges are added to $\mathcal{E}^\text{M}$ for all levels of refinement, i.e., for the finest mesh, $M^6$, as well as for $M^5$, $M^4$, $M^3$, $M^2$, $M^1$ and $M^0$. This is straightforward because of how the refinement process works: the nodes of $M^{\rr-1}$ are always a subset of the nodes in $M^{\rr}$. Therefore, nodes introduced at lower refinement levels serve as hubs for longer range communication, independent of the maximum level of refinement. The resulting graph that contains the joint set of edges from all of the levels of refinement is what we refer to as the ``multi-mesh''. See \cref{fig:schematic}e,g for a depiction of all individual meshes in the refinement hierarchy, as well as the full multi-mesh.

For each edge $e^\text{M}_{v^\text{M}_\text{s} \rightarrow v^\text{M}_\text{r}}$ connecting a sender mesh node $v^\text{M}_\text{s}$ to a receiver mesh node $v^\text{M}_\text{r}$, we build edge features $\edge^\text{M,features}_{v^\text{M}_\text{s} \rightarrow v^\text{M}_\text{r}}$ using the position on the unit sphere of the mesh nodes.
This includes the length of the edge, and the vector difference between the 3d positions of the sender node and the receiver node computed in a local coordinate system of the receiver.
The local coordinate system of the receiver is computed by applying a rotation that changes the azimuthal angle until that receiver node lies at longitude 0, followed by a rotation that changes the polar angle until the receiver also lies at latitude 0.
This results in a total of 327,660 mesh edges (See \cref{tab:app:icosahedral-mesh}), each with 4 input features.

\paragraph{Grid2Mesh edges}%
$\mathcal{E}^\text{G2M}$ are unidirectional edges that connect sender grid nodes to receiver mesh nodes.
An edge $e^\text{G2M}_{v^\Gsuper_\text{s} \rightarrow v^\text{M}_\text{r}}$ is added if the distance between the mesh node and the grid node is smaller or equal than 0.6 times\footnote{Technically it is 0.6 times the ``longest'' edge in $M^6$, since there is some variance in the length of the edges caused by the split-and-reproject mechanism.} the length of the edges in mesh $M^6$ (see \cref{fig:schematic}) which ensures every grid node is connected to at least one mesh node.
Features $\edge^\text{G2M,features}_{v^\Gsuper_\text{s} \rightarrow v^\text{M}_\text{r}}$ are built the same way as those for the mesh edges. This results on a total of 1,618,746 Grid2Mesh edges, each with 4 input features.

\paragraph{Mesh2Grid edges}
$\mathcal{E}^\text{M2G}$ are unidirectional edges that connect sender mesh nodes to receiver grid nodes. For each grid point, we find the triangular face in the mesh $M^6$ that contains it and add three Mesh2Grid edges of the form $e^\text{M2G}_{v^\text{M}_\text{s} \rightarrow v^\Gsuper_\text{r}}$, to connect the grid node to the three mesh nodes adjacent to that face (see \cref{fig:schematic}). Features $\edge^\text{M2G,features}_{v^\text{M}_\text{s} \rightarrow v^\Gsuper_\text{r}}$ are built on the same way as those for the mesh edges. This results on a total of 3,114,720 Mesh2Grid edges (3 mesh nodes connected to each of the $721\times 1440$ latitude-longitude grid points), each with four input features.

\subsection{Encoder}\label{sec:app:encoder}
The purpose of the encoder is to prepare data into latent representations for the processor, which will run exclusively on the multi-mesh.

\paragraph{Embedding the input features}\label{sec:app:embeddinginputfeatures}
As part of the encoder, we first embed the features of each of the grid nodes, mesh nodes, mesh edges, grid to mesh edges, and mesh to grid edges into a latent space of fixed size using five multi-layer perceptrons (MLP),
\begin{equation}
\begin{split}
\mathbf{v}^\Gsuper_i & = \text{MLP}^\text{embedder}_{\mathcal{V}^\Gsuper}(\mathbf{v}^\text{\G,features}_i)
\\
\mathbf{v}^\text{M}_i & = \text{MLP}^\text{embedder}_{\mathcal{V}^\text{M}}(\mathbf{v}^\text{M,features}_i)
\\
\mathbf{e}^\text{M}_{v^\text{M}_\text{s} \rightarrow v^\text{M}_\text{r}} & = \text{MLP}^\text{embedder}_{\mathcal{E}^\text{M}}(\mathbf{e}^\text{M,features}_{v^\text{M}_\text{s} \rightarrow v^\text{M}_\text{r}})
\\
\mathbf{e}^\text{G2M}_{v^\Gsuper_\text{s} \rightarrow v^\text{M}_\text{r}} & = \text{MLP}^\text{embedder}_{\mathcal{E}^\text{G2M}}(\mathbf{e}^\text{G2M,features}_{v^\Gsuper_\text{s} \rightarrow v^\text{M}_\text{r}})
\\
\mathbf{e}^\text{M2G}_{v^\text{M}_\text{s} \rightarrow v^\Gsuper_\text{r}} & = \text{MLP}^\text{embedder}_{\mathcal{E}^\text{M2G}}(\mathbf{e}^\text{M2G,features}_{v^\text{M}_\text{s} \rightarrow v^\Gsuper_\text{r}})
\end{split}
\end{equation}

\paragraph{Grid2Mesh GNN}\label{sec:app:grid2mesh}%
Next, in order to transfer information of the state of atmosphere from the grid nodes to the mesh nodes, we perform a single message passing step over the Grid2Mesh bipartite subgraph $\mathcal{G}_{\text{G2M}} (\mathcal{V}^\Gsuper, \mathcal{V}^\text{M}, \mathcal{E}^\text{G2M})$ connecting grid nodes to mesh nodes.
This update is performed using an interaction network \cite{battaglia2016interaction, battaglia2018relational}, augmented to be able to work with multiple node types \cite{allen2022learning}.
First, each of the Grid2Mesh edges are updated using information from the adjacent nodes,
\begin{equation}
{\mathbf{e}^\text{G2M}_{v^\Gsuper_\text{s} \rightarrow v^M_\text{r}}}' = \text{MLP}^\text{Grid2Mesh}_{\mathcal{E}^\text{G2M}}([
\mathbf{e}^\text{G2M}_{v^\Gsuper_\text{s} \rightarrow v^M_\text{r}},
\mathbf{v}^\Gsuper_s, 
\mathbf{v}^M_r]).
\end{equation}
Then each of the mesh nodes is updated by aggregating information from all of the edges arriving at that mesh node:
\begin{equation}
 {\mathbf{v}^\text{M}_i}' =  \text{MLP}^\text{Grid2Mesh}_{\mathcal{V}^\text{M}}\big(\big[
 \mathbf{v}^\text{M}_i, 
 \sum_{e^\text{G2M}_{v^\Gsuper_\text{s} \rightarrow v^\text{M}_\text{r}} : \; v^\text{M}_\text{r} = v^\text{M}_i} {\mathbf{e}^\text{G2M}_{v^\Gsuper_\text{s} \rightarrow v^\text{M}_\text{r}}}'\big]\big).
\end{equation}

Each of the grid nodes are also updated, but with no aggregation, because grid nodes are not receivers of any edges in the Grid2Mesh subgraph,
\begin{equation}
 {\mathbf{v}^\Gsuper_i}' =  \text{MLP}^\text{Grid2Mesh}_{\mathcal{V}^\Gsuper}\big(
 \mathbf{v}^\Gsuper_i\big).
\end{equation}
After updating all three elements, the model includes a residual connection, and for simplicity of the notation, reassigns the variables,
\begin{equation}
\begin{split}
 \mathbf{v}^\Gsuper_i & \leftarrow \mathbf{v}^\Gsuper_i + {\mathbf{v}^\Gsuper_i}',
\\
 \mathbf{v}^\text{M}_i & \leftarrow \mathbf{v}^\text{M}_i + {\mathbf{v}^\text{M}_i}',
\\
 \mathbf{e}^\text{G2M}_{v^\Gsuper_\text{s} \rightarrow v^M_\text{r}} & \leftarrow \mathbf{e}^\text{G2M}_{v^\Gsuper_\text{s} \rightarrow v^M_\text{r}} + {\mathbf{e}^\text{G2M}_{v^\Gsuper_\text{s} \rightarrow v^M_\text{r}}}'.
\end{split}
\end{equation}

\subsection{Processor}\label{sec:app:processor}
The processor is a deep GNN that operates on the Mesh subgraph $\mathcal{G}_{\text{M}} (\mathcal{V}^\text{M}, \mathcal{E}^\text{M})$ which only contains the Mesh nodes and and the Mesh edges. Note the Mesh edges contain the full multi-mesh, with not only the edges of $M^6$, but all of the edges of $M^5$, $M^4$, $M^3$, $M^2$, $M^1$ and $M^0$, which will enable long distance communication.

\paragraph{Multi-mesh GNN}\label{sec:app:meshgnn}
A single layer of the Mesh GNN is a standard interaction network \cite{battaglia2016interaction,battaglia2018relational} which first updates each of the mesh edges using information of the adjacent nodes:
\begin{equation}
{\mathbf{e}^\text{M}_{v^M_\text{s} \rightarrow v^M_\text{r}}}' = \text{MLP}^\text{Mesh}_{\mathcal{E}^\text{M}}([
\mathbf{e}^\text{M}_{v^M_\text{s} \rightarrow v^M_\text{r}},
\mathbf{v}^M_s, 
\mathbf{v}^M_r]).
\end{equation}
Then it updates each of the mesh nodes, aggregating information from all of the edges arriving at that mesh node:
\begin{equation}
 {\mathbf{v}^\text{M}_i}' =  \text{MLP}^\text{Mesh}_{\mathcal{V}^\text{M}}\big(\big[
 \mathbf{v}^\text{M}_i, 
 \sum_{e^\text{M}_{v^\text{M}_\text{s} \rightarrow v^\text{M}_\text{r}} : \; v^\text{M}_\text{r} = v^\text{M}_i} {\mathbf{e}^\text{M}_{v^\text{M}_\text{s} \rightarrow v^\text{M}_\text{r}}}'\big]\big)
\end{equation}
And after updating both, the representations are updated with a residual connection and for simplicity of the notation, also reassigned to the input variables:
\begin{equation}
\begin{split}
 \mathbf{v}^\text{M}_i & \leftarrow \mathbf{v}^\text{M}_i + {\mathbf{v}^\text{M}_i}'
\\
 \mathbf{e}^\text{M}_{v^M_\text{s} \rightarrow v^M_\text{r}} & \leftarrow \mathbf{e}^\text{M}_{v^N_\text{s} \rightarrow v^M_\text{r}} + {\mathbf{e}^\text{M}_{v^M_\text{s} \rightarrow v^M_\text{r}}}'
\end{split}
\end{equation}

The previous paragraph describes a single layer of message passing, but following a similar approach to \cite{sanchez2020learning, pfaff2021learning}, we applied this layer iteratively 16 times, using unshared neural network weights for the MLPs in each layer.

\subsection{Decoder}\label{sec:app:decoder}
The role of the decoder is to bring back information to the grid, and extract an output.

\paragraph{Mesh2Grid GNN}\label{sec:app:mesh2grid}%
Analogous to the Grid2Mesh GNN, the Mesh2Grid GNN performs a single message passing over the Mesh2Grid bipartite subgraph $\mathcal{G}_{\text{M2G}} (\mathcal{V}^\Gsuper, \mathcal{V}^\text{M}, \mathcal{E}^\text{M2G})$.
The Grid2Mesh GNN is functionally equivalent to the Mesh2Grid GNN, but using the Mesh2Grid edges to send information in the opposite direction.
The GNN first updates each of the Grid2Mesh edges using information of the adjacent nodes:
\begin{equation}
{\mathbf{e}^\text{M2G}_{v^M_\text{s} \rightarrow v^\Gsuper_\text{r}}}' = \text{MLP}^\text{Mesh2Grid}_{\mathcal{E}^\text{M2G}}([
\mathbf{e}^\text{M2G}_{v^M_\text{s} \rightarrow v^\Gsuper_\text{r}},
\mathbf{v}^M_s, 
\mathbf{v}^\Gsuper_r])
\end{equation}
Then it updates each of the grid nodes, aggregating information from all of the edges arriving at that grid node:
\begin{equation}
 {\mathbf{v}^\Gsuper_i}' =  \text{MLP}^\text{Mesh2Grid}_{\mathcal{V}^\Gsuper}\big(\big[
 \mathbf{v}^\Gsuper_i, 
 \sum_{e^\text{M2G}_{v^\text{M}_\text{s} \rightarrow v^\Gsuper_\text{r}} : \; v^\Gsuper_\text{r} = v^\Gsuper_i} {\mathbf{e}^\text{M2G}_{v^\text{M}_\text{s} \rightarrow v^\Gsuper_\text{r}}}'\big]\big).
\end{equation}
In this case we do not update the mesh nodes, as they won't play any role from this point on.

Here again we add a residual connection, and for simplicity of the notation, reassign the variables, this time only for the grid nodes, which are the only ones required from this point on:
\begin{equation}
 \mathbf{v}^\Gsuper_i \leftarrow \mathbf{v}^\Gsuper_i + {\mathbf{v}^\Gsuper_i}'.
\end{equation}

\paragraph{Output function}\label{sec:app:outputfunction}
Finally the prediction $\mathbf{\hat{y}}_i$ for each of the grid nodes is produced using another MLP,
\begin{equation}
 \mathbf{\ypred}^\Gsuper_i =  \text{MLP}^\text{Output}_{\mathcal{V}^\Gsuper}\big(
 \mathbf{v}^\Gsuper_i\big)
\end{equation}
which contains all 227 predicted variables for that grid node. Similar to \cite{sanchez2020learning, pfaff2021learning}, the next weather state, $\Xpred^{\ttt+1}$,  is computed by adding the per-node prediction, $\Ypred^{\ttt}$, to the input state for all grid nodes,
\begin{equation}
\Xpred^{\ttt+1} = \text{\ourmodel}(\X^{\ttt}, \X^{\ttt-1}) = \X^{\ttt} + \Ypred^{\ttt}.
\label{eq:output}
\end{equation}

\subsection{Normalization and network parameterization}\label{sec:app:othermodeldetails}

\paragraph{Input normalization}\label{sec:app:inputnormalization}
Similar to \cite{sanchez2020learning,pfaff2021learning}, we normalized all inputs. For each physical variable, we computed the per-pressure level mean and standard deviation over 1979--2015, and used that to normalize them to zero mean and unit variance.
For relative edge distances and lengths, we normalized the features to the length of the longest edge. For simplicity, we omit this output normalization from the notation.

\paragraph{Output normalization}\label{sec:app:outputnormalization}
Because our model outputs a difference, $\Ypred^{\ttt}$, which, during inference, is added to $\X^{\ttt}$ to produce $\Xpred^{\ttt+1}$, we normalized the output of the model by computing per-pressure level standard deviation statistics for the time difference $\Y^{\ttt} = \X^{\ttt+1} - \X^{\ttt}$ of each variable\footnote{We ignore the mean in the output normalization, as the mean of the time differences is zero.}.
When the GNN produces an output, we multiply this output by this standard deviation to obtain $\Ypred^{\ttt}$ before computing $\Xpred^{\ttt+1}$, as in \cref{eq:output}. For simplicity, we omit this output normalization from the notation.

\paragraph{Neural network parameterizations}\label{sec:app:neuralnets}
The neural networks within \ourmodel{} are all MLPs, with one hidden layer, and hidden and output layers sizes of 512 (except the final layer of the Decoder's MLP, whose output size is 227, matching the number of predicted variables for each grid node). We chose the ``swish''~\cite{ramachandran2017searching} activation function for all MLPs.
All MLPs are followed by a LayerNorm \cite{ba2016layer} layer (except for the Decoder's MLP).

\newpage
\section{Training details}\label{sec:app:trainingdetails}

This section provides details pertaining to the training of \ourmodel, including the data split used to develop the model (\cref{sec:app:trainingsplit}), the full definition of the objective function with the weight associated with each variable and vertical level (\cref{sec:app:trainingobjective}), the autoregressive training approach (\cref{sec:app:artraining}), optimization settings (\cref{sec:app:optimization}), curriculum training used to reduce training cost (\cref{sec:app:trainingschedule}), technical details used to reduce the memory footprint of \ourmodel (\cref{sec:app:memoryfootprint}), training time (\cref{sec:app:trainingtime}) and the software stacked we used (\cref{sec:app:softwarehardwarestack}).

\subsection{Training split}\label{sec:app:trainingsplit}

To mimic real deployment conditions, in which the forecast cannot depend on information from the future, we split the data used to develop \ourmodel and data used to test its performance ``causally'', in that the ``development set'' only contained dates earlier than those in the ``test set''.
The development set comprises the period 1979--2017, and the test set contains the years 2018--2021. 
Neither the researchers, nor the model training software, were allowed to view data from the test set until we had finished the development phase.
This prevented our choices of model architecture and training protocol from being able to exploit any information from the future.

Within our development set, we further split the data into a training set comprising the years 1979--2015, and a validation set that includes 2016--2017. We used the training set as training data for our models and the validation set for hyperparameter optimization and model selection, i.e., to decide on the best-performing model architecture.
We then froze the model architecture and all the training choices and moved to the test phase. In preliminary work, we also explored training on earlier data from 1959--1978, but found it had little benefit on performance, so in the final phases of our work we excluded 1959--1978 for simplicity.

\subsection{Training objective}\label{sec:app:trainingobjective}
\ourmodel{} was trained to minimize an objective function over 12-step forecasts (3 days) against ERA5 targets, using gradient descent.
The training objective is defined as the mean square error (MSE) between the target output $\X$ and predicted output $\Xpred$,
\begin{align}
  \mathcal{L}_{\text{MSE}} &= 
  \underbrace{\frac{1}{|D_{\text{batch}}|}\sum_{\dinit \in D_{\text{batch}}}}_{\text{forecast date-time}}
  \underbrace{\frac{1}{\T_{\text{train}}}\sum_{\lt \in 1:\T_{\text{train}}}}_{\text{lead time}}
  \underbrace{\frac{1}{|\Gquarterdegree|}\sum_{\sll \in \Gquarterdegree}}_{\text{spatial location}}
  \underbrace{\sum_{j \in J}}_{\text{variable-level}}
  \stdweight_{j} \varweight_j 
  \latitudeweight_{\sll} 
  \underbrace{(\xpred^{\dinit+\lt}_{\sll,j} - \x^{\dinit+\lt}_{\sll, j})^2}_{\text{squared error}}
\end{align}
where
\begin{itemize}[nosep]
    \item $\lt \in 1:T_{\text{train}}$ are the lead times that correspond to the $T_{\text{train}}$ autoregressive steps. 
    \item $\dinit \in D_{\text{batch}}$ represent forecast initialization date-times in a batch of forecasts in the training set, 
    \item $j \in J$ indexes the variable, and for atmospheric variables the pressure level. E.g.
    $J=\{\varlevel{z}{1000}, \varlevel{z}{850}, \dots, \varlevel{2t}, \varlevel{msl}\}$,
    \item $\sll \in \Gquarterdegree$ are the location (latitude and longitude) coordinates in the grid,
    \item $\xpred^{\dinit+\lt}_{j, \sll}$ and $\x^{\dinit+\lt}_{j, \sll}$ are predicted and target values for some variable-level, location, and lead time,
    \item $\stdweight_j$ is the per-variable-level inverse variance of time differences,
    \item $\varweight_j$ is the per-variable-level loss weight,
    \item $\latitudeweight_{\sll}$ is the area of the latitude-longitude grid cell, which varies with latitude, and is normalized to unit mean over the grid.
\end{itemize}

In order to build a single scalar loss, we took the average across latitude-longitude, pressure levels, variables, lead times, and batch size. We averaged across latitude-longitude axes, with a weight proportional to the latitude-longitude cell size (normalized to mean 1). We applied uniform averages across time and batch.

\begin{figure}[t]
  \centering
  \includegraphics[width=\textwidth]{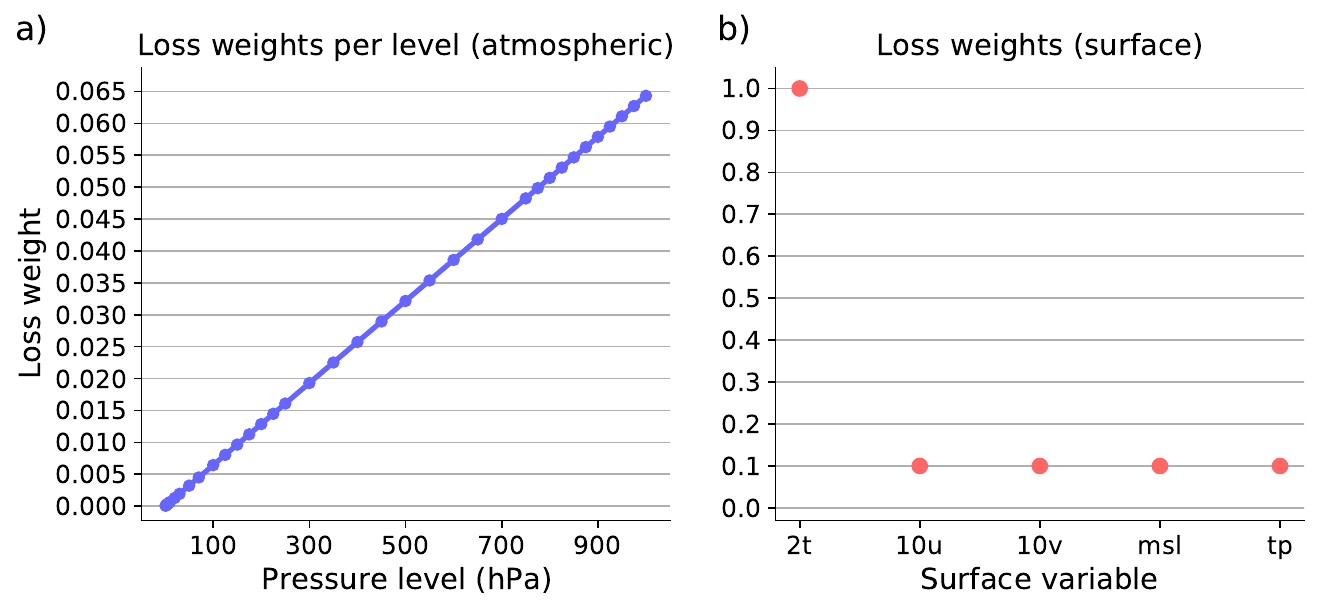}
  \caption{\small\textbf{Training loss weights.} (a) Loss weights per pressure level, for atmospheric variables. (b) Loss weights for surface variables.}
  \label{fig:app:lossweights}
\end{figure}

The quantities $\stdweight_{j} = \mathbb{V}_{i,t}\left[x_{i,j}^{t+1}-x_{i,j}^{t}\right]^{-1}$ are per-variable-level inverse variance estimates of the time differences, which aim to standardize the targets (over consecutive steps) to unit variance. These were estimated from the training data. We then applied per-variable-level loss weights, $\varweight_j$. For atmospheric variables, we averaged across levels, with a weight proportional to the pressure of the level (normalized to unit mean), as shown in \cref{fig:app:lossweights}a. We use pressure here as a proxy for the density~\cite{keisler2022forecasting}.
Note that the loss weight applied to pressure levels at or below 50 \unit{hPa}, where HRES tends to perform better than \ourmodel{}, is only $0.66\%$ of the total loss weight across all variables and levels.
We tuned the loss weights for the surface variables during model development, so as to produce roughly comparable validation performance across all variables: the weight on \varlevel{2t} was 1.0, and the weights on \varlevel{10u}, \varlevel{10v}, \varlevel{msl}, and \varlevel{tp} were each 0.1, as shown in \cref{fig:app:lossweights}b. 
The loss weights across all variables sum to $7.4$, i.e., ($6 \times 1.0$ for the atmospheric variables, plus ($1.0 + 0.1 + 0.1 + 0.1 + 0.1$) for the surface variables listed above, respectively).

\subsection{Training on autoregressive objective}\label{sec:app:artraining}
In order to improve our model's ability to make accurate forecasts over more than one step, we used an autoregressive training regime, where the model's predicted next step was fed back in as input for predicting the next step. The final \ourmodel{} version was trained on 12 autoregressive steps, following a curriculum training schedule described below. The optimization procedure computed the loss on each step of the forecast, with respect to the corresponding ground truth step, error gradients with respect to the model parameters were backpropagated through the full unrolled sequence of model iterations (i.e., using backpropagation-through-time).

\subsection{Optimization}\label{sec:app:optimization}
The training objective function was minimized using gradient descent, with mini-batches. We sampled ground truth trajectories from our ERA5 training dataset, with replacement, for batches of size 32. We used the AdamW optimizer \cite{loshchilov2017decoupled,kingma2014adam} with parameters $(\text{beta1}=0.9, \text{beta2}=0.95)$. We used weight decay of $0.1$ on the weight matrices. We used gradient (norm) clipping with a maximum norm value of 32.

\subsection{Curriculum training schedule}\label{sec:app:trainingschedule}
\begin{figure}[t]
  \centering
  \includegraphics[width=\textwidth]{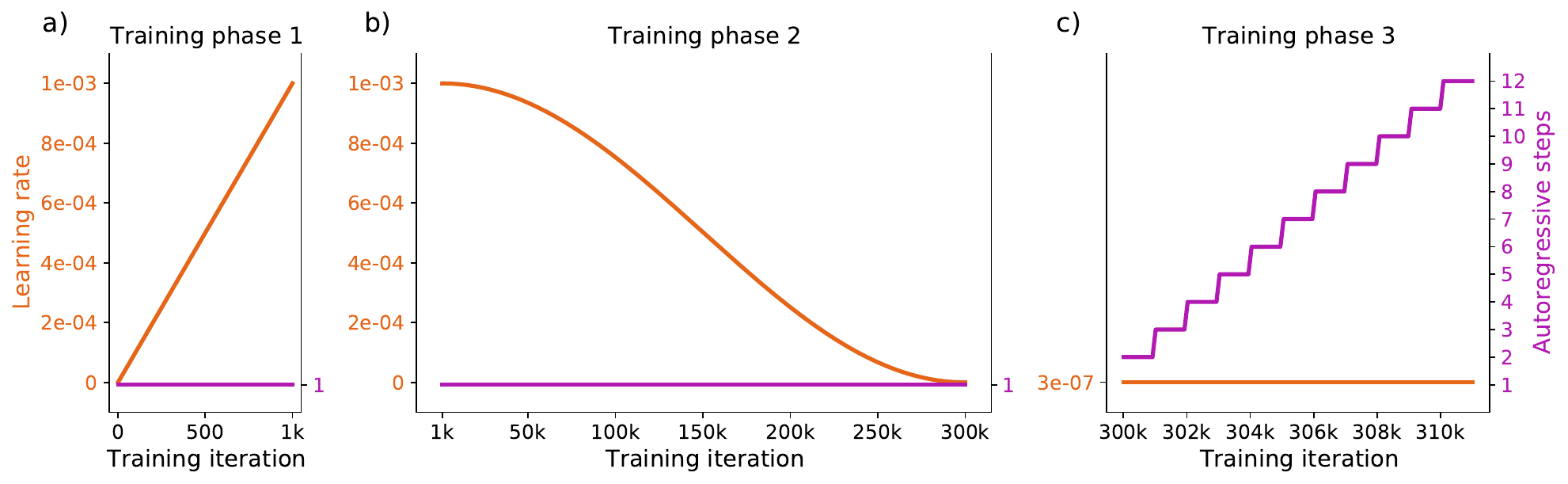}
  \caption{\small\textbf{Training schedule.} (a) First phase of training. (b) Second phase of training. (c) Third phase of training.}
  \label{fig:app:trainingschedule}
\end{figure}
Training the model was conducted using a curriculum of three phases, which varied the learning rates and number of autoregressive steps. The first phase consisted of 1000 gradient descent updates, with one autoregressive step, and a learning rate schedule that increased linearly from \enotation{0} to \enotation{1}{-3} (\cref{fig:app:trainingschedule}a). The second phase consisted of 299,000 gradient descent updates, again with one autoregressive step, and a learning rate schedule that decreased back to \enotation{0} with half-cosine decay function (\cref{fig:app:trainingschedule}b). The third phase consisted of 11,000 gradient descent updates, where the number of autoregressive steps increased from 2 to 12, increasing by 1 every 1000 updates, and with a fixed learning rate of \enotation{3}{-7} (\cref{fig:app:trainingschedule}c).

\subsection{Reducing memory footprint}\label{sec:app:memoryfootprint}
To fit long trajectories (12 autoregressive steps) into the 32GB of a Cloud TPU v4 device, we use several strategies to reduce the memory footprint of our model.
First, we use batch parallelism to distribute data across 32 TPU devices (i.e., one data point per device).
Second, we use bfloat16 floating point precision to decrease the memory taken by activations (note, we use full-precision numerics (i.e. float32) to compute performance metrics at evaluation time).
Finally, we use gradient check-pointing \cite{chen2016training} to further reduce memory footprint at the cost of a lower training speed.

\subsection{Training time}\label{sec:app:trainingtime}
Following the training schedule that ramps up the number of autoregressive steps, as detailed above, training \ourmodel{} took about four weeks on 32 TPU devices.

\subsection{Software and hardware stack}\label{sec:app:softwarehardwarestack}
We use JAX \cite{jax2018github}, Haiku \cite{haiku2020github}, Jraph \cite{jraph2020github}, Optax, Jaxline \cite{deepmind2020jax} and xarray \cite{hoyer2017xarray} to build and train our models.

\newpage
\section{Verification methods}\label{sec:app:evaluationdetails}

This section provides details on our evaluation protocol. \cref{sec:app:trainvaltest} details our approach to splitting data in a causal way, ensuring our evaluation tests for meaningful generalization, i.e., without leveraging information from the future. 
\cref{sec:app:comparingGraphcastHRES} explains in further details our choices to evaluate HRES skill and compare it to \ourmodel, starting from the need for a ground truth specific to HRES to avoid penalizing it at short lead times (\cref{sec:app:choice_of_ground_truth}),
the impact of ERA5 and HRES using different assimilation windows on the lookahead each state incorporates (\cref{sec:app:ensuring_equal_lookahead}), the resulting choice of initialization time for \ourmodel and HRES to ensure that all methods benefit from the same lookahead in their inputs as well as in their targets (\cref{sec:app:alignment_of_times_of_day}), and finally the evaluation period we used to report performance on 2018 (\cref{sec:app:evalperiod}). 
\cref{sec:app:evaluationmetrics} provides the definition of the metrics used to measure skill in our main results, as well as metrics used in complementary results in the Supplements.
Finally, \cref{sec:app:statistical_testing} details our statistical testing methodology.

\subsection{Training, validation, and test splits}\label{sec:app:trainvaltest}

\begin{figure}
  \centering
  \includegraphics[width=\textwidth]{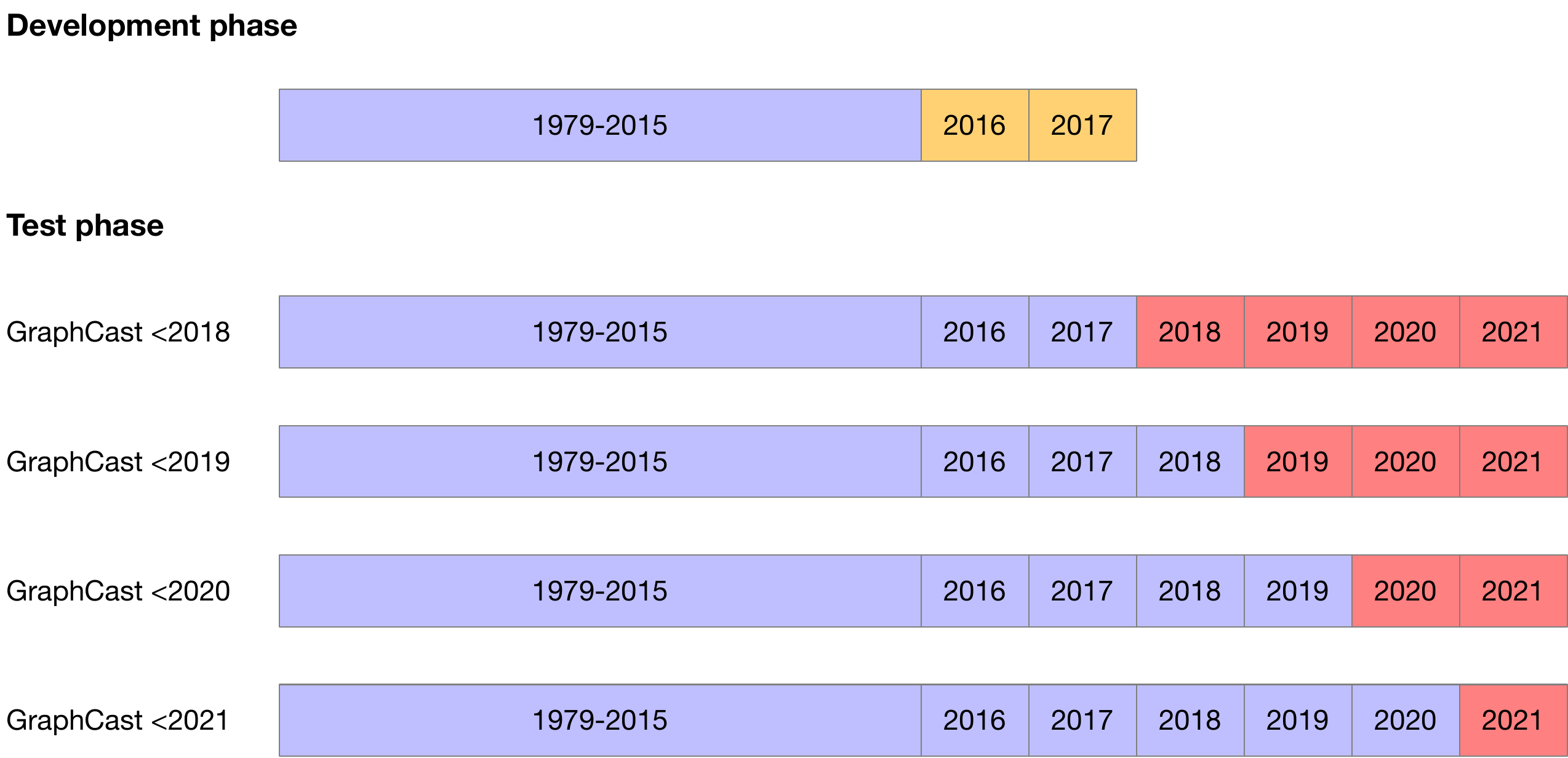}
  \caption{\small\textbf{Data split summary.} In the development phase,  \ourmodel was trained on 1979--2015 (blue) and validated on 2016--2017 (yellow) until the training protocol was frozen. In the test phase, four versions of \ourmodel were trained on larger and more recent train sets. Blue years represent training years for a given version of \ourmodel, and red years represent the data that can be used at test time while satisfying split causality.}
  \label{fig:app:splits}
\end{figure}

In the test phase, using protocol frozen at the end of the development phase (\cref{sec:app:trainingsplit}), we trained four versions of \ourmodel, each of them on a different period.
The models were trained on data from 1979--2017, 1979--2018, 1979--2019 and 1979--2020 for evaluation on the periods 2018--2021, 2019--2021, 2020--2021 and 2021, respectively. Again, these splits maintained a causal separation between the data used to train a version of the model and the data used to evaluate its performance (see \cref{fig:app:splits}).
Most of our results were evaluated on 2018 (i.e., with the model trained on 1979--2017), with several exceptions.
For cyclone tracking experiments, we report results on 2018--2021 because cyclones are not that common, so including more years increases the sample size. We use the most recent version of \ourmodel to make forecast on a given year: \ourmodel $<$2018 for 2018 forecast, \ourmodel $<$2019 for 2019 forecast, etc.  For training data recency experiments, we evaluated how different models trained up to different years compared on 2021 test performance.

\subsection{Comparing \ourmodel{} to HRES}\label{sec:app:comparingGraphcastHRES}

\subsubsection{Choice of ground truth datasets}
\label{sec:app:choice_of_ground_truth}

\ourmodel was trained to predict ERA5 data, and to take ERA5 data as input; we also use ERA5 as ground truth for evaluating our model. HRES forecasts, however, are initialized based on HRES analysis. Generally, verifying a model against its own analysis gives the best skill estimates~\cite{swinbank2016tigge}.
So rather than evaluating HRES forecasts against ERA5 ground truth, which would mean that even the zeroth step of HRES forecasts would have non-zero error, we constructed an ``HRES forecast at step 0'' (\hresfczero{}) dataset, which contains the initial time step of HRES forecasts at future initializations (see \cref{sec:app:hresfczero}). We use \hresfczero{} as ground truth for evaluating HRES forecasts.

\subsubsection{Ensuring equal lookahead in assimilation windows}
\label{sec:app:ensuring_equal_lookahead}

\begin{figure}[ht]
  \centering
  \includegraphics[width=0.8\textwidth]{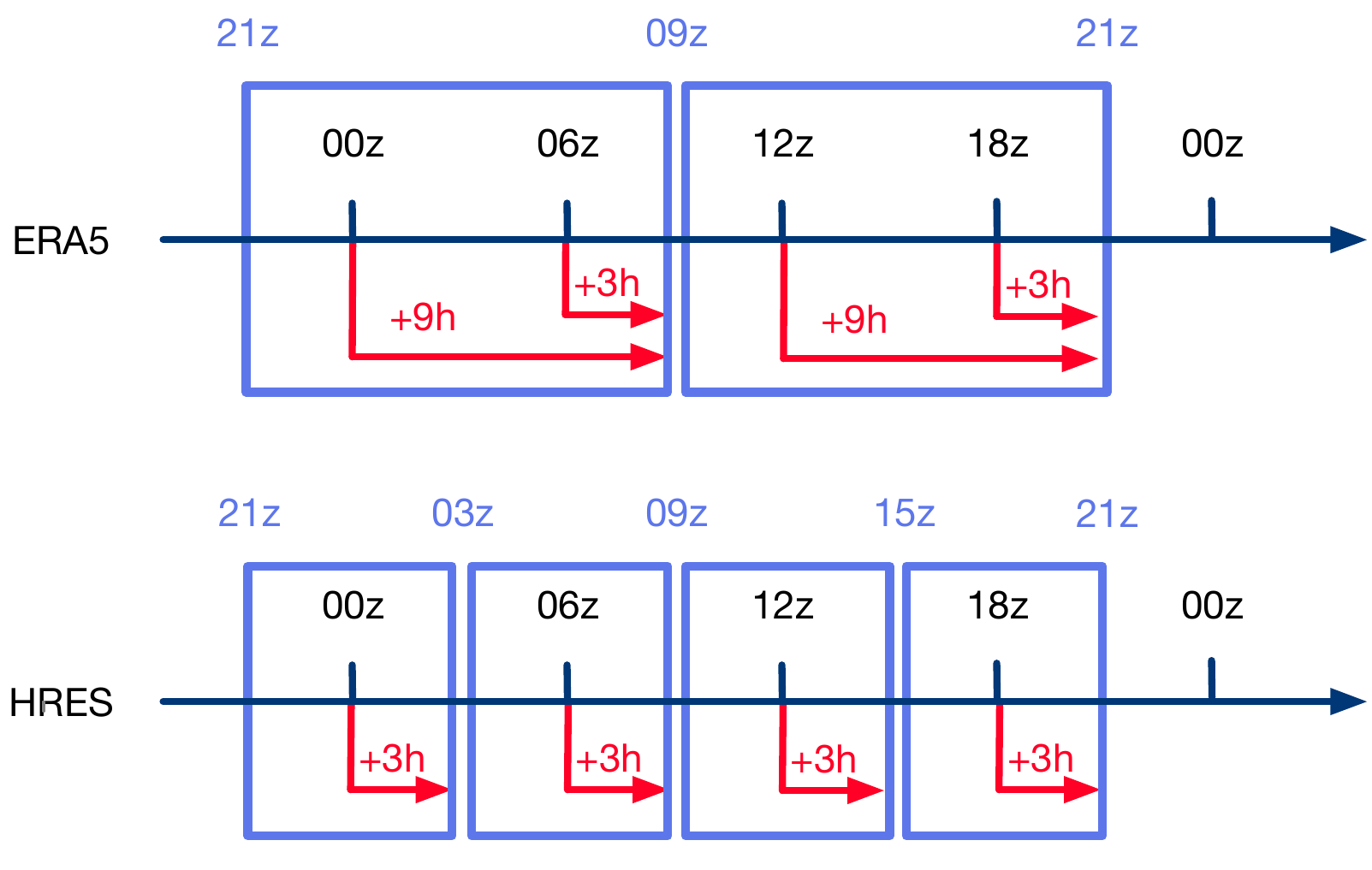}
  \caption{\small\textbf{Schematic of the assimilation windows for ERA5 and HRES.} Data assimilation windows are marked as blue rectangles spanning 12h for ERA5 and 6h for HRES. The red arrows represent the duration of the effective lookahead that is incorporated in the corresponding state.}
  \label{fig:app:assimilation_window_schematic}
\end{figure}

\begin{figure}[ht]
  \centering
  \includegraphics[width=0.85\textwidth]{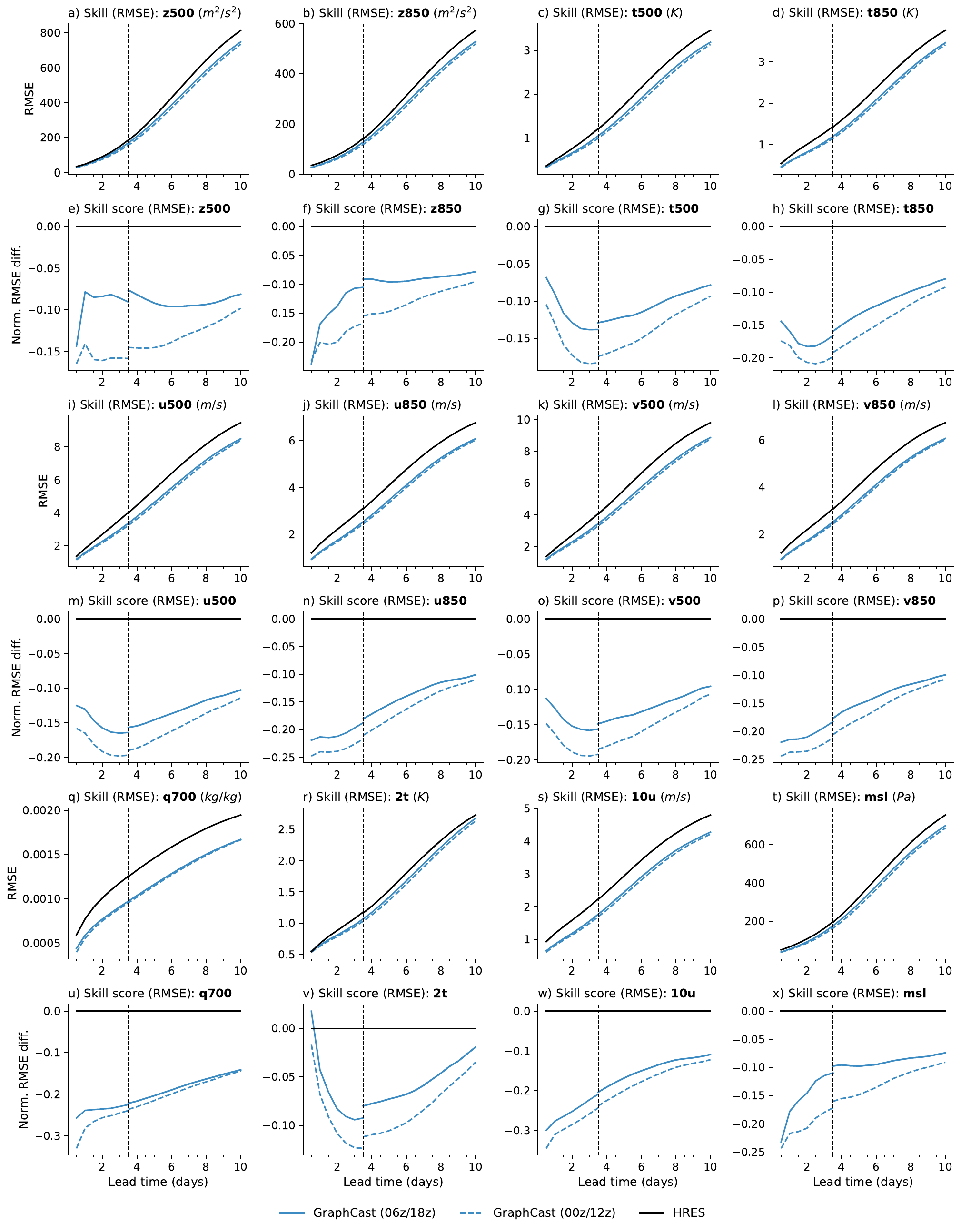}
  \caption{\small\textbf{Effect of initialization time.} When initialized from an ERA5 state benefiting from longer lookahead (00z/12z), \ourmodel performs better than when initialized from an ERA5 state benefiting from a shorter lookahead (06z/18z). The improvement is measurable from short to long prediction lead time. This supports our choice to evaluate all models from 06z/18z, in order to avoid giving an advantage to \ourmodel. The x-axis represents lead time, at 12-hour steps over 10 days. The y-axis represents the RMSE skill or skill score.}
  \label{fig:app:appendix_gc0012_vs_gc0618}
\end{figure}

When comparing the skills of \ourmodel and HRES, we made several choices to control for differences between the ERA5 and \hresfczero data assimilation windows.
As described in~\cref{sec:app:datasets}, each day HRES assimilates observations using four +/-3h windows centered on 00z, 06z, 12z and 18z (where 18z means 18:00 UTC in Zulu convention), while ERA5 uses two +9h/-3h windows centered on 00z and 12z, or equivalently two +3h/-9h windows centered on 06z and 18z.
See \cref{fig:app:assimilation_window_schematic} for an illustration.
We chose to evaluate \ourmodel's forecasts from the 06z and 18z initializations, ensuring its inputs carry information from +3h of future observations, matching HRES's inputs. We did not evaluate \ourmodel's 00z and 12z initializations, to avoid a mismatch between having a +9h lookahead in ERA5 inputs versus +3h lookahead for HRES inputs.
\cref{fig:app:appendix_gc0012_vs_gc0618} show the performance of \ourmodel initialized from 06z/18z, and 00z/12z. When initialized from a state with a larger lookahead, \ourmodel gets a visible improvement that persists at longer lead times, supporting our choice to initialized evaluation from 06z/18z. 
We applied the same logic when choosing the target on which to evaluate: we only evaluate targets which incorporate a 3h lookahead for both HRES and ERA5. Given our choice of initialization at 06z and 18z, this corresponds to evaluating every 12h, on future 06z and 18z analysis times.
As a practical example, if we were to evaluate \ourmodel and HRES initialized at 06z, at lead time 6h (i.e., 12z), the target for \ourmodel would integrate a +9h lookahead, while the target for HRES would only incorporate +3h lookahead. At equal lead time, this could result in a harder task for \ourmodel.

\subsubsection{Alignment of initialization and validity times-of-day}
\label{sec:app:alignment_of_times_of_day}

As stated above, a fair comparison with HRES requires us to evaluate \ourmodel using 06z and 18z initializations, and with lead times which are multiples of 12\unit{h}, meaning validity times are also 06z and 18z.

For lead times up to 3.75 days there are archived HRES forecasts available using 06z and 18z initialization and validity times, and we use these to perform a like-for-like comparison with \ourmodel at these lead times. Note, because we evaluate only on 12 hour lead time increments, this means the final lead time is 3.5 days.

For lead times of 4 days and beyond, archived HRES forecasts are only available at 00z and 12z initializations, which given our 12-hour-multiple lead times means 00z and 12z validity times. At these lead times we have no choice but to compare \ourmodel at 06z and 18z, with HRES at 00z and 12z.

\begin{figure}[ht]
  \centering
  \includegraphics[width=0.99\textwidth]{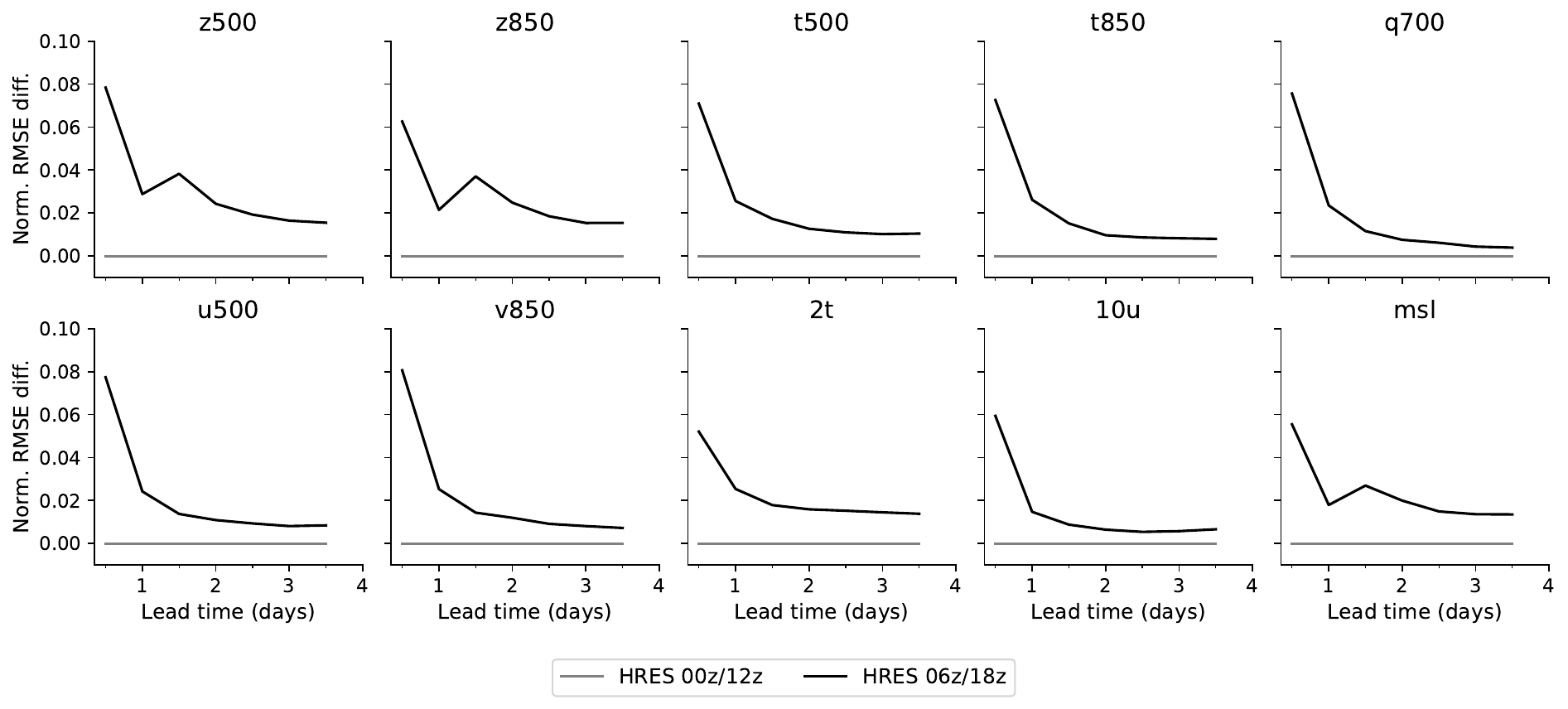}
  \caption{\small\textbf{RMSE skill scores for HRES at 06z/18z vs HRES at 00z/12z.} Plots show the RMSE skill score of HRES initialized at 06z/18z (black line), relative to HRES initialized at 00z/12z (grey line), as a function of lead time. We plot lead times up to 3.5 days which is the longest for which HRES predictions initialized at 06z/18z are available. The y axis scales are shared.}
  \label{fig:app:rmse_skill_score_hres_0012_0618}
\end{figure}

In these comparisons of globally-defined RMSEs, we expect the difference in time-of-day to give HRES a slight advantage. In \cref{fig:app:rmse_skill_score_hres_0012_0618}, we can see that up to 3.5 day lead times, HRES RMSEs tend to be smaller on average over 00z and 12z initialization/validity times than they are at the 06z and 18z times which \ourmodel is evaluated on. We can also see that the difference decreases as lead time increases, and that the 06z/18z RMSEs generally appear to be tending towards an asymptote above the 00z/12z RMSE, but within 2\% of it. We expect these differences to continue to favor HRES at longer lead times, and regardless to remain small, and so we do not believe that they compromise our conclusions in cases where \ourmodel has greater skill than HRES.

Whenever we plot RMSE and other evaluation metrics as a function of lead time, we indicate with a dotted line the 3.5 day changeover point where we switch from evaluating HRES on 06z/18z to evaluating on 00z/12z. At this changeover point, we plot both the 06z/18z and 00z/12z metrics, showing the discontinuity clearly.

\FloatBarrier

\subsubsection{Evaluation period}\label{sec:app:evalperiod}

Most of our main results are reported for the year 2018 (from our test set), for which the first forecast initialization time was \datetime{2018-01-01}{06:00:00} UTC and the last \datetime{2018-12-31}{18:00:00}, or when evaluating HRES at longer lead times, \datetime{2018-01-01}{00:00:00} and \datetime{2018-12-31}{12:00:00}.
Additional results on cyclone tracking and the effect of data recency use years 2018--2021 and 2021 respectively.

\subsection{Evaluation metrics}\label{sec:app:evaluationmetrics}

We quantify the skillfulness of \ourmodel{}, other ML models, and HRES using the root mean square error~(RMSE) and the anomaly correlation coefficient~(ACC), which are both computed against the models' respective ground truth data. The RMSE measures the magnitude of the differences between forecasts and ground truth for a given variable indexed by $j$ and a given lead time $\tau$ (see \cref{eq:app:rmse}). The ACC, $\mathcal{L}_{\text{ACC}}^{j,\lt}$, is defined in \cref{eq:app:acc} and measures how well forecasts' differences from climatology, i.e., the average weather for a location and date, correlate with the ground truth's differences from climatology. For skill scores we use the normalized RMSE difference between model $A$ and baseline $B$ as $(\mathrm{RMSE}_A - \mathrm{RMSE}_B) / \mathrm{RMSE}_B$, and the normalized ACC difference as $(\mathrm{ACC}_A - \mathrm{ACC}_B) / (1 - \mathrm{ACC}_B)$.

All metrics were computed using float32 precision and reported using the native dynamic range of the variables, without normalization.

\paragraph{Root mean square error (RMSE).}\label{sec:app:rmse}
We quantified forecast skill for a given variable, $\x_j$, and lead time, $\lt = \ttt\dsize$, using a latitude-weighted root mean square error (RMSE) given by
\begin{align}
  \text{RMSE}(j,\lt) &= 
  \frac{1}{|D_{\text{eval}}|} \sum_{\dinit \in D_{\text{eval}}}
  \sqrt{
  \frac{1}{|\Gquarterdegree|} \sum_{\sll \in \Gquarterdegree} 
  \latitudeweight_{\sll} {\left(\xpred^{\dinit+\lt}_{j,\sll} - \x^{\dinit+\lt}_{j,\sll}\right)}^2
  }
\label{eq:app:rmse}
\end{align}
where
\begin{itemize}[nosep]
    \item $\dinit \in D_{\text{eval}}$ represent forecast initialization date-times in the evaluation dataset, 
    \item $j \in J$ index variables and levels, e.g., $J=\{\varlevel{z}{1000}, \varlevel{z}{850}, \dots, \varlevel{2t}, \varlevel{msl}\}$,
    \item $\sll \in \Gquarterdegree$ are the location (latitude and longitude) coordinates in the grid,
    \item $\xpred^{\dinit+\lt}_{j,\sll}$ and $\x^{\dinit+\lt}_{j,\sll}$ are predicted and target values for some variable-level, location, and lead time,
    \item $\latitudeweight_{\sll}$ is the area of the latitude-longitude grid cell (normalized to unit mean over the grid) which varies with latitude.
\end{itemize}

By taking the square root inside the mean over forecast initializations we follow the convention of WeatherBench \cite{rasp2020weatherbench}. However we note that this differs from how RMSE is defined in many other contexts, where the square root is only applied to the final mean, that is,
\begin{align}
  \text{RMSE}_{\text{trad}}(j,\lt) &=  \sqrt{
  \frac{1}{|D_{\text{eval}}|} \sum_{\dinit \in D_{\text{eval}}}
  \frac{1}{|\Gquarterdegree|} \sum_{\sll \in \Gquarterdegree} 
  \latitudeweight_{\sll} {\left(\xpred^{\dinit+\lt}_{j,\sll} - \x^{\dinit+\lt}_{j,\sll}\right)}^2
  }
\label{eq:app:rmse_trad_grid}.
\end{align}

\paragraph{Root mean square error (RMSE), spherical harmonic domain.}%

In all comparisons involving predictions that are filtered, truncated or decomposed in the spherical harmonic domain, for convenience we compute RMSEs directly in the spherical harmonic domain, with all means taken inside the square root,
\begin{align}
  \text{RMSE}_{\text{sh}}(j,\lt) &=  \sqrt{
  \frac{1}{|D_{\text{eval}}|} \sum_{\dinit \in D_{\text{eval}}}
  \frac{1}{4\pi}
  \sum_{l=0}^{l_{max}} \sum_{m=-l}^{l}
  {\left(\hat{f}^{\dinit+\lt}_{j,l,m} - f^{\dinit+\lt}_{j,l,m}\right)}^2
  }
\label{eq:app:rmse_sh}
\end{align}
Here $\hat{f}^{\dinit+\lt}_{j,l,m}$ and $f^{\dinit+\lt}_{j,l,m}$ are predicted and target coefficients of spherical harmonics with total wavenumber $l$ and longitudinal wavenumber $m$. We compute these coefficients from grid-based data using a discrete spherical harmonic transform \cite{driscoll1994-discrete-sht} with triangular truncation at wavenumber 719, which was chosen to resolve the \quarterdegree{} (28km) resolution of our grid at the equator. This means that $l$ ranges from 0 to $l_{max} = 719$ and $m$ from $-l$ to $l$.

This RMSE closely approximates the grid-based definition of RMSE given in~\cref{eq:app:rmse_trad_grid}, however it is not exactly comparable, in part because the triangular truncation at wavenumber 719 does not resolve the additional resolution of the equiangular grid near the poles.

\paragraph{Root mean square error (RMSE), per location.}\label{sec:app:rmse_per_location}

This is computed following the RMSE definition of \cref{eq:app:rmse_trad_grid}, but for a single location:
\begin{align}
  \text{RMSE}_{\text{by-lat-lon}}(\sll,j,\lt) &=  \sqrt{
    \frac{1}{|D_{\text{eval}}|}
    \sum_{\dinit \in D_{\text{eval}}} 
    \left(\xpred^{\dinit+\lt}_{j,\sll} - \x^{\dinit+\lt}_{j,\sll}\right)^2
  }.
\label{eq:app:rmse_lat_lon}
\end{align}
We also break down RMSE by latitude only:
\begin{align}
  \text{RMSE}_{\text{by-lat}}(l,j,\lt) &=  \sqrt{
      \frac{1}{|D_{\text{eval}}|}
      \sum_{\dinit \in D_{\text{eval}}}
      \frac{1}{|\text{lon}(\Gquarterdegree)|}
         \sum_{\sll \in \Gquarterdegree : \text{lat}(\sll) = l}
        {(\xpred^{\dinit+\lt}_{j,\sll} - \x^{\dinit+\lt}_{j,\sll})}^2
  }
\label{eq:app:rmse_above_ground_lat}
\end{align}
where $|\text{lon}(\Gquarterdegree)| = 1440$ is the number of distinct longitudes in our regular \quarterdegree grid.

\paragraph{Root mean square error (RMSE), by surface elevation.}\label{sec:app:rmse_by_surface_elevation}

This is computed following the RMSE definition of \cref{eq:app:rmse_trad_grid} but restricted to a particular range of surface elevations, given by bounds $z_l \leq z_{\text{surface}} < z_u$ on the surface geopotential:
\begin{align}
  \text{RMSE}_{\text{by-elevation}}(z_l,z_u,j,\lt) &=  \sqrt{
      \frac{
      \sum_{\dinit \in D_{\text{eval}}}
         \sum_{\sll \in \Gquarterdegree} \mathbb{I}[z_l \leq z_{surface}(\sll) < z_u]
         \latitudeweight_{\sll}
         (\xpred^{\dinit+\lt}_{j,\sll} - \x^{\dinit+\lt}_{j,\sll})^2
      }{
      |D_{\text{eval}}|
         \sum_{\sll \in \Gquarterdegree} \mathbb{I}[z_l \leq z_{surface}(\sll) < z_u]
         \latitudeweight_{\sll}
      }
  },
\label{eq:app:rmse_by_surface_elevation}
\end{align}
where $\mathbb{I}$ denotes the indicator function.

\paragraph{Mean bias error (MBE), per location.}\label{sec:app:mbe}

This quantity is defined as
\begin{align}
  \text{MBE}_{\text{by-lat-lon}}(\sll,j,\lt) &= 
        \frac{1}{|D_{\text{eval}}|}
        \sum_{\dinit \in D_{\text{eval}}}
        \left( \xpred^{\dinit+\lt}_{j,\sll} - \x^{\dinit+\lt}_{j,\sll} \right).
\label{eq:app:mbe_lat_lon}
\end{align}

\paragraph{Root-mean-square per-location mean bias error (RMS-MBE).}\label{sec:app:rms_mbe}
This quantifies the average magnitude of the per-location biases from \cref{eq:app:mbe_lat_lon} and is given by
\begin{align}
  \text{RMS-MBE}(j,\lt) &=  \sqrt{
      \frac{1}{|\Gquarterdegree|} \sum_{\sll \in \Gquarterdegree} \latitudeweight_{\sll}\;
      \text{MBE}_{\text{by-lat-lon}}(\sll,j,\lt)^2
  }.
\label{eq:app:rms_mbe}
\end{align}

\paragraph{Correlation of per-location mean bias errors.}\label{sec:app:corr_mbe}
This quantifies the correlation between per-location biases (\cref{eq:app:mbe_lat_lon}) of two different models A and B. We use an uncentered correlation coefficient because of the significance of the origin zero in measurements of bias, and compute this quantity according to
\begin{align}
  \text{Corr-MBE}(j,\lt) &=  \frac{
      \frac{1}{|\Gquarterdegree|} \sum_{\sll \in \Gquarterdegree} \latitudeweight_{\sll}\;
      \text{MBE}_A(\sll,j,\lt) \text{MBE}_B(\sll,j,\lt)
    }{
      \text{RMS-MBE}_A(j,\lt) \text{RMS-MBE}_B(j,\lt)
    }.
\label{eq:app:corr_mbe}
\end{align}

\paragraph{Anomaly correlation coefficient (ACC).}\label{sec:app:acc}
We also computed the anomaly correlation coefficient for a given variable, $\x_j$,
and lead time, $\lt = \ttt\dsize$, according to
\begin{align}
\mathcal{L}_{\text{ACC}}^{j,\lt} &=
\frac{1}{|D_{\text{eval}}|} \sum_{\dinit \in D_{\text{eval}}}
\frac{
\sum_{\sll \in \Gquarterdegree}
\latitudeweight_{\sll}
\left(\xpred^{\dinit+\lt}_{j,\sll} - \climatology^{\dinit+\lt}_{j,\sll}\right)
\left(\x^{\dinit+\lt}_{j,\sll} - \climatology^{\dinit+\lt}_{j,\sll}\right)
}{
\sqrt{
\left[
\sum_{\sll \in \Gquarterdegree}
\latitudeweight_{\sll}
{\left(\xpred^{\dinit+\lt}_{j,\sll} - \climatology^{\dinit+\lt}_{j,\sll}\right)}^2
\right]
\left[
\sum_{\sll \in \Gquarterdegree}
\latitudeweight_{\sll}
{\left(\x^{\dinit+\lt}_{j,\sll} - \climatology^{\dinit+\lt}_{j,\sll}\right)}^2 
\right]
}
}
\label{eq:app:acc}
\end{align}
where $\climatology^{\dinit+\lt}_{j,\sll}$ is the climatological mean for a given variable, level, latitude and longitude, and for the day-of-year containing the validity time $\dinit+\lt$. Climatological means were computed using ERA5 data between 1993 and 2016. All other variables are defined as above.

\subsection{Statistical methodology}
\label{sec:app:statistical_testing}

\subsubsection{Significance tests for difference in means}
\label{sec:app:significance_tests}

For each lead time $\lt$ and variable-level $j$, we test for a difference in means between per-initialization-time RMSEs (defined in \cref{eq:app:rmse_per_init}) for \ourmodel and HRES. We use a paired two-sided $t$-test with correction for auto-correlation, following the methodology of \cite{geer2016significance}. This test assumes that time series of differences in forecast scores are adequately modelled as stationary Gaussian AR(2) processes. This assumption does not hold exactly for us, but is motivated as adequate for verification of medium range weather forecasts by the ECMWF in \cite{geer2016significance}.

The nominal sample size for our tests is $n=730$ at lead times under 4 days, consisting of two forecast initializations per day over the 365 days of 2018. (For lead times over 4 days we have $n=729$, see \cref{sec:app:stat_testing_forecast_alignment}). However these data (differences in forecast RMSEs) are auto-correlated in time. Following \cite{geer2016significance} we estimate an inflation factor $k$ for the standard error which corrects for this. Values of $k$ range between 1.21 and 6.75,
with the highest values generally seen at short lead times and at the lowest pressure levels. These correspond to reduced effective sample sizes $n_{\text{eff}} = n / k^2$ in the range of 16 to 501.

See \cref{tab:app:significance_test_details} for detailed results of our significance tests, including $p$-values, values of the $t$ test statistic and of $n_{\text{eff}}$.

\subsubsection{Forecast alignment}
\label{sec:app:stat_testing_forecast_alignment}

For lead times $\lt$ less than 4 days, we have forecasts available at 06z and 18z initialization and validity times each day for both \ourmodel and HRES, and we can test for differences in RMSEs between these paired forecasts. Defining the per-initialization-time RMSE as:
\begin{align}
  \mathrm{RMSE}(j, \lt, \dinit) &= \sqrt{
  \frac{1}{|\Gquarterdegree|} \sum_{\sll \in \Gquarterdegree} 
  \latitudeweight_{\sll} {\left(\xpred^{\dinit+\lt}_{j,\sll} - \x^{\dinit+\lt}_{j,\sll}\right)}^2
  }
\label{eq:app:rmse_per_init}
\end{align}
We compute differences
\begin{align}
\text{diff-RMSE}(j, \lt, \dinit) &=  \text{RMSE}_{GC}(j, \lt, \dinit) -  \text{RMSE}_{HRES}(j, \lt, \dinit),
\end{align}
which we use to test the null hypothesis that $\mathbb{E}[\text{diff-RMSE}(j, \lt, \dinit)] = 0$ against the two-sided alternative. Note that by our stationarity assumption this expectation does not depend on $\dinit$.

As discussed in \cref{sec:app:alignment_of_times_of_day}, at lead times of 4 days or more we only have HRES forecasts available at 00z and 12z initialization and validity times, while for the fairest comparison (\cref{sec:app:ensuring_equal_lookahead}) \ourmodel forecasts must be evaluated using 06z and 18z initialization and validity times. In order to perform a paired test, we compare the RMSE of a \ourmodel forecast with an interpolated RMSE of the two HRES forecasts either side of it: one initialized and valid 6 hours earlier, and the other initialized and valid 6 hours later, all with the same lead time. Specifically we compute differences:
\begin{align}
\text{diff-RMSE}_{\text{interp}}(j, \lt, \dinit) =\; & \text{RMSE}_{GC}(j, \lt, \dinit) \\
 &- \frac12 \Bigl( \text{RMSE}_{HRES}(j, \lt, \dinit - 6h) + \text{RMSE}_{HRES}(j, \lt, \dinit + 6h)
\Bigr). \nonumber
\end{align}

We can use these to test the null hypothesis $\mathbb{E}[\text{diff-RMSE}_{\text{interp}}(j, \lt, \dinit)] = 0$, which again doesn't depend on $\dinit$ by the stationarity assumption on the differences. If we further assume that the HRES RMSE time series itself is stationary (or at least close enough to stationary over a 6 hour  window) then $\mathbb{E}[\text{diff-RMSE}_{\text{interp}}(j, \lt, \dinit)] = \mathbb{E}[\text{diff-RMSE}(j, \lt, \dinit)]$ and the interpolated differences can also be used to test deviations from the original null hypothesis that $\mathbb{E}[\text{diff-RMSE}(j, \lt, \dinit)] = 0$.

This stronger stationarity assumption for HRES RMSEs is violated by diurnal periodicity, and in \cref{sec:app:alignment_of_times_of_day} we do see some systematic differences in HRES RMSEs between 00z/12z and 06z/18z validity times. However as discussed there, these systematic differences reduce substantially as lead time grows and they tend to favour HRES, and so we believe that a test of $\mathbb{E}[\text{diff-RMSE}(j, \lt, \dinit)] = 0$ based on $\text{diff-RMSE}_{\text{interp}}$ will be conservative in cases where \ourmodel appears to have greater skill than HRES.

\subsubsection{Confidence intervals for RMSEs}
\label{sec:app:conf_int_rmse}

The error bars in our RMSE skill plots correspond to separate confidence intervals for $\mathbb{E}[\text{RMSE}_{GC}]$ and $\mathbb{E}[\text{RMSE}_{HRES}]$ (eliding for now the arguments $j, \lt, \dinit$). These are derived from the two-sided $t$-test with correction for auto-correlation that is described above, applied separately to \ourmodel and HRES RMSE time-series.

These confidence intervals make a stationarity assumption for the separate \ourmodel and HRES RMSE time series, which as stated above is a stronger assumption that stationarity of the differences and is violated somewhat. Thus these single-sample confidence intervals should be treated as approximate; we do not rely on them in our significance statements.

\subsubsection{Confidence intervals for RMSE skill scores}
\label{sec:app:conf_int_rmse_skill_score}

From the $t$-test described in \cref{sec:app:significance_tests} we can also derive in the standard way confidence intervals for the true difference in RMSEs, however in our skill score plots we would like to show confidence intervals for the true RMSE skill score, in which the true difference is normalized by the true RMSE of HRES:
\begin{align}
\text{RMSE-SS}_{\text{true}} &= \frac{\mathbb{E}[\text{RMSE}_{GC} - \text{RMSE}_{HRES}]}{\mathbb{E}[\text{RMSE}_{HRES}]}
\end{align}
A confidence interval for this quantity should take into account the uncertainty of our estimate of the true HRES RMSE.
Let $[l_{\text{diff}}, u_{\text{diff}}]$ be our $1-\alpha/2$ confidence interval for the numerator (difference in RMSEs), and $[l_{\text{HRES}}, u_{\text{HRES}}]$ our $1-\alpha/2$  confidence interval for the denominator (HRES RMSE). Given that $0<l_{HRES}$ in every case for us, using interval arithmetic and the union bound we obtain a conservative $1-\alpha$ confidence interval
\begin{align}
[\min \{
  l_{\text{diff}}/u_{\text{HRES}},
  l_{\text{diff}}/l_{\text{HRES}}
\}, 
\max \{
  u_{\text{diff}}/u_{\text{HRES}},
  u_{\text{diff}}/l_{\text{HRES}},
\}]
\end{align}
for $\text{RMSE-SS}_\text{true}$. We plot these confidence intervals alongside our estimates of the RMSE skill score, however note that we don't rely on them for significance testing.

\FloatBarrier

\newpage
\section{Comparison with previous machine learning baselines}\label{sec:app:baselinedetails}

\begin{figure}
  \centering
  \includegraphics[width=\textwidth]{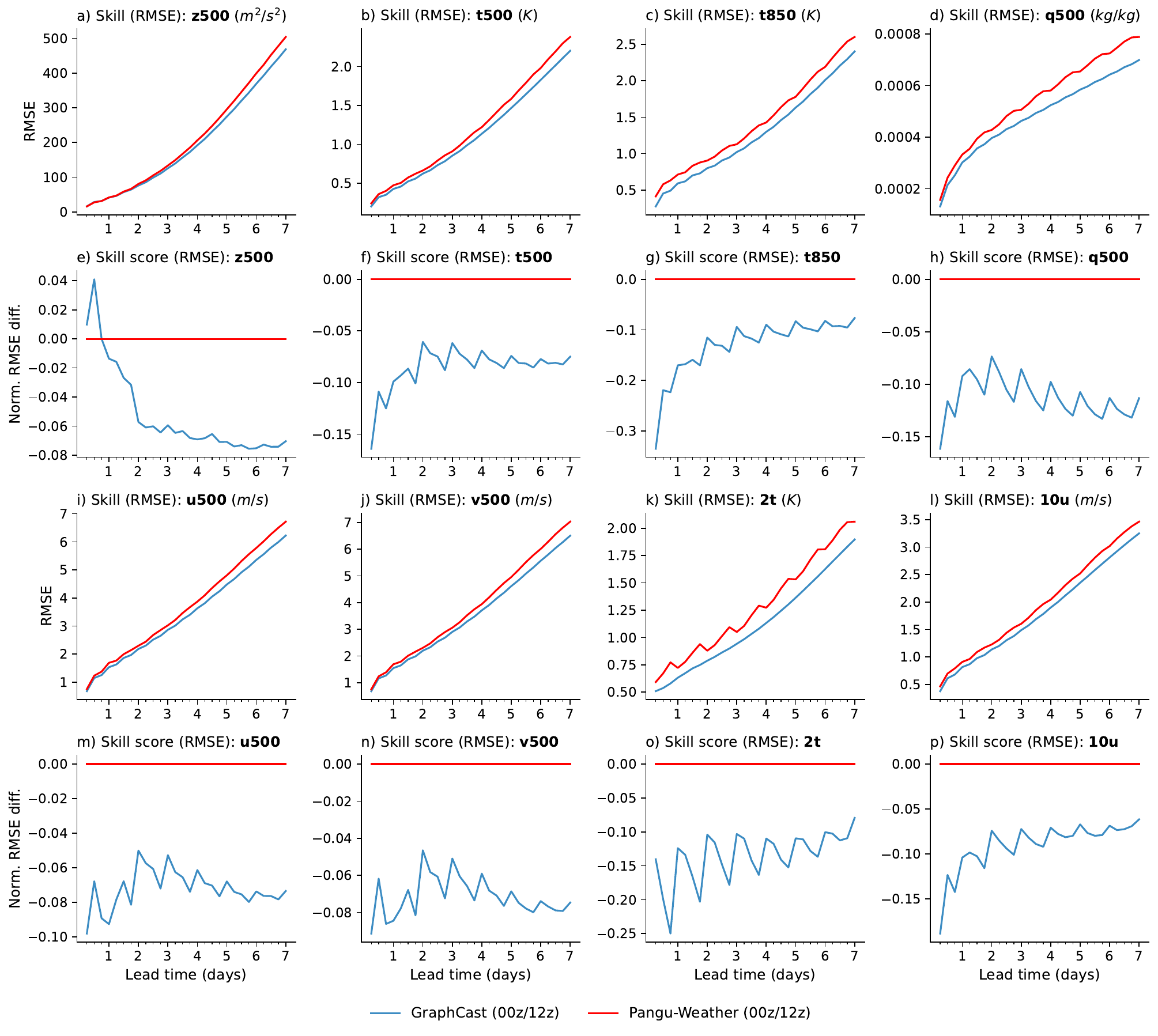}
  \caption{\small\textbf{Comparison between \ourmodel{} and \pangumodel, on RMSE skill.}  Rows 1 and 3 show absolute RMSE for \ourmodel{} (blue lines),
  and \pangumodel{}~\cite{bi2022pangu} (red lines);
  rows 2 and 4 show normalized RMSE differences between the models with respect to \pangumodel{}. Each subplot represents a single variable (and pressure level, for atmospheric variables), as indicated in the subplot titles. The x-axis represents lead time, at 6-hour steps over 10 days. The y-axis represents (absolute or normalized) RMSE. The variables and levels were chosen to be those reported by \cite{bi2022pangu}.
    We did not include \varlevel{10v}, because \varlevel{10u} is already present, and the two are highly correlated.
  }
  \label{fig:baselines}
\end{figure}

To determine how \ourmodel's performance compares to other ML methods, we focus on \pangumodel{}~\cite{bi2022pangu}, a strong MLWP baseline that operates at \quarterdegree{} resolution.
To make the most direct comparison, 
we depart from our evaluation protocol, and use the one described in \cite{bi2022pangu}.
Because published \pangumodel{} results are obtained from the 00z/12z initializations, we use those same initializations for \ourmodel, instead of 06z/18z, as in the rest of this paper. This allows both models to be initialized on the same inputs, which incorporate the same amount of lookahead (+9 hours, see \cref{sec:app:ensuring_equal_lookahead,sec:app:alignment_of_times_of_day}). As HRES initialization incorporates at most +3 hours lookahead, even if initialized from 00z/12z, we do not show the evaluation of HRES (against ERA5 or against \hresfczero) in this comparison as it would disadvantage it.
The second difference with our protocol is to report performance every 6 hours, rather than every 12 hours. Since both models are evaluated against ERA5, their targets are identical, in particular, for a given lead time, the target incorporates +3 hours or +9 hours of lookahead for both \ourmodel and \pangumodel{}, allowing for a fair comparison.
\pangumodel{}\cite{bi2022pangu} reports its 7-day forecast accuracy (RMSE and ACC) on: \varlevel{z}{500}, \varlevel{t}{500}, \varlevel{t}{850}, \varlevel{q}{500}, \varlevel{u}{500}, \varlevel{v}{500}, \varlevel{2t}, \varlevel{10u}, \varlevel{10v}, and \varlevel{msl}.

As shown in \cref{fig:baselines}, \ourmodel{} (blue lines) outperforms \pangumodel{}~\cite{bi2022pangu} (red lines) on $\percentbetterthanpangu\%$ of targets. For the surface variables (\varlevel{2t}, \varlevel{10u}, \varlevel{10v}, \varlevel{msl}), \ourmodel's error in the first several days is around 10-20\% lower, and over the longer lead times plateaus to around 7-10\% lower error. The only two (of the 252 total) metrics on which \pangumodel{} outperformed \ourmodel{} was \varlevel{z}{500}, at lead times 6 and 12 hours, where \ourmodel{} had $1.7\%$ higher average RMSE (\cref{fig:baselines}a,e).

\newpage
\section{Additional forecast verification results}\label{sec:app:additional_results}

This section provides additional analysis of \ourmodel's performance, giving a fuller picture of its strengths and limitations.
\cref{sec:app:additional_variables} complements the
main results of the paper on additional variables and levels beyond
\varlevel{z}{500}.
\cref{sec:app:resultsalongvariousaxis} further analyses \ourmodel performance broken down by regions, latitude and pressure levels (in particular distinguishing the performance below and above the tropopause), illustrates the biases and the RMSE by latitude longitude and elevation.
\cref{sec:app:ablations} demonstrates that both the multi-mesh and the autoregressive loss play an important role in the performance of \ourmodel.
\cref{sec:app:optimalfiltering} details the approach of optimal blurring applied to HRES and \ourmodel, to ensure that \ourmodel improved performance is not only due to its ability to blur its predictions. It also shows the connection between the number of autoregressive steps in the loss and blurring, demonstrating that autoregressive training does more than just optimally blur predictions.
Finally, \cref{sec:app:spectralanalysis} shows various spectral analyses, demonstrating that in most cases \ourmodel has improved performance over HRES across all horizontal length scales and resolutions.
We also discuss the impact of differences in spectra between ERA5 and HRES.
Together, those results show an extensive evaluation of \ourmodel and a rigorous comparison to HRES.

\subsection{Detailed results for additional variables}\label{sec:app:additional_variables}

\subsubsection{RMSE and ACC}

\begin{figure}%
  \centering
  \includegraphics[width=0.85\textwidth]{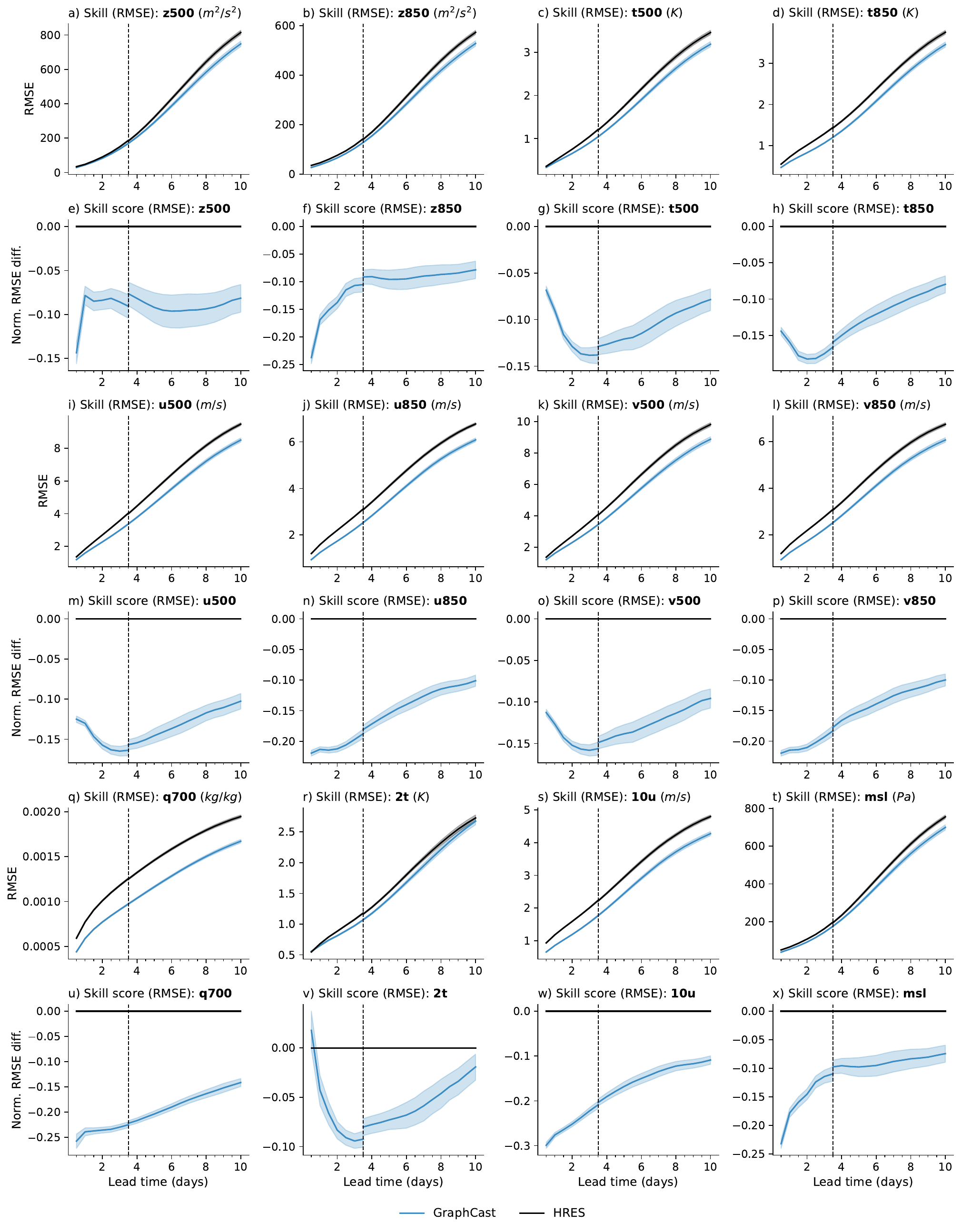}
  \caption{\small\textbf{\ourmodel{}’s RMSE skill versus HRES in 2018 \textit{(lower is better)}}  Rows 1, 3 and 5 show absolute RMSE
for \ourmodel (blue lines) and HRES (black lines), with 95\% confidence interval error bars (see \cref{sec:app:conf_int_rmse}); rows 2, 4 and 6 show RMSE skill score (normalized RMSE differences between
\ourmodel{}’s RMSE and HRES’s) with 95\% confidence interval error bars (see \cref{sec:app:conf_int_rmse_skill_score}). Each subplot represents a single variable (and pressure level), as indicated
in the subplot titles. The x-axis represents lead time, at 12-hour steps over 10 days. The y-axis represents
(absolute or normalized) RMSE.
The vertical dashed line represents 3.5 days, which which marks the transition from HRES forecasts initialized at 06z/18z, to forecast initialized at 00z/12z. This transition explains the discontinuity observed in \ourmodel{}'s skill score curves. }
  \label{fig:app:appendix_main_rmse_with_add_variables}
\end{figure}

\begin{figure}%
  \centering
  \includegraphics[width=0.85\textwidth]{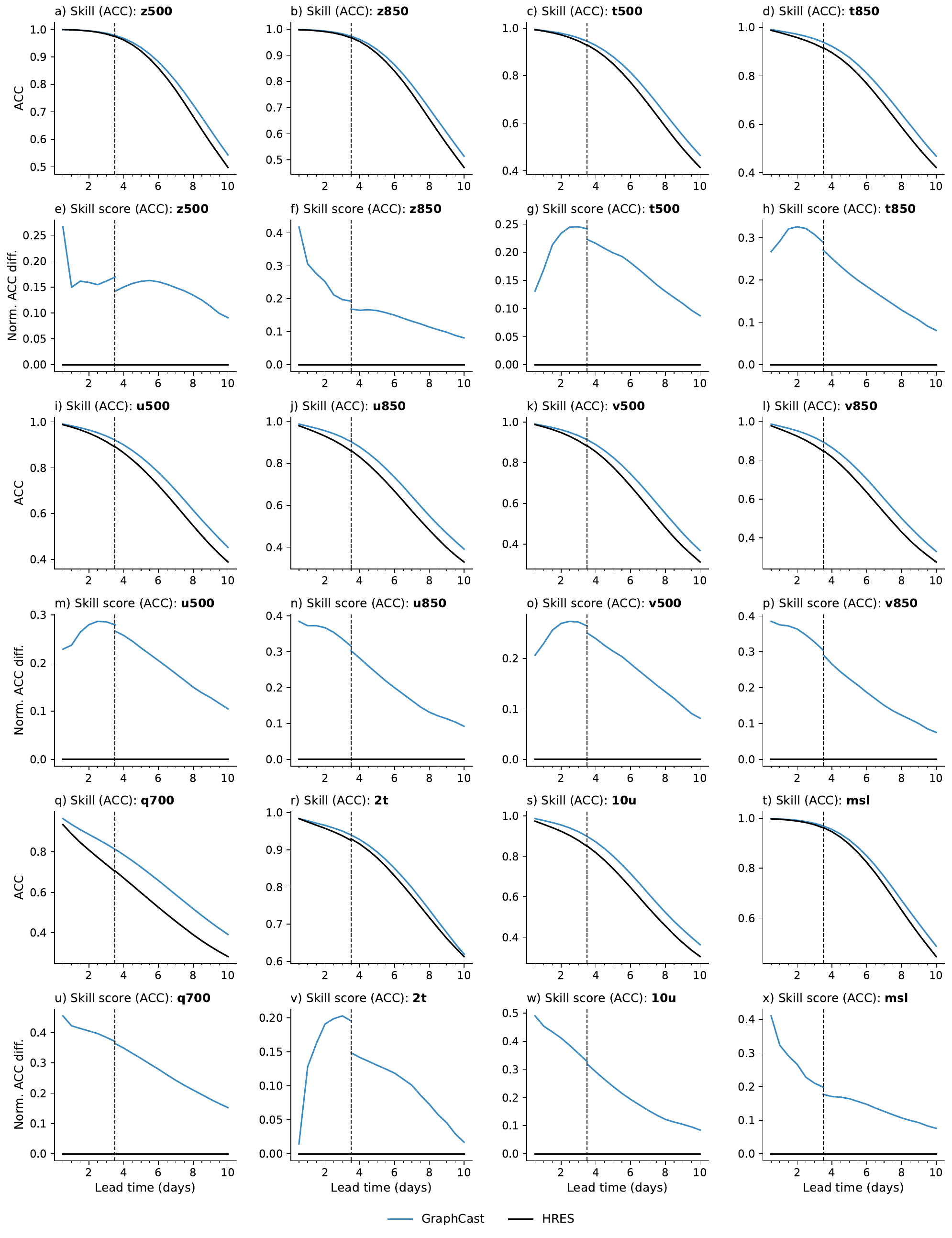}
  \caption{\small\textbf{\ourmodel{}’s ACC skill versus HRES in 2018 \textit{(higher is better)}.} Rows 1, 3 and 5 show absolute ACC
for \ourmodel (blue lines) and HRES (black lines); rows 2, 4 and 6 show ACC skill score (normalized ACC differences between
\ourmodel{}’s RMSE and HRES’s). Each subplot represents a single variable (and pressure level), as indicated
in the subplot titles. The x-axis represents lead time, at 12-hour steps over 10 days. The y-axis represents
(absolute or normalized) ACC.
The vertical dashed line represents 3.5 days, which which marks the transition from HRES forecasts initialized at 06z/18z, to forecast initialized at 00z/12z. This transition explains the discontinuity observed in \ourmodel{}'s skill score curves. }
  \label{fig:app:appendix_main_acc_with_add_variables}
\end{figure}

\cref{fig:app:appendix_main_rmse_with_add_variables} complements \cref{fig:resultshres}a--b and shows the RMSE and normalized RMSE difference with respect to HRES for \ourmodel and HRES on a combination of 12 highlight variables.
\cref{fig:app:appendix_main_acc_with_add_variables} shows the ACC and normalized ACC difference with respect to HRES for \ourmodel and HRES on the same a combination of 12 variables and complements \cref{fig:resultshres}c.
The ACC skill score is the normalized ACC difference between model $A$ and baseline $B$ as $(\mathrm{ACC}_A - \mathrm{ACC}_B) / (1-\mathrm{RMSE}_B)$.


\subsubsection{Detailed significance test results for RMSE comparisons}

\cref{tab:app:significance_test_details} provides further information about the statistical significance claims made in the main section about differences in RMSE between \ourmodel and HRES. Details of the methodology are in \cref{sec:app:statistical_testing}. Here we give $p$-values, test statistics and effective sample sizes for all variables. For reasons of space we limit ourselves to three key lead times (12 hours, 2 days and 10 days) and a subset of 7 pressure levels chosen to include all cases where $p > 0.05$ at these lead times.

\begin{table}
    \centering
  \begin{math}
  \begin{array}{|c|r r r|r r r|r r r|}
  \hline
    \text{Lead time} & \multicolumn{3}{|c|}{\text{12 hours}} & \multicolumn{3}{|c|}{\text{2 days}} & \multicolumn{3}{|c|}{\text{10 days}} \\
\hline
  \text{Variable} & \multicolumn{1}{|c}{p} & \multicolumn{1}{c}{t} & \multicolumn{1}{c|}{n_{\text{eff}}} & \multicolumn{1}{|c}{p} & \multicolumn{1}{c}{t} & \multicolumn{1}{c|}{n_{\text{eff}}} & \multicolumn{1}{|c}{p} & \multicolumn{1}{c}{t} & \multicolumn{1}{c|}{n_{\text{eff}}} \\
  \hline
    \varlevel{z}{50} & <10^{-9} & 11 & 27.8 & <10^{-9} & 6.87 & 37.3 & <10^{-9} & 14.9 & 84.5 \\
  \varlevel{z}{100} & 0.0045 & -2.85 & 88 & 0.044 & 2.01 & 38.3 & 0.0088 & -2.63 & 178 \\
  \varlevel{z}{300} & <10^{-9} & -23.3 & 239 & <10^{-9} & -17.9 & 311 & <10^{-9} & -12.7 & 275 \\
  \varlevel{z}{500} & <10^{-9} & -30.1 & 319 & <10^{-9} & -20.6 & 337 & <10^{-9} & -12.8 & 268 \\
  \varlevel{z}{700} & <10^{-9} & -50 & 465 & <10^{-9} & -28.3 & 334 & <10^{-9} & -12.3 & 257 \\
  \varlevel{z}{850} & <10^{-9} & -59.8 & 452 & <10^{-9} & -33.5 & 332 & <10^{-9} & -12.1 & 259 \\
  \varlevel{z}{1000} & <10^{-9} & -76.9 & 462 & <10^{-9} & -40.5 & 349 & <10^{-9} & -12.1 & 261 \\
  \varlevel{t}{50} & <10^{-9} & 31.8 & 27.2 & <10^{-9} & 8.56 & 32.9 & <10^{-9} & 40.5 & 54.3 \\
  \varlevel{t}{100} & <10^{-9} & 20.6 & 94.5 & 0.32 & -0.995 & 35.6 & 1.7 \times 10^{-6} & 4.83 & 55.2 \\
  \varlevel{t}{300} & 0.35 & 0.941 & 228 & <10^{-9} & -59.6 & 251 & <10^{-9} & -20 & 198 \\
  \varlevel{t}{500} & <10^{-9} & -40.7 & 224 & <10^{-9} & -71.2 & 366 & <10^{-9} & -17.1 & 279 \\
  \varlevel{t}{700} & <10^{-9} & -60.4 & 186 & <10^{-9} & -72.1 & 202 & <10^{-9} & -17.6 & 298 \\
  \varlevel{t}{850} & <10^{-9} & -78.4 & 243 & <10^{-9} & -90.3 & 229 & <10^{-9} & -16.9 & 244 \\
  \varlevel{t}{1000} & <10^{-9} & -23 & 82.9 & <10^{-9} & -59.7 & 169 & <10^{-9} & -11.8 & 275 \\
  \varlevel{u}{50} & <10^{-9} & 25.5 & 20.9 & <10^{-9} & 10.1 & 20.9 & <10^{-9} & 19.6 & 55 \\
  \varlevel{u}{100} & 4.5 \times 10^{-5} & 4.1 & 158 & <10^{-9} & -14.7 & 86 & <10^{-9} & -10.1 & 133 \\
  \varlevel{u}{300} & <10^{-9} & -41.8 & 235 & <10^{-9} & -76.9 & 295 & <10^{-9} & -25.3 & 294 \\
  \varlevel{u}{500} & <10^{-9} & -114 & 339 & <10^{-9} & -103 & 337 & <10^{-9} & -26.8 & 269 \\
  \varlevel{u}{700} & <10^{-9} & -162 & 285 & <10^{-9} & -124 & 263 & <10^{-9} & -27 & 227 \\
  \varlevel{u}{850} & <10^{-9} & -183 & 275 & <10^{-9} & -134 & 336 & <10^{-9} & -27.6 & 243 \\
  \varlevel{u}{1000} & <10^{-9} & -155 & 183 & <10^{-9} & -134 & 383 & <10^{-9} & -26.6 & 231 \\
  \varlevel{v}{50} & <10^{-9} & 35.5 & 31.8 & <10^{-9} & 9.96 & 36.4 & 0.34 & 0.951 & 175 \\
  \varlevel{v}{100} & 0.023 & 2.28 & 175 & <10^{-9} & -14 & 77.5 & <10^{-9} & -16.5 & 234 \\
  \varlevel{v}{300} & <10^{-9} & -27.2 & 198 & <10^{-9} & -78.7 & 343 & <10^{-9} & -19 & 261 \\
  \varlevel{v}{500} & <10^{-9} & -101 & 331 & <10^{-9} & -96 & 365 & <10^{-9} & -21 & 256 \\
  \varlevel{v}{700} & <10^{-9} & -159 & 297 & <10^{-9} & -127 & 315 & <10^{-9} & -25.3 & 241 \\
  \varlevel{v}{850} & <10^{-9} & -181 & 272 & <10^{-9} & -129 & 335 & <10^{-9} & -25.8 & 260 \\
  \varlevel{v}{1000} & <10^{-9} & -211 & 345 & <10^{-9} & -130 & 367 & <10^{-9} & -25.5 & 275 \\
  \varlevel{q}{50} & <10^{-9} & 24.5 & 43.9 & <10^{-9} & 20.7 & 34.1 & <10^{-9} & 8.25 & 56.6 \\
  \varlevel{q}{100} & 1.8 \times 10^{-8} & -5.69 & 177 & <10^{-9} & -8.91 & 77.6 & 7.9 \times 10^{-5} & -3.97 & 22.5 \\
  \varlevel{q}{300} & <10^{-9} & -170 & 224 & <10^{-9} & -139 & 188 & <10^{-9} & -42.5 & 125 \\
  \varlevel{q}{500} & <10^{-9} & -70.6 & 78.9 & <10^{-9} & -137 & 214 & <10^{-9} & -40.4 & 129 \\
  \varlevel{q}{700} & <10^{-9} & -54.2 & 50 & <10^{-9} & -150 & 180 & <10^{-9} & -49.2 & 166 \\
  \varlevel{q}{850} & <10^{-9} & -128 & 92.1 & <10^{-9} & -222 & 199 & <10^{-9} & -61.4 & 163 \\
  \varlevel{q}{1000} & <10^{-9} & -85.6 & 89.3 & <10^{-9} & -128 & 140 & <10^{-9} & -28.8 & 215 \\
  \varlevel{2t} & 0.037 & 2.09 & 38.9 & <10^{-9} & -23.4 & 108 & 0.00075 & -3.39 & 249 \\
  \varlevel{10u} & <10^{-9} & -175 & 143 & <10^{-9} & -156 & 370 & <10^{-9} & -29.9 & 239 \\
  \varlevel{10v} & <10^{-9} & -281 & 298 & <10^{-9} & -160 & 365 & <10^{-9} & -28.8 & 283 \\
  \varlevel{msl} & <10^{-9} & -82.4 & 501 & <10^{-9} & -41.2 & 360 & <10^{-9} & -12 & 260 \\
  \hline
\end{array}
\end{math}
    \caption{\small\textbf{Detailed significance test results for the comparison of \ourmodel and HRES RMSE}. We list the $p$-value, the test statistic $t$ and the effective sample size $n_{\text{eff}}$ for all variables at three key lead times, and a subset of 7 levels chosen to include all cases where $p > 0.05$ at these lead times. Nominal sample size $n \in \{729, 730\}$.}
    \label{tab:app:significance_test_details}
\end{table}

\subsubsection{Effect of data recency on \ourmodel}\label{sec:app:datarecency}

An important feature of MLWP methods is they can be re-trained periodically with the most recent data. This, in principle, allows them to model recent weather patterns that change over time, such as the ENSO cycle and other oscillations, as well as the effects of climate change.
To explore how the recency of the training data influences \ourmodel's test performance, we trained four variants of \ourmodel, with training data that always began in 1979, but ended in 2017, 2018, 2019, and 2020, respectively (we label the variant ending in 2017 as ``\ourmodel:$<$2018'', etc). We evaluated the variants, and HRES, on 2021 test data.

\begin{figure}%
  \centering
  \includegraphics[width=0.9\textwidth]{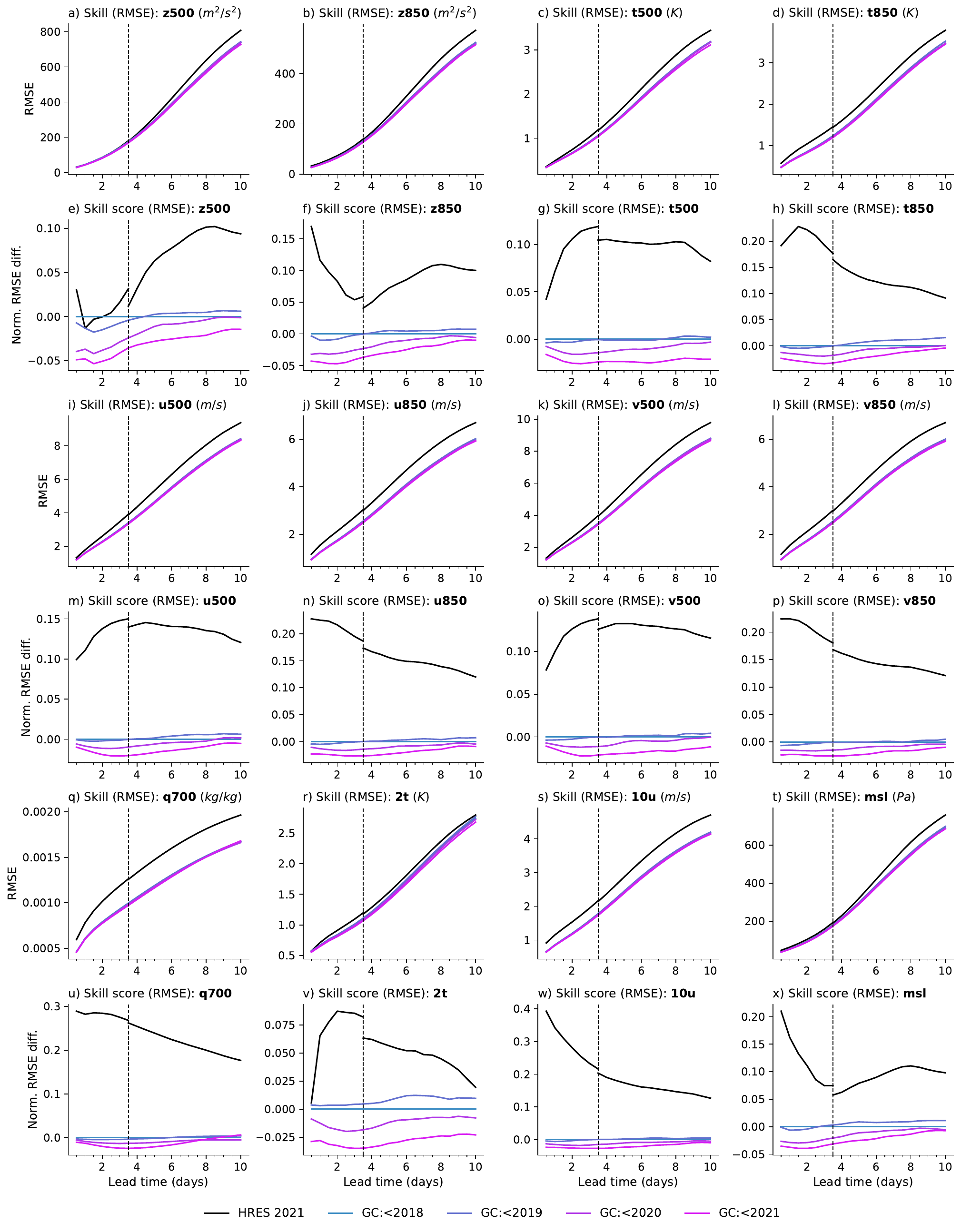}
  \caption{\small\textbf{Effects of data recency.} Each color line in the plot represent a variant of \ourmodel trained on different amounts of available data, while the black line represents HRES. All models are evaluated on the test year 2021. Rows 1, 3, and 5 show the RMSE, and rows 2, 4, and 6 show the normalized RMSE difference with respect to HRES. Each subplot represents a single variable (and pressure level), as indicated in the subplot titles. \ourmodel's performance can be improved by retraining on the most recent data.}
  \label{fig:app:staggered_multi_variables}
\end{figure}

\cref{fig:app:staggered_multi_variables} shows the skill and skill scores (with respect to HRES) of the four variants of \ourmodel, for several variables and complements \cref{fig:resultsablations}a. There is a general trend where variants trained to years closer to the test year have generally improved skill score against HRES. The reason for this improvement is not fully understood, though we speculate it is analogous to long-term bias correction, where recent statistical biases in the weather are being exploited to improve accuracy. It is also important to note that HRES is not a single NWP across years: it tends to be upgraded once or twice a year, with generally increasing skill on \varlevel{z}{500} and other fields~\cite{ifs-upgrade-2018,ifs-upgrade-2019,ifs-upgrade-2020,ifs-upgrade-2021a,ifs-upgrade-2021b}.
This may also contribute to why \ourmodel:$<$2018 and \ourmodel:$<$2019, in particular, have lower skill scores against HRES at early lead times for the 2021 test evaluation.
We note that for other variables, \ourmodel:$<$2018 and \ourmodel:$<$2019 tend to still outperform HRES.
These results highlight a key feature of \ourmodel, in allowing performance to be automatically improved by re-training on recent data.

\FloatBarrier

\subsection{Disaggregated results}\label{sec:app:resultsalongvariousaxis}

\subsubsection{RMSE by region}

\begin{figure}
  \centering
  \includegraphics[width=0.8\textwidth]{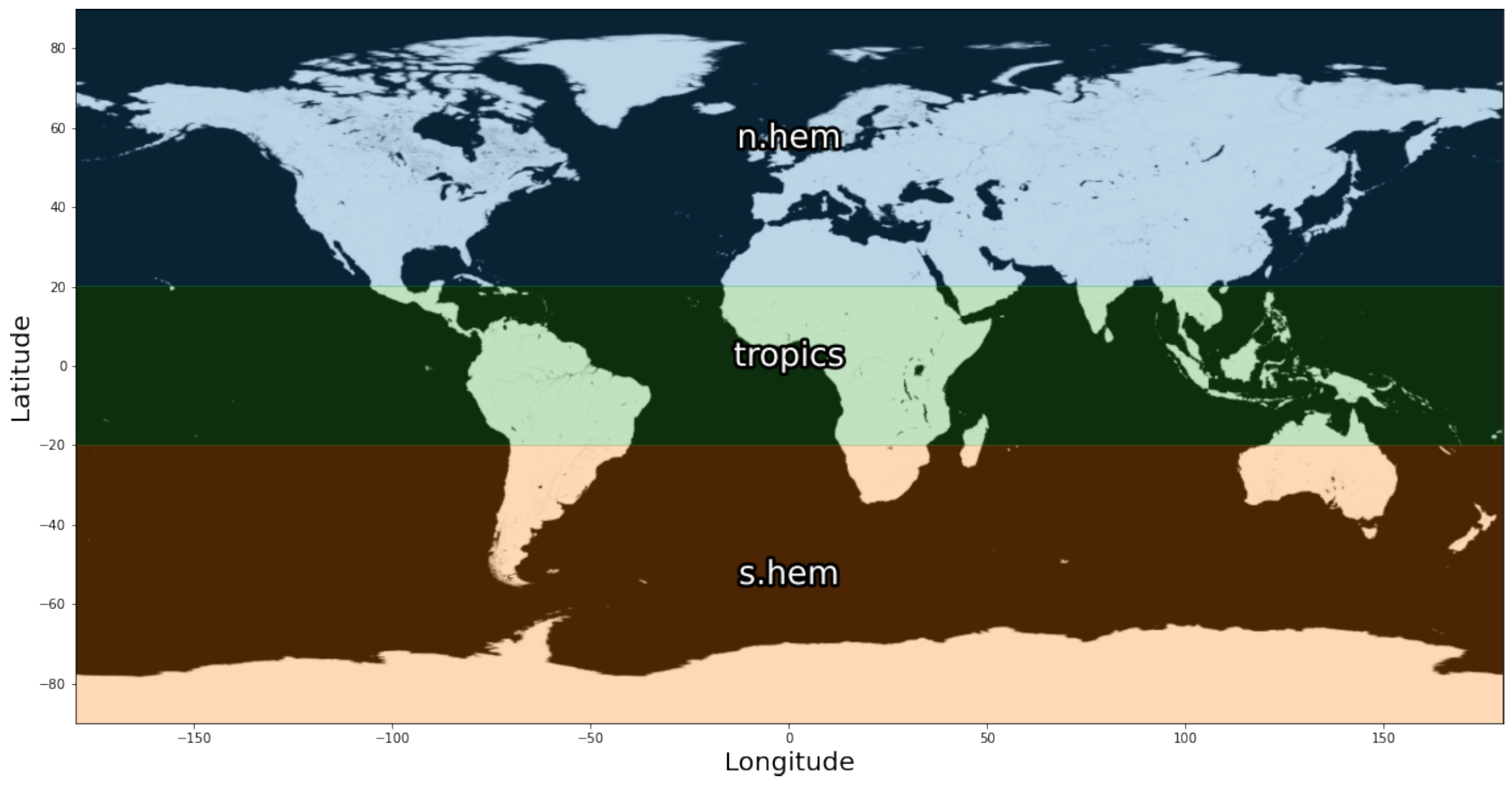}
  \includegraphics[width=0.8\textwidth]{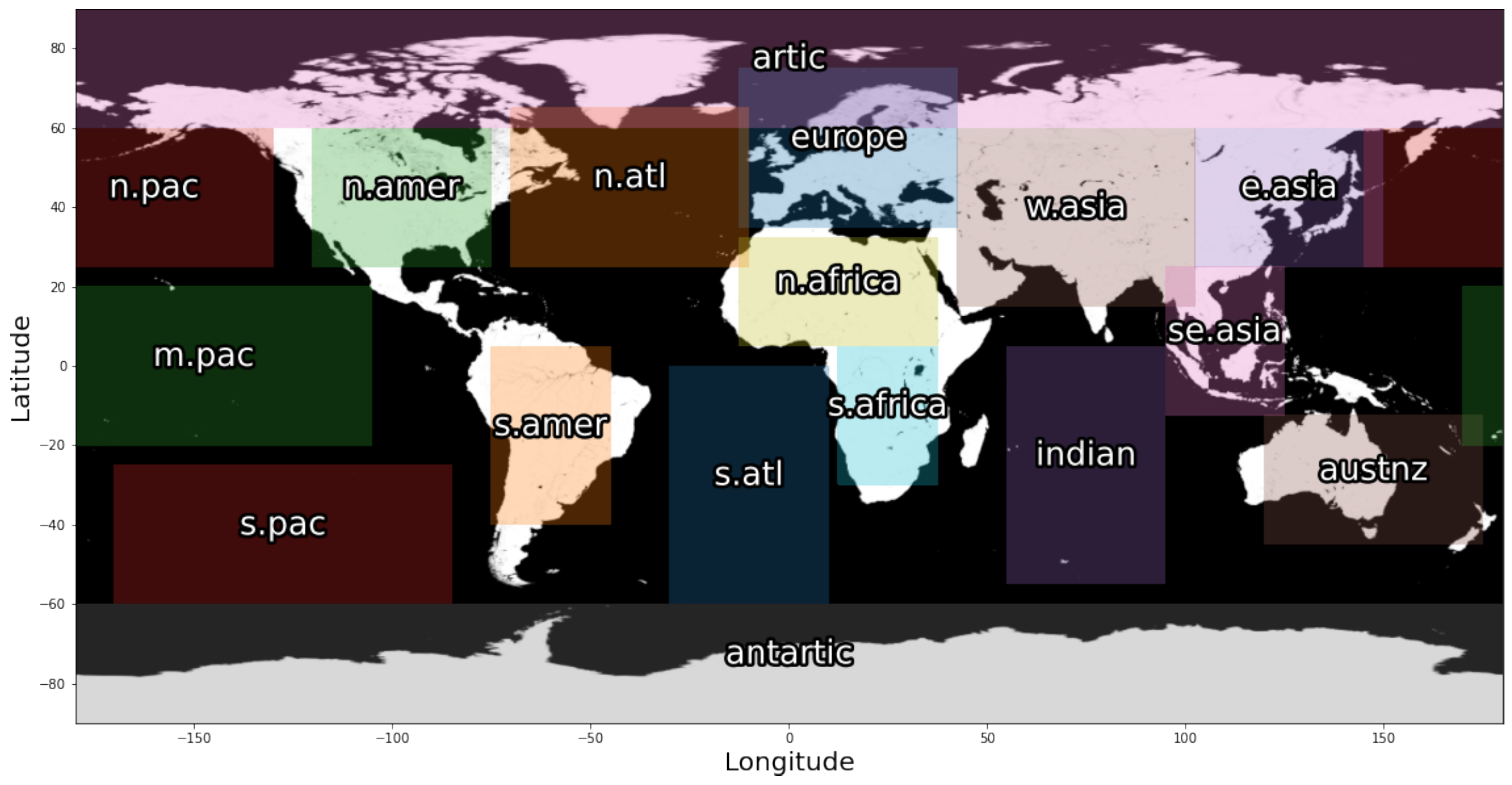}
  \caption{\small\textbf{Region specification for the regional analysis.} We use the same regions and naming convention as in the ECMWF scorecards (\protect\url{https://sites.ecmwf.int/ifs/scorecards/scorecards-47r3HRES.html}), and add some additional regions for better coverage of the entire planet. Per-region evaluation is provided in \cref{fig:app:regional_results_1} and \cref{fig:app:regional_results_2}.}
  \label{fig:app:region_diagram}
\end{figure}

\begin{figure}
  \centering
  \includegraphics[width=.91\textwidth]{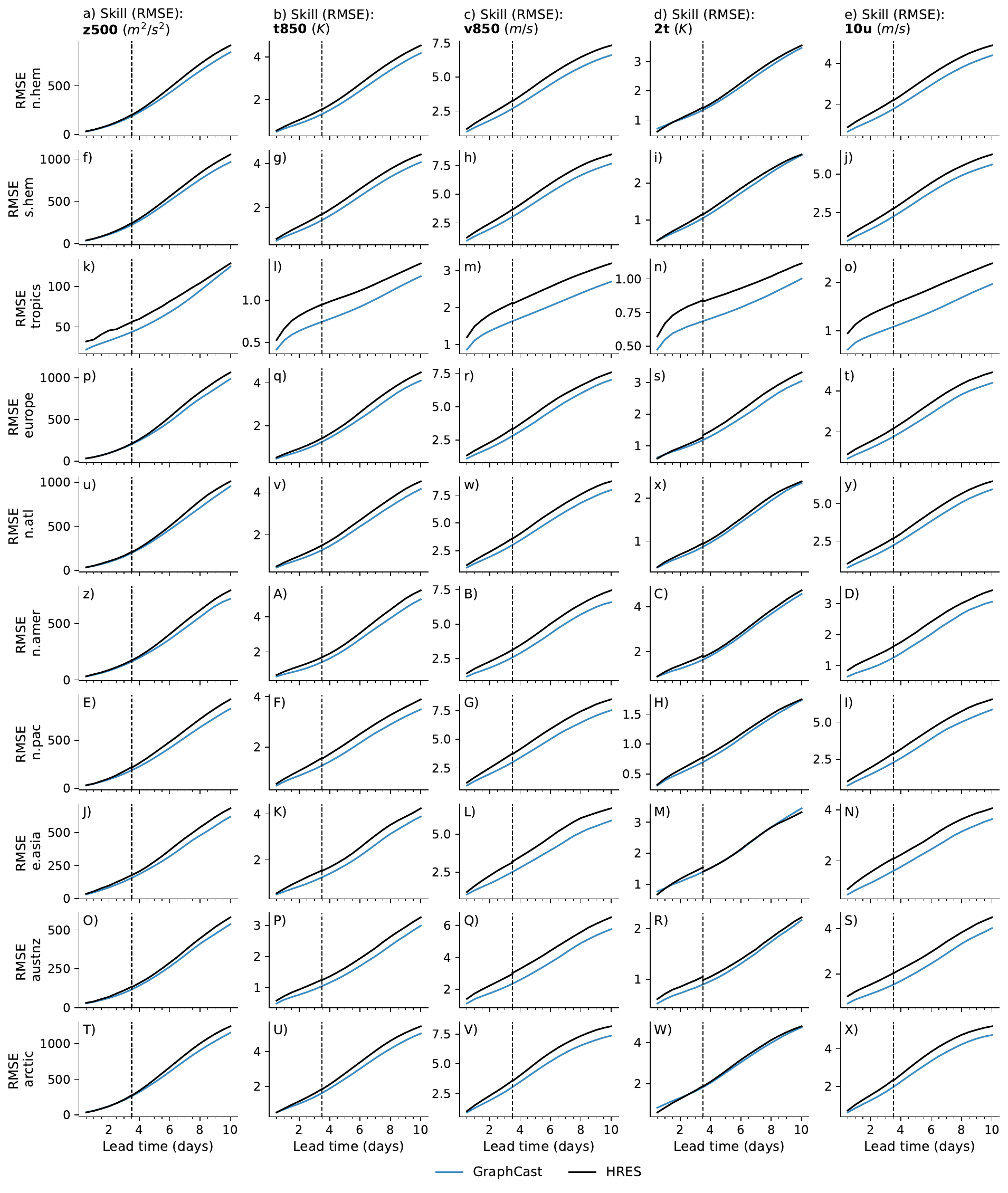}
  \caption{\small\textbf{Skill (RMSE) of \ourmodel versus HRES, per-region.} Each column is a different variable (and level), for a representative set of variables. Each row is a different region. The x-axis is lead time, in days. The y-axis is RMSE, with units specified in the column titles. \ourmodel's RMSEs are the blue lines, and HRES's RMSEs are the black lines. The regions are: \texttt{n.hem}, \texttt{s.hem}, \texttt{tropics}, \texttt{europe}, \texttt{n.atl}, \texttt{n.amer}, \texttt{n.pac}, \texttt{e.asia}, \texttt{austnz}, and \texttt{arctic}. See \cref{fig:app:region_diagram} for a legend of the region names.}
  \label{fig:app:regional_results_1}
\end{figure}

\begin{figure}
  \centering
  \includegraphics[width=.93\textwidth]{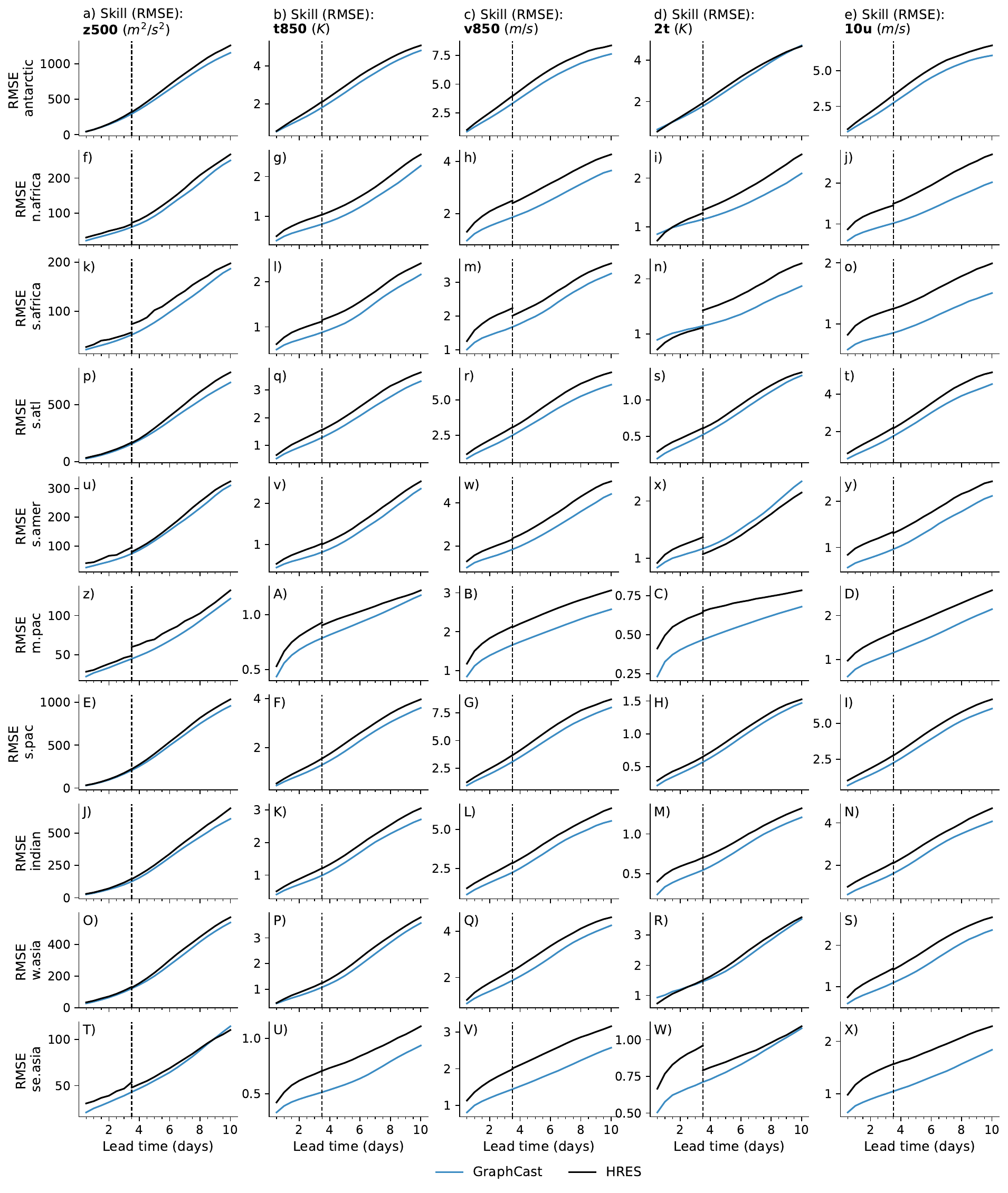}
  \caption{\small\textbf{Skill (RMSE) of \ourmodel{} versus HRES, per-region.} This plot is the same as \cref{fig:app:regional_results_1}, except for regions (in \cref{fig:app:region_diagram}): \texttt{antarctic}, \texttt{n.africa}, \texttt{s.africa}, \texttt{s.atl}, \texttt{s.amer}, \texttt{m.pac}, \texttt{s.pac}, \texttt{indian}, \texttt{w.asia}, and \texttt{se.asia}.}
  \label{fig:app:regional_results_2}
\end{figure}

Per-region evaluation of forecast skill is provided in \cref{fig:app:regional_results_1,fig:app:regional_results_2}, using the same regions and naming convention as in the ECMWF scorecards (\url{https://sites.ecmwf.int/ifs/scorecards/scorecards-47r3HRES.html}). We added some additional regions for better coverage of the entire planet. These regions are shown in \cref{fig:app:region_diagram}.

\FloatBarrier

\subsubsection{RMSE skill score by latitude and pressure level}
\label{app:sec:resultslatlevel}

\begin{figure}[htb]
  \centering
  \includegraphics[width=0.99\textwidth]{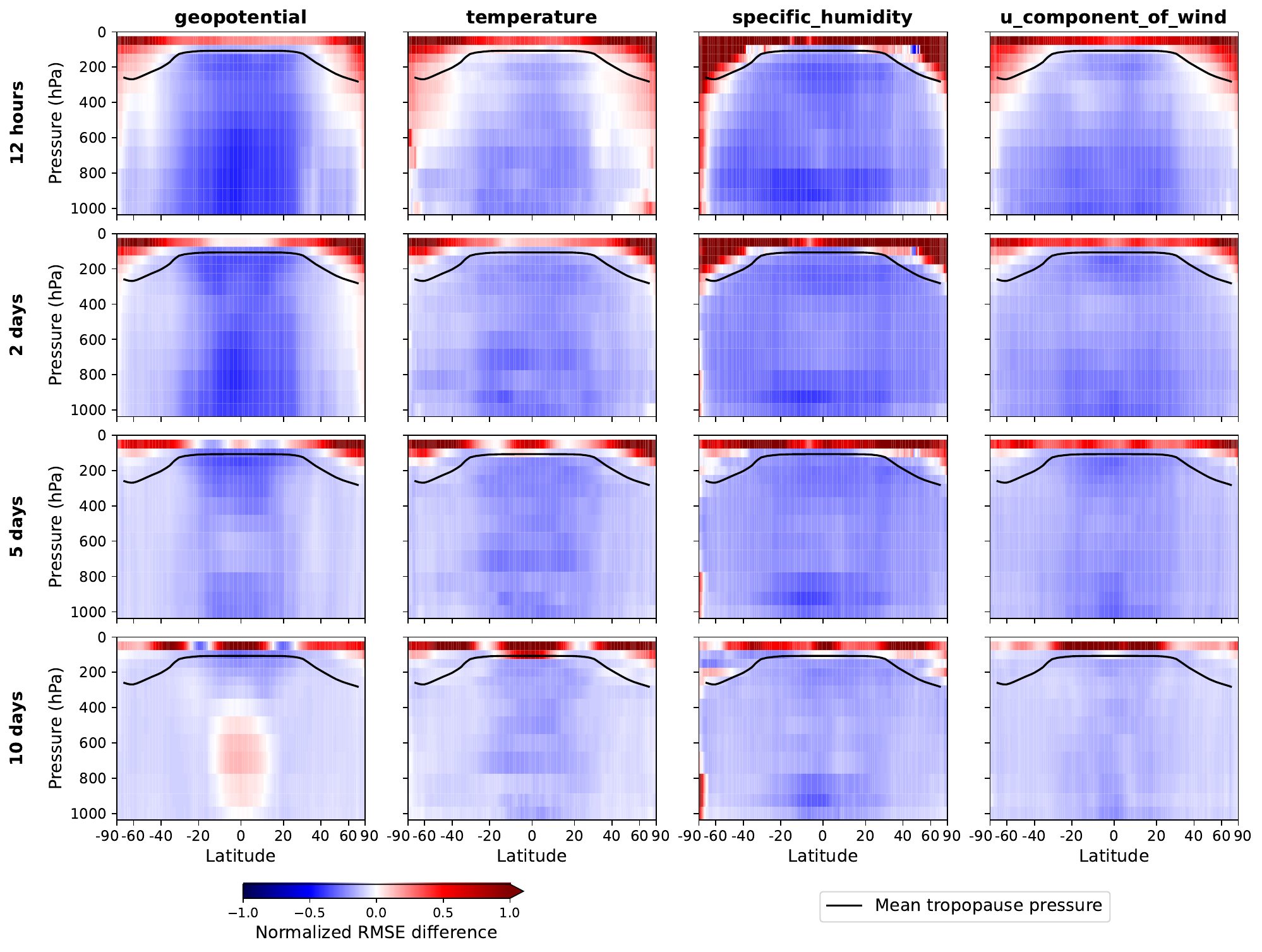}
  \caption{\small\textbf{Normalized RMSE difference of Graphcast relative to HRES, by pressure level and latitude}. Black lines indicate the mean pressure of the tropopause at each latitude; the area above these lines in each plot corresponds roughly to the stratosphere. Latitude spacing is proportional to surface area. Red indicates that HRES has a lower RMSE than \ourmodel, blue the opposite. GraphCast was evaluated using 06z/18z initializations; HRES was evaluated using 06z/18z initializations at 12 hour and 2 day lead times, and 00z/12z at 5 and 10 day lead times (see \cref{sec:app:evaluationdetails}).}
  \label{fig:app:rmse_skill_score_by_lat_level}
\end{figure}

In \cref{fig:app:rmse_skill_score_by_lat_level}, we plot normalized RMSE differences between \ourmodel and HRES, as a function of both pressure level and latitude. We plot only the 13 pressure levels from WeatherBench~\cite{rasp2020weatherbench} on which we have evaluated HRES. 

On these plots, we indicate at each latitude the mean pressure of the tropopause, which separates the troposphere from the stratosphere. We use values computed for the ERA-15 dataset (1979-1993), given in Figure 1 of \cite{santer2003tropopause}.
These will not be quite the same as for ERA5 but are intended only as a rough aid to interpretation. We can see from the scorecard in \cref{fig:resultshres} that \ourmodel performs worse than HRES at the lowest pressure levels evaluated (50hPa). \cref{fig:app:rmse_skill_score_by_lat_level} shows that the pressure level at which \ourmodel starts to get worse is often latitude-dependent too, in some cases roughly following the mean level of the tropopause.

The reasons for \ourmodel's reduced skill in the stratosphere are currently poorly understood. We use a lower loss weighting for lower pressure levels and this may be playing some role; it is also possible that there may be differences between the ERA5 and \hresfczero datasets in the predictability of variables in the stratosphere.

\FloatBarrier

\subsubsection{Biases by latitude and longitude}\label{sec:app:bias_by_lat_lon}

\begin{figure}[hb]
  \centering
  \includegraphics[width=\textwidth]{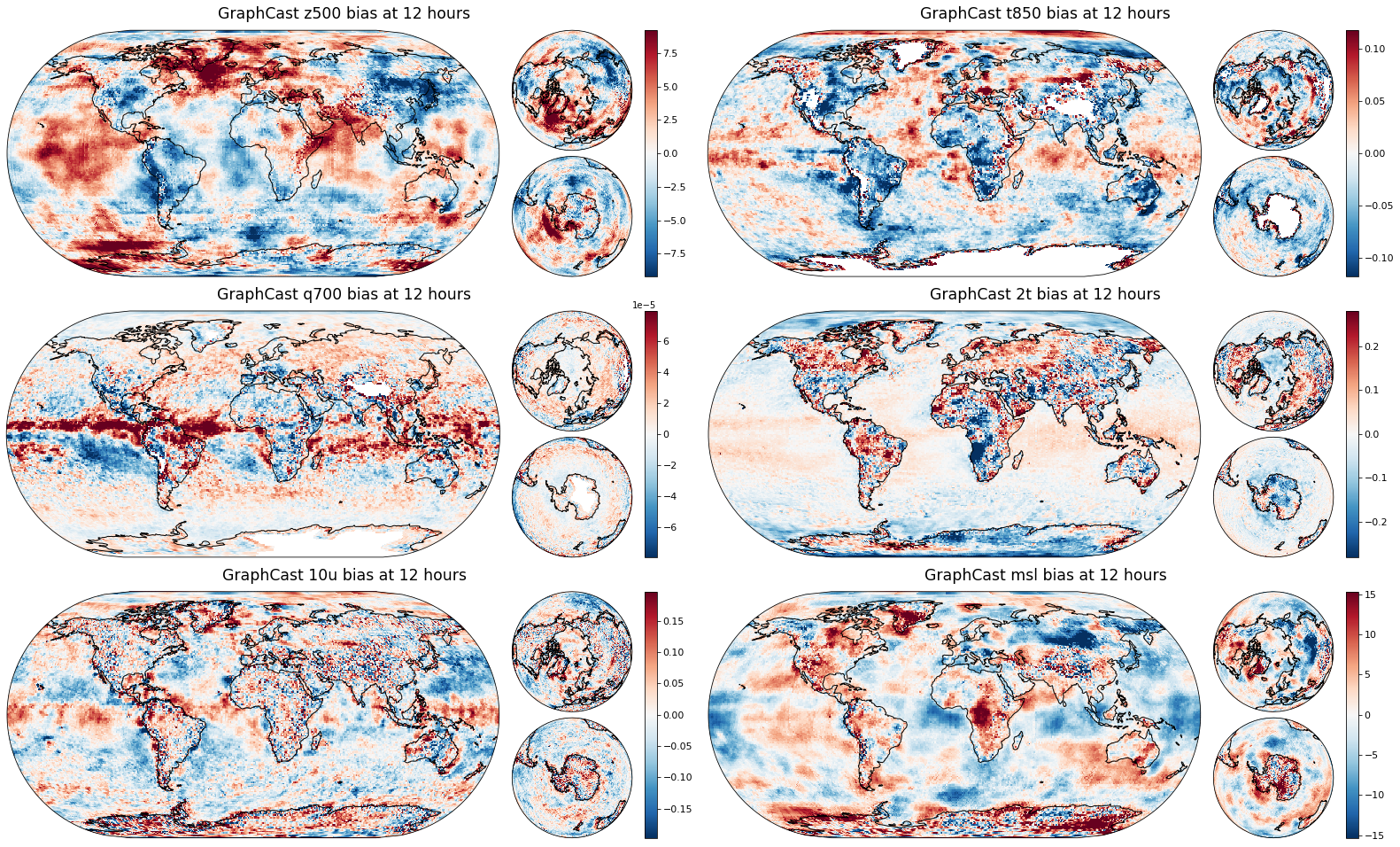}
  \caption{\small\textbf{Mean bias error for \ourmodel at 12 hour lead time, over the 2018 test set.}}
  \label{fig:app:bias_by_lat_lon_graphcast_6h}
\end{figure}

\begin{figure}
  \centering
  \includegraphics[width=\textwidth]{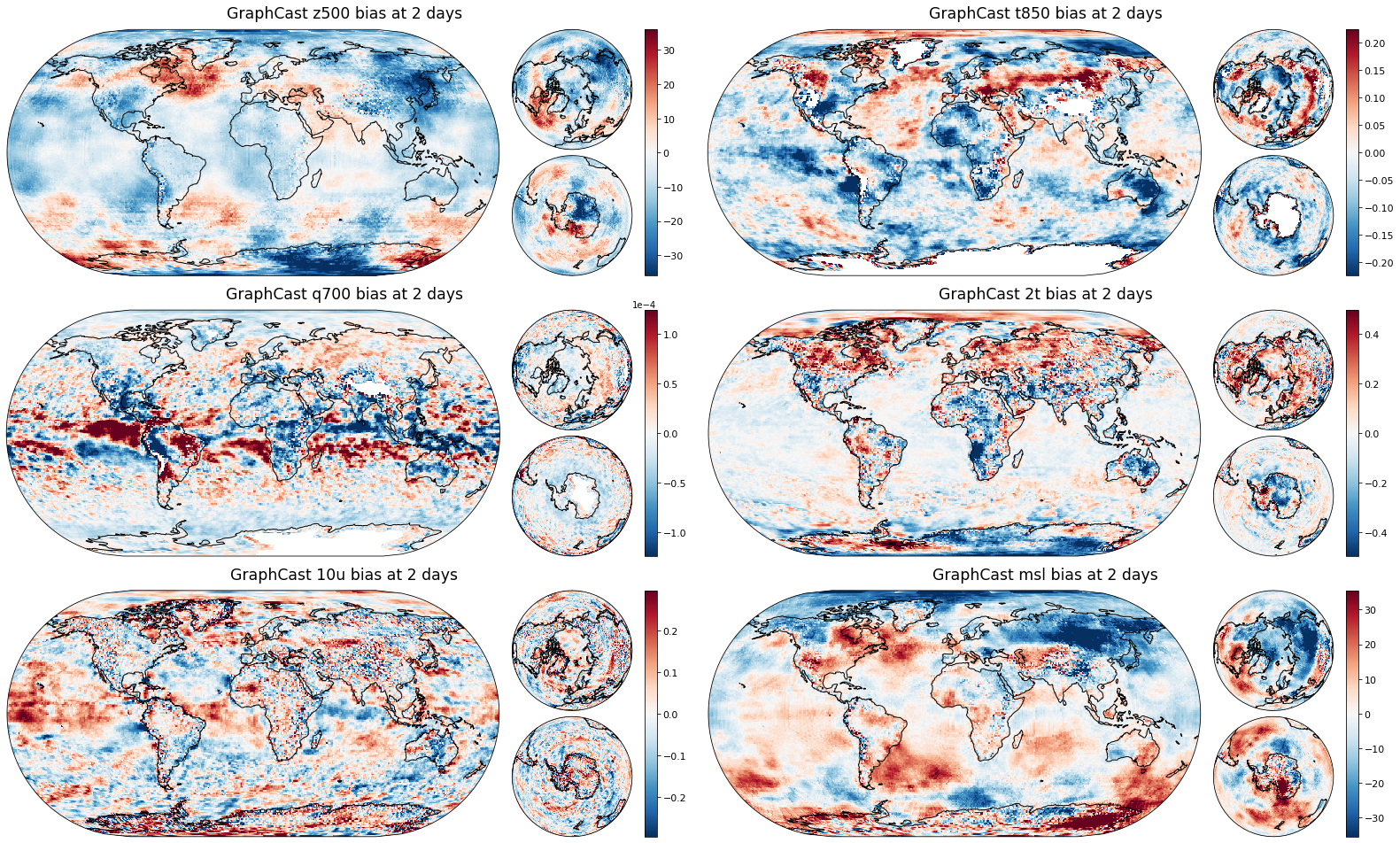}
  \caption{\small\textbf{Mean bias error for \ourmodel at 2 day lead time, over the 2018 test set.}}
  \label{fig:app:bias_by_lat_lon_graphcast_2d}
\end{figure}

\begin{figure}
  \centering
  \includegraphics[width=\textwidth]{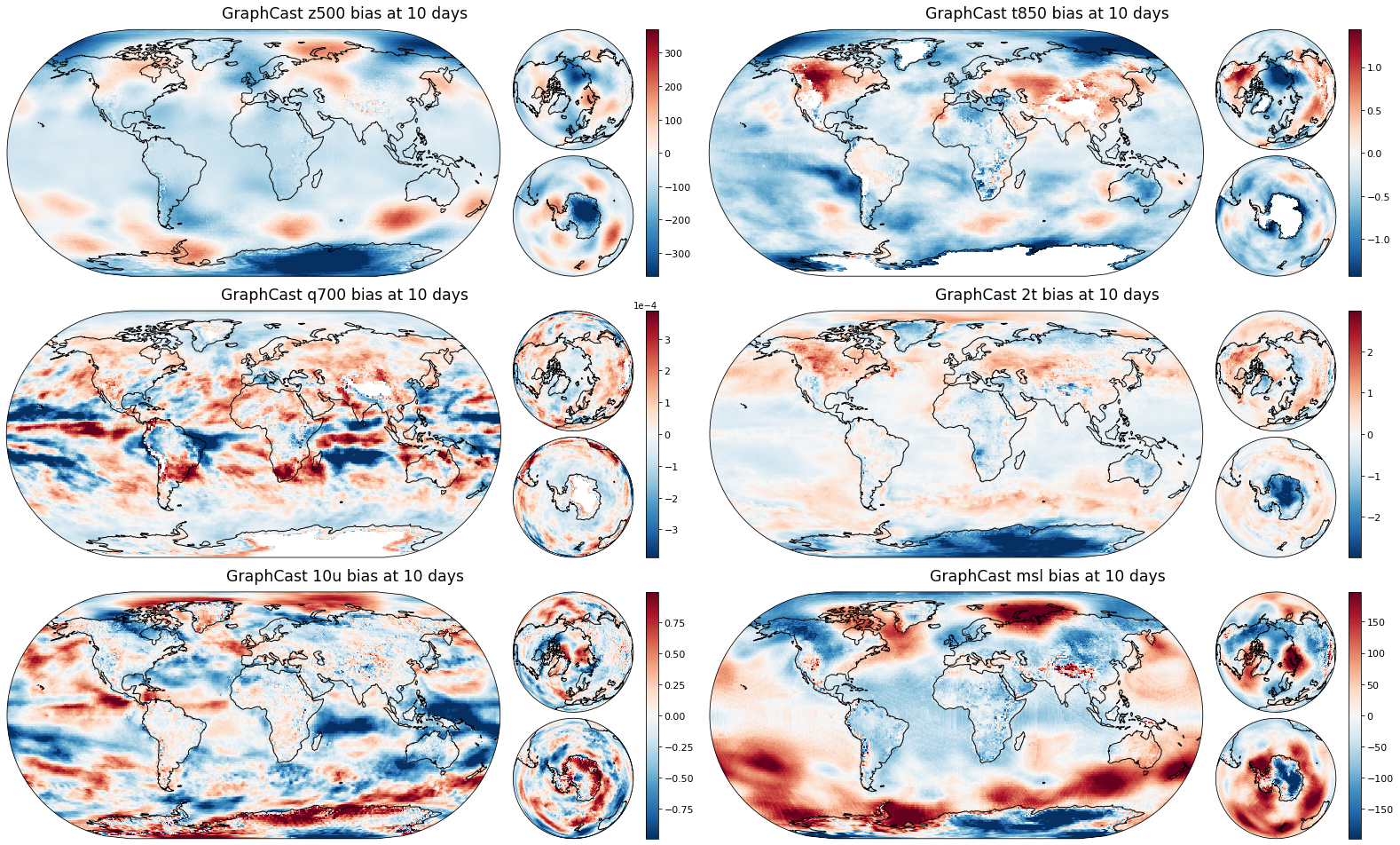}
  \caption{\small\textbf{Mean bias error for \ourmodel at 10 day lead time, over the 2018 test set.}}
  \label{fig:app:bias_by_lat_lon_graphcast_10d}
\end{figure}

In \cref{fig:app:bias_by_lat_lon_graphcast_6h,fig:app:bias_by_lat_lon_graphcast_2d,fig:app:bias_by_lat_lon_graphcast_10d}, we plot the mean bias error (MBE, or just `bias', defined in \cref{eq:app:mbe_lat_lon}) of \ourmodel as a function of latitude and longitude, at three lead times: 12 hours, 2 days and 10 days.

In the plots for variables given on pressure levels, we have masked out regions whose surface elevation is high enough that the pressure level is below ground on average. We determine this to be the case when the surface geopotential exceeds a climatological mean geopotential at the same location and pressure level. In these regions the variable will typically have been interpolated below ground and will not represent a true atmospheric value.


\FloatBarrier

\begin{figure}[ht]
  \centering
  \includegraphics[width=\textwidth]{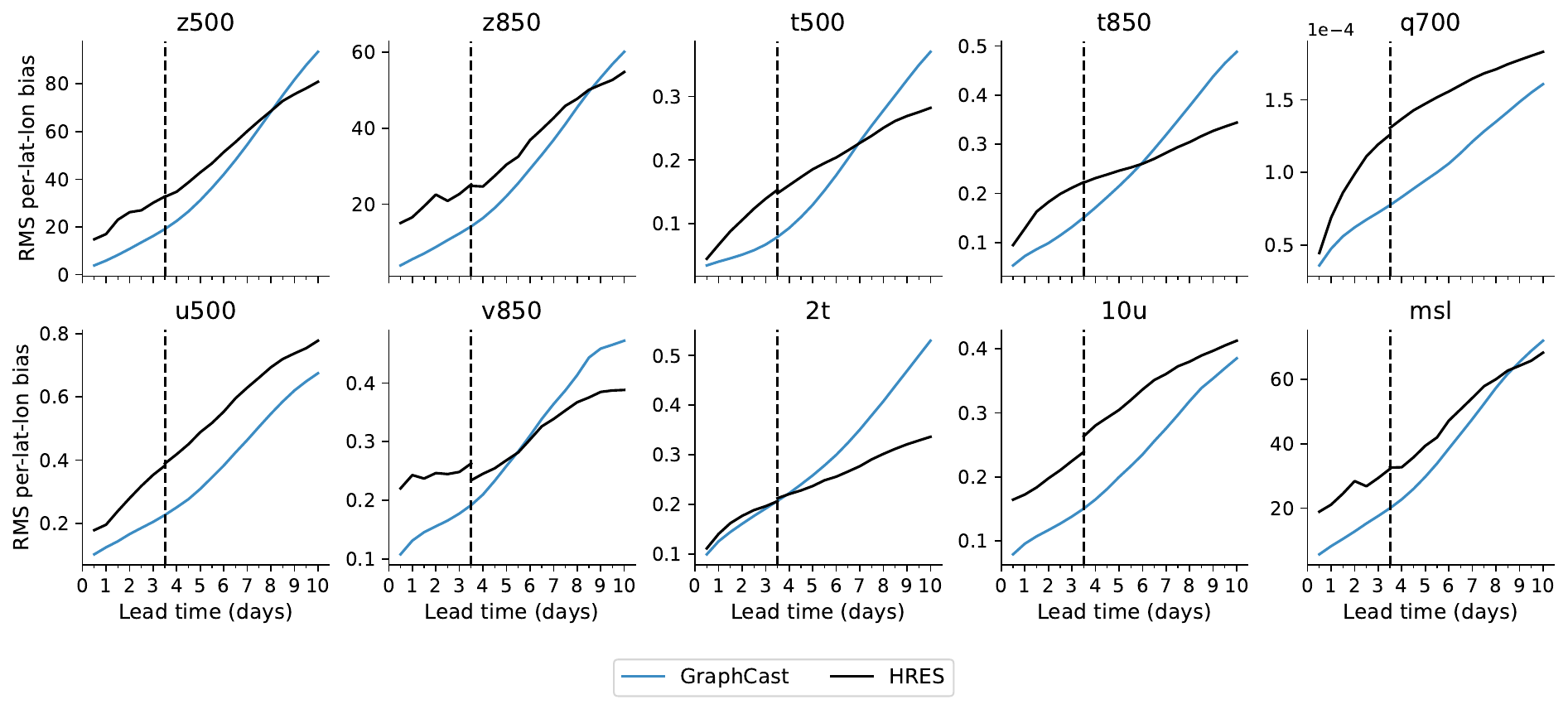}
  \caption{\small\textbf{Root-mean-square of per-location biases for \ourmodel and HRES} as a function of lead time.}
  \label{fig:app:rms_mbe}
\end{figure}

To quantify the average magnitude of the per-location biases shown in \cref{fig:app:bias_by_lat_lon_graphcast_6h,fig:app:bias_by_lat_lon_graphcast_2d,fig:app:bias_by_lat_lon_graphcast_10d}, we computed the root-mean-square of per-location mean bias errors (RMS-MBE, defined in \cref{sec:app:rms_mbe}). These are plotted in \cref{fig:app:rms_mbe} for \ourmodel and HRES as a function of lead time. We can see that \ourmodel's biases are smaller on average than HRES' for most variables up to 6 days. However they generally start to exceed HRES' biases at longer lead times, and at 4 days in the case of 2m temperature.

\FloatBarrier

\begin{figure}[ht]
  \centering
  \includegraphics[width=\textwidth]{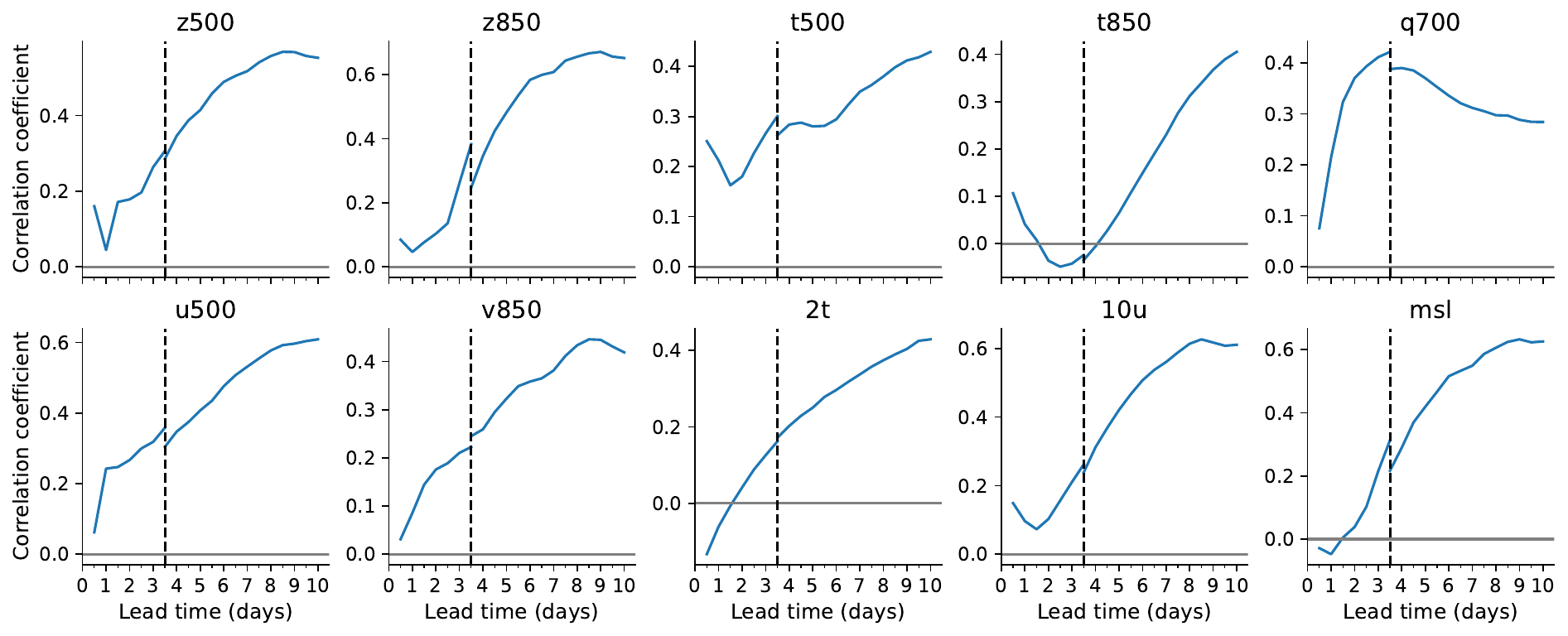}
  \caption{\small\textbf{Correlation of per-location biases between \ourmodel and HRES} as a function of lead time.}
  \label{fig:app:corr_mbe}
\end{figure}

We also computed a correlation coefficient between \ourmodel and HRES' per-location mean bias errors (defined in \cref{sec:app:corr_mbe}), which is plotted as a function of lead time in \cref{fig:app:corr_mbe}. We can see that \ourmodel and HRES' biases are uncorrelated or weakly correlated at the shortest lead times, but the correlation coefficient generally grows with lead time, reaching values as high as 0.6 at 10 days.

\FloatBarrier

\subsubsection{RMSE skill score by latitude and longitude}

\begin{figure}
  \centering
  \includegraphics[width=\textwidth]{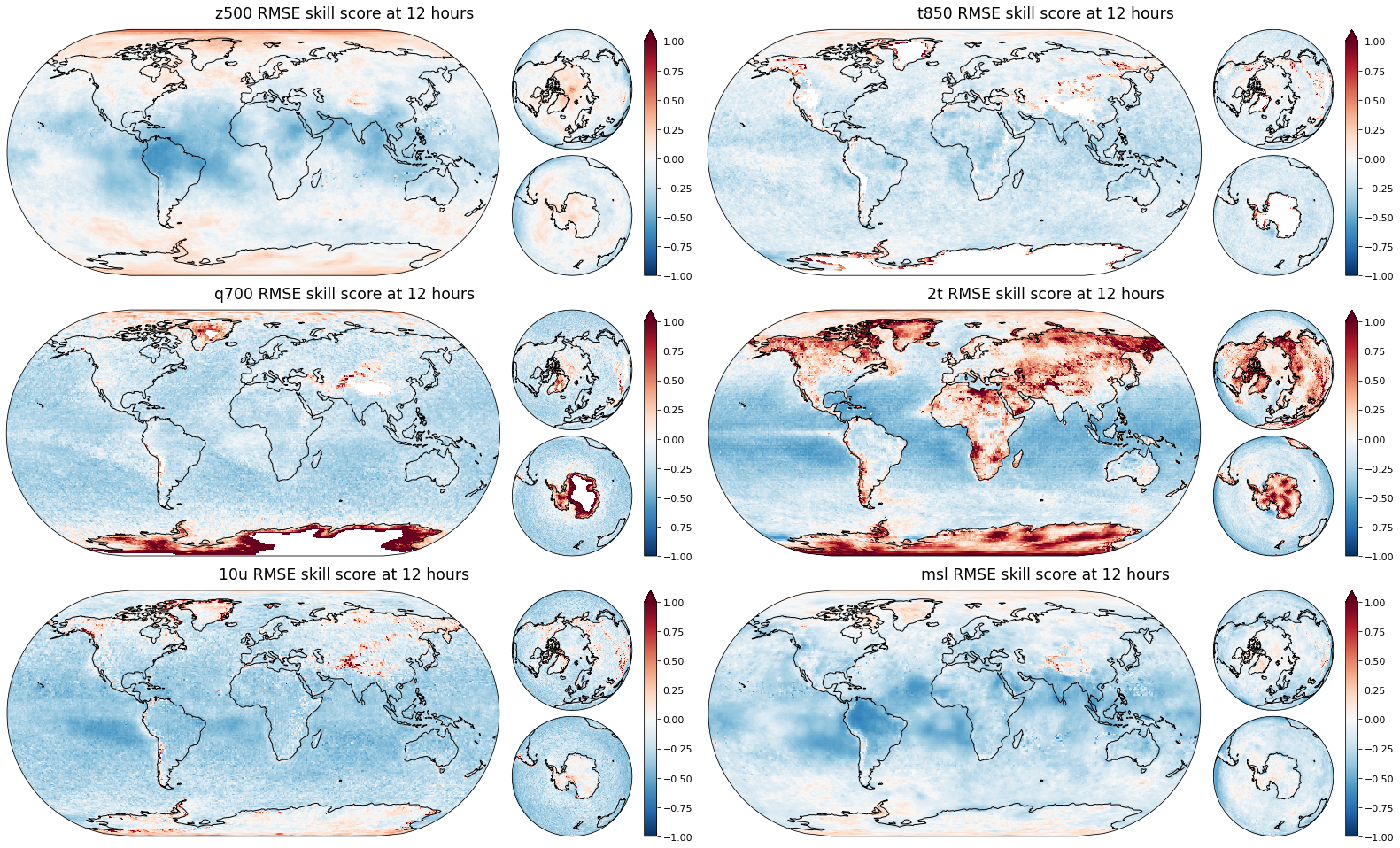}
  \caption{\small\textbf{Normalized RMSE difference of \ourmodel relative to HRES, by location, at 12 hours.} Blue indicates that \ourmodel has greater skill than HRES, Red that HRES has greater skill.}
  \label{fig:app:rmse_skill_score_by_lat_lon_12h}
\end{figure}

\begin{figure}
  \centering
  \includegraphics[width=\textwidth]{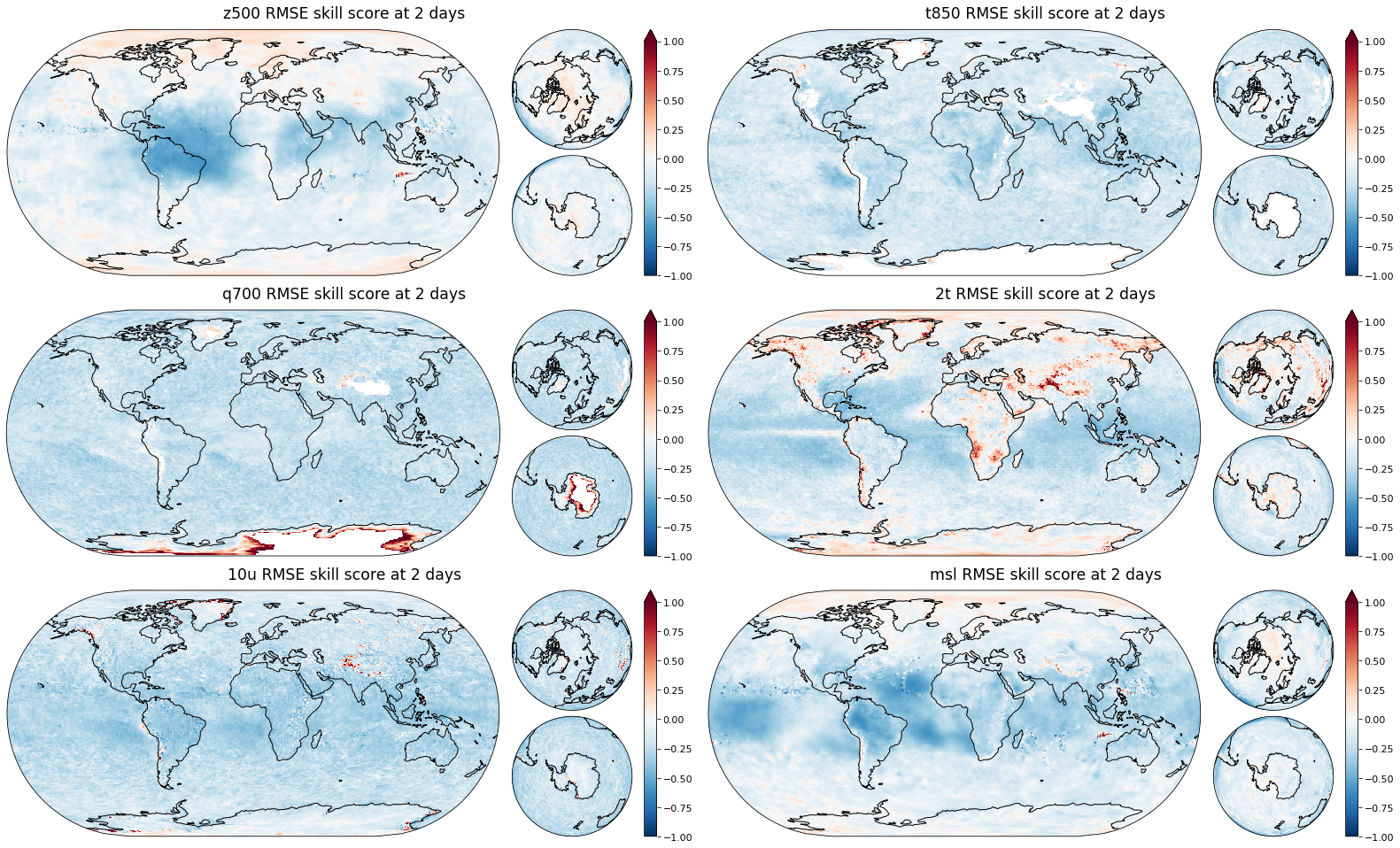}
  \caption{\small\textbf{Normalized RMSE difference of \ourmodel relative to HRES, by location, at 2 days.} Blue indicates that \ourmodel has greater skill than HRES, Red that HRES has greater skill.}
  \label{fig:app:rmse_skill_score_by_lat_lon_2d}
\end{figure}

\begin{figure}
  \centering
  \includegraphics[width=\textwidth]{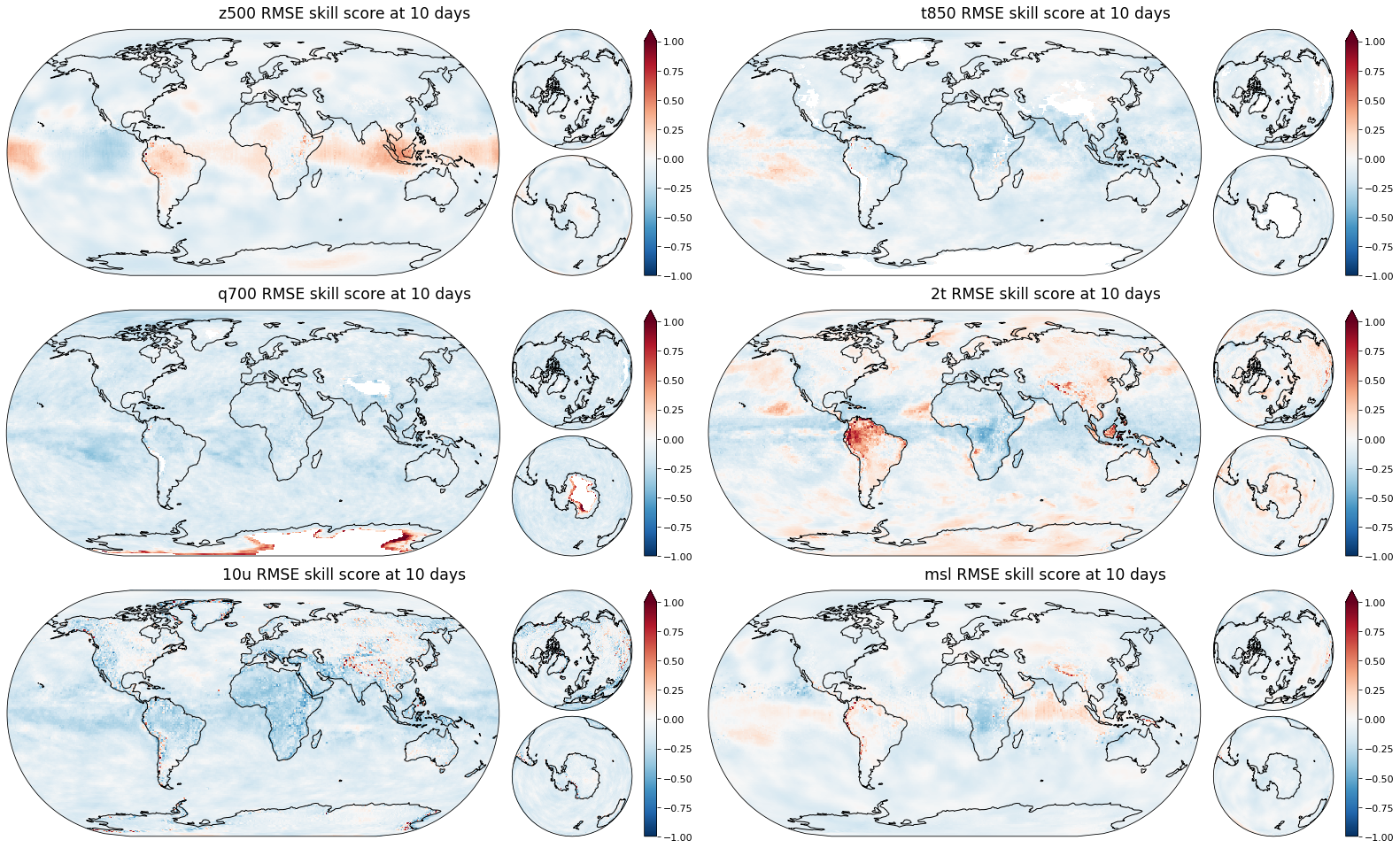}
  \caption{\small\textbf{Normalized RMSE difference of \ourmodel relative to HRES, by location, at 10 days.} Blue indicates that \ourmodel has greater skill than HRES, Red that HRES has greater skill. In these 10 day plots we must compare GraphCast at 06z/18z with HRES at 00z/12z (see \cref{sec:app:evaluationdetails}). This difference in time-of-day may confound some comparisons, e.g. of \varlevel{2t}.}
  \label{fig:app:rmse_skill_score_by_lat_lon_10d}
\end{figure}

In \cref{fig:app:rmse_skill_score_by_lat_lon_12h,fig:app:rmse_skill_score_by_lat_lon_2d,fig:app:rmse_skill_score_by_lat_lon_10d}, we plot the normalized RMSE difference between \ourmodel and HRES by latitude and longitude. As in \cref{sec:app:bias_by_lat_lon}, for variables given on pressure levels, we have masked out regions whose surface elevation is high enough that the pressure level is below ground on average.

Notable areas where HRES outperforms \ourmodel include specific humidity near the poles (particularly the south pole); geopotential near the poles; 2m temperature near the poles and over many land areas; and a number of surface or near-surface variables in regions of high surface elevation (see also \cref{sec:app:rmse_skill_score_by_surface_elevation}). \ourmodel's skill in these areas generally improves over longer lead times. However HRES outperforms \ourmodel on geopotential in some tropical regions at longer lead times.

At 12 hour and 2 day lead times both \ourmodel and HRES are evaluated at 06z/18z initialization and validity times, however at 10 day lead times we must compare \ourmodel at 06z/18z with HRES at 00z/12z (see \cref{sec:app:evaluationdetails}). This difference in time-of-day may confound comparisons at specific locations for variables like 2m temperature (\varlevel{2t}) with a strong diurnal cycle.


\FloatBarrier

\subsubsection{RMSE skill score by surface elevation}

\label{sec:app:rmse_skill_score_by_surface_elevation}

\begin{figure}[ht]
  \centering
  \includegraphics[width=\textwidth]{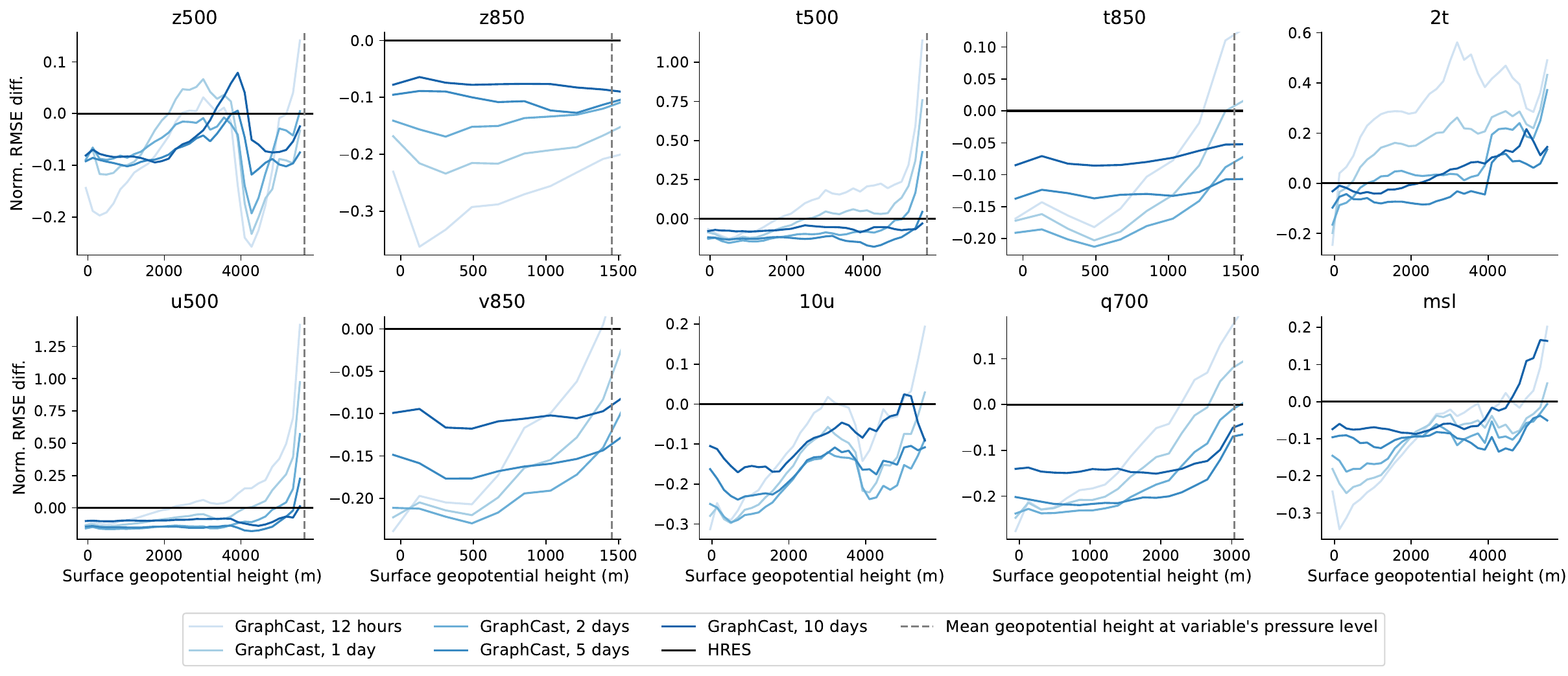}
  \caption{\small\textbf{Normalized RMSE difference of \ourmodel relative to HRES, by surface geopotential height.} For pressure-level variables, we crop the x-axis to exclude surface geopotential heights at which the variable is typically below ground (those greater than the mean geopotential height at the variable's pressure level, indicated via a dotted vertical line).
  }
  \label{fig:app:rmse_skill_score_by_surface_elevation}
\end{figure}

In \cref{fig:app:rmse_skill_score_by_lat_lon_12h}, we can see that \ourmodel appears to have reduced skill in high-elevation regions for many variables at 12 hour lead time. To investigate this further we divided the earth surface into 32 bins by surface elevation (given in terms of geopotential height) and computed RMSEs within each bin according to \cref{sec:app:rmse_by_surface_elevation}. These are plotted in \cref{fig:app:rmse_skill_score_by_surface_elevation}.

At short lead times and especially at 6 hours, \ourmodel's skill relative to HRES tends to decrease with higher surface elevation, in most cases dropping below the skill of HRES at sufficiently high elevations. At longer lead times of 5 to 10 days this effect is less noticeable, however.

We note that \ourmodel is trained on variables defined using a mix of pressure-level coordinates (for atmospheric variables) and height above surface coordinates (for surface-level variables like 2m temperature or 10m wind). The relationship between these two coordinates systems depends on surface elevation. Despite \ourmodel conditioning on surface elevation we conjecture that it may struggle to learn this relationship, and to extrapolate it well to the highest surface elevations.
In further work we would propose to try training the model on a subset of ERA5's native model levels instead of pressure levels; these use a hybrid coordinate system 
\cite{ecmwf2016-ifs-cy41r2-partiii}
which follows the land surface at the lowest levels, and this may make the relationship between surface and atmospheric variables easier to learn, especially at high surface elevations.

Variables using pressure-level coordinates are interpolated below ground when the pressure level exceeds surface pressure. \ourmodel is not given any explicit indication that this has happened and this may add to the challenge of learning to forecast at high surface elevations. In further work using pressure-level coordinates we propose to provide additional signal to the model indicating when this has happened.

Finally, our loss weighting is lower for atmospheric variables at lower pressure levels, and this may affect skill at higher-elevation locations. Future work might consider taking surface elevation into account in this weighting.

\FloatBarrier

\subsection{\ourmodel ablations}\label{sec:app:ablations}

\subsubsection{Multi-mesh ablation}\label{sec:app:multimeshablation}

To better understand how the multi-mesh representation affects the performance of \ourmodel, we compare \ourmodel performance to a version of the model trained without the multi-mesh representation. The architecture of the latter model is identical to \ourmodel (including same encoder and decoder, and the same number of nodes), except that in the process block, the graph only contains the edges from the finest icosahedron mesh $M^6$ (245,760 edges, instead of 327,660 for \ourmodel).
As a result, the ablated model can only propagate information with short-range edges, while \ourmodel contains additional long-range edges.

\begin{figure}%
  \centering
  \includegraphics[width=0.3\textwidth]{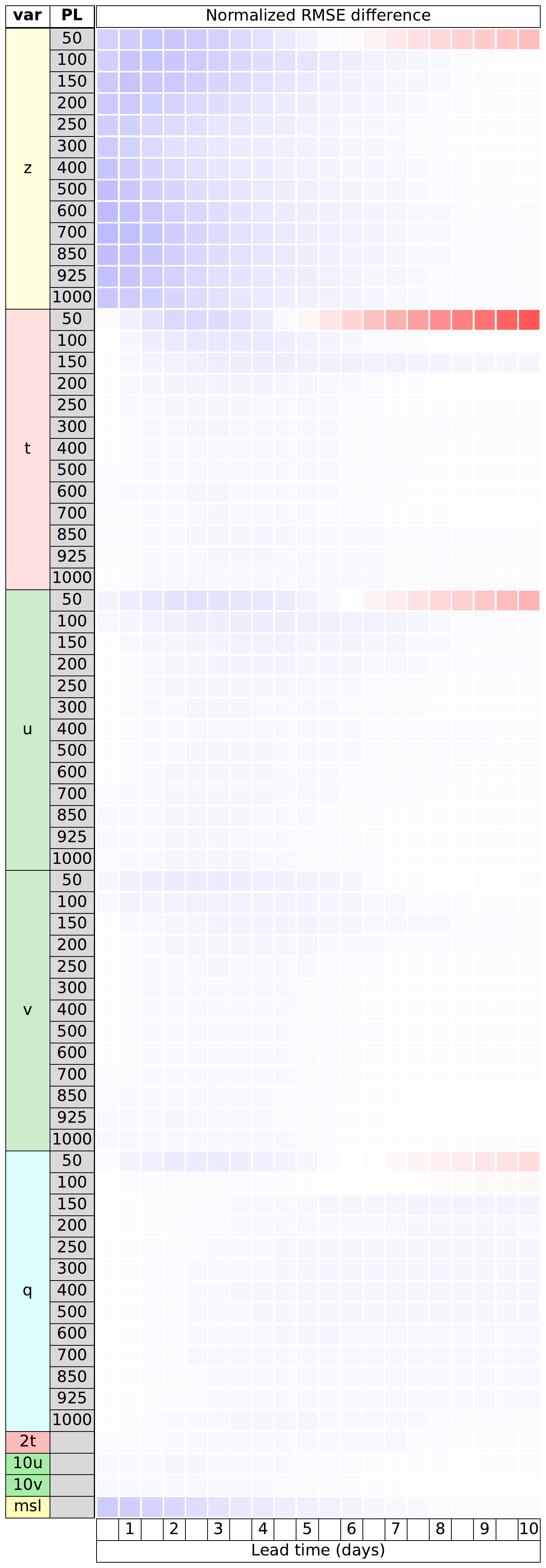}
  ~~\includegraphics[width=0.3\textwidth]{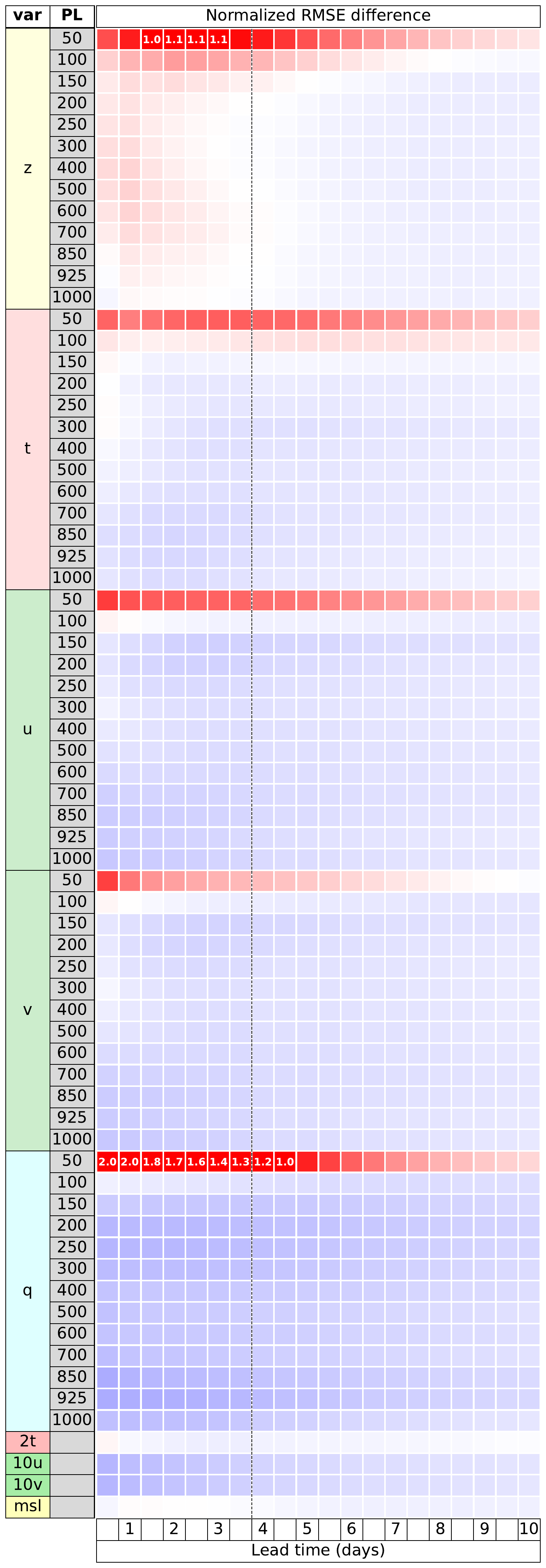}
  ~~\includegraphics[width=0.3\textwidth]{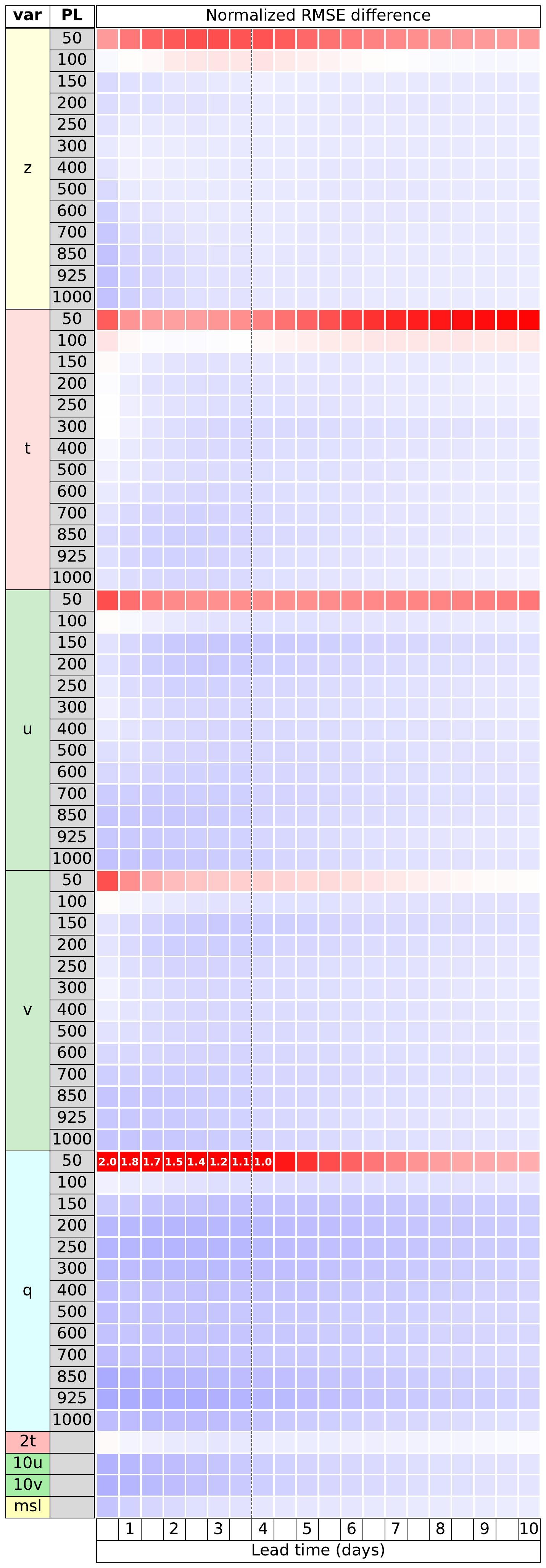}
\caption{\small\textbf{Scorecards comparing \ourmodel to the ablated model without multi-mesh edges (left panel), the ablated model to HRES (middle panel) and \ourmodel to HRES (right panel)}. In the left panel, blue cells represent variables and lead time where \ourmodel is better than the ablated model, showing that training a model with the multi-mesh improves performance for all variables, except at 50~\unit{hPa} past 5 days of lead time. In the middle panel, blue cells represent variables and lead time where the ablated model is better than HRES. Comparing the middle panel to the right one, where blue cells indicate that \ourmodel is better than HRES, shows that the multi-mesh is necessary to outperform HRES on geopotential for lead times under 5 days.}
  \label{fig:multi-mesh_ablation}
\end{figure}

\cref{fig:multi-mesh_ablation} (left panel) shows the scorecard comparing \ourmodel to the ablated model. \ourmodel benefits from the multi-mesh structure for all predicted variables, except for lead times beyond 5 days at 50~\unit{hPa}.
The improvement is especially pronounced for geopotential across all pressure levels and for mean sea-level pressure for lead times under 5 days.
The middle panel shows the scorecard comparing the ablated model to HRES, while the right panel compares \ourmodel to HRES, demonstrating that the multi-mesh is essential for \ourmodel to outperform HRES on geopotential at lead times under 5 days.

\subsubsection{Effect of autoregressive training}\label{sec:app:AR_ablation}

\begin{figure}%
  \centering
  \includegraphics[width=0.85\textwidth]{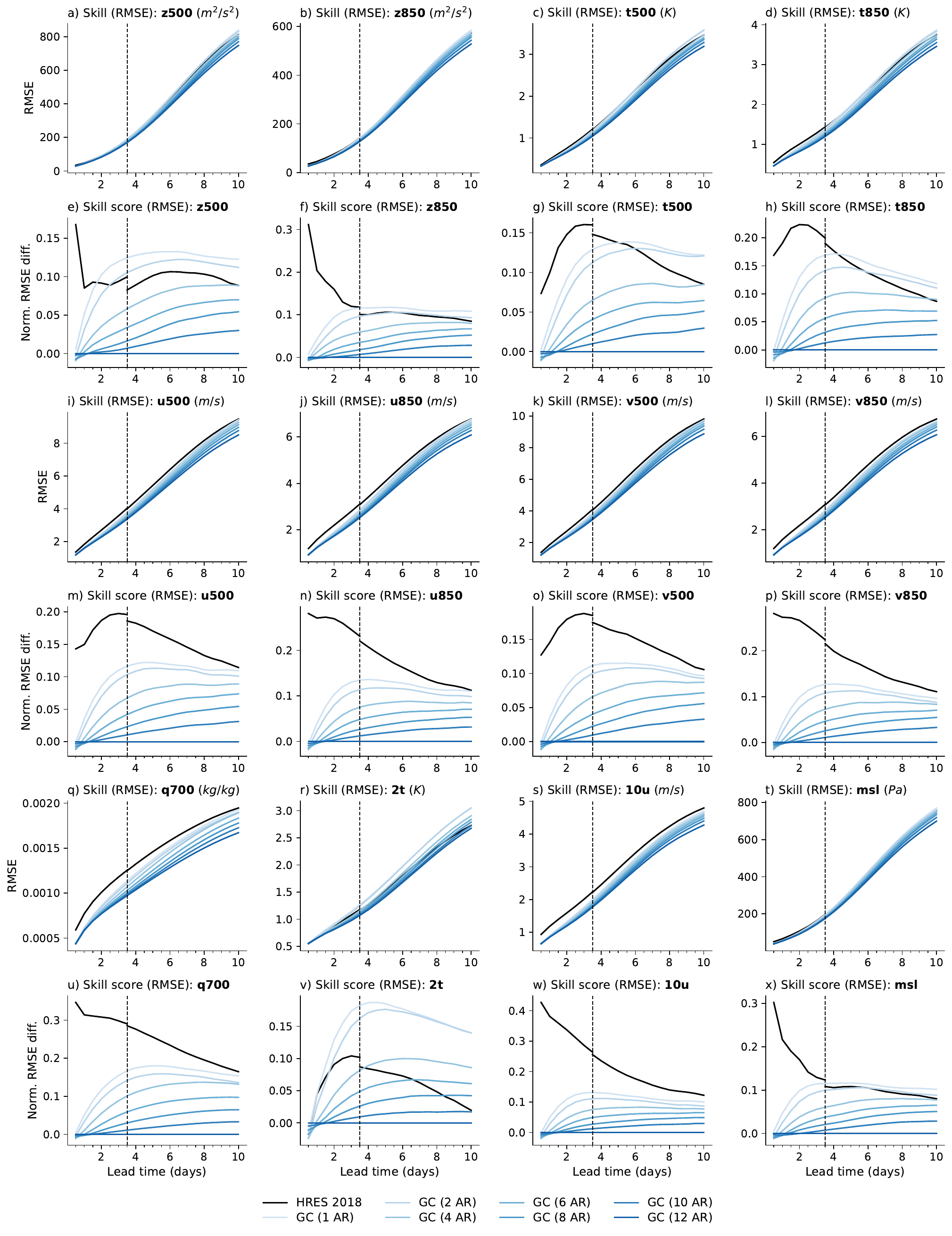}
  \caption{\small\textbf{Effects of autoregressive training.} Each line in the plots represents \ourmodel{}, fine-tuned with different numbers of autoregressive steps, where increasing numbers of steps are represented with darker shades of blue. Rows 1, 3 and 5 show absolute RMSE for \ourmodel{}. Rows 2, 4 and 6 show normalized RMSE differences, with respect to our full 12 autoregressive-step \ourmodel.  Each subplot represents a single variable (and pressure level), as indicated in the subplot titles. The x-axis represents lead time, at 12-hour steps over 10 days. The y-axis represents (absolute or normalized) RMSE. }
  \label{fig:app:resultsar}
\end{figure}

We analyzed the performance of variants of \ourmodel that were trained with fewer autoregressive (AR) steps\footnote{Each of these models were trained using a curriculum where the 1 AR-step model was fine-tuned for 1000 gradient updates each, on increasing numbers of AR steps, from 2-12 (see \cref{sec:app:trainingdetails} and \cref{fig:app:trainingschedule} for details). Each model shown in \cref{fig:app:resultsar} completed its respective number of AR-step training. This means the higher AR-step models had slightly more training than the others, though we do not believe that each had generally converged, so training the lower AR-step models longer likely would not have made much difference.}, which should encourage them to improve their short lead time performance at the expense of longer lead time performance. As shown in \cref{fig:app:resultsar} (with the lighter blue lines corresponding to training with fewer AR steps) we found that models trained with fewer AR steps tended to trade longer for shorter lead time accuracy. These results suggest potential for combining multiple models with varying numbers of AR steps, e.g., for short, medium and long lead times, to capitalize on their respective advantages across the entire forecast horizon.
The connection between number of autoregressive steps and blurring is discussed in Supplements~\cref{sec:app:optimal_filtering_ar_training}.

\FloatBarrier

\subsection{Optimal blurring}
\label{sec:app:optimalfiltering}

\subsubsection{Effect on the comparison of skill between \ourmodel and HRES}

\begin{figure}%
  \centering
  \includegraphics[width=\textwidth]{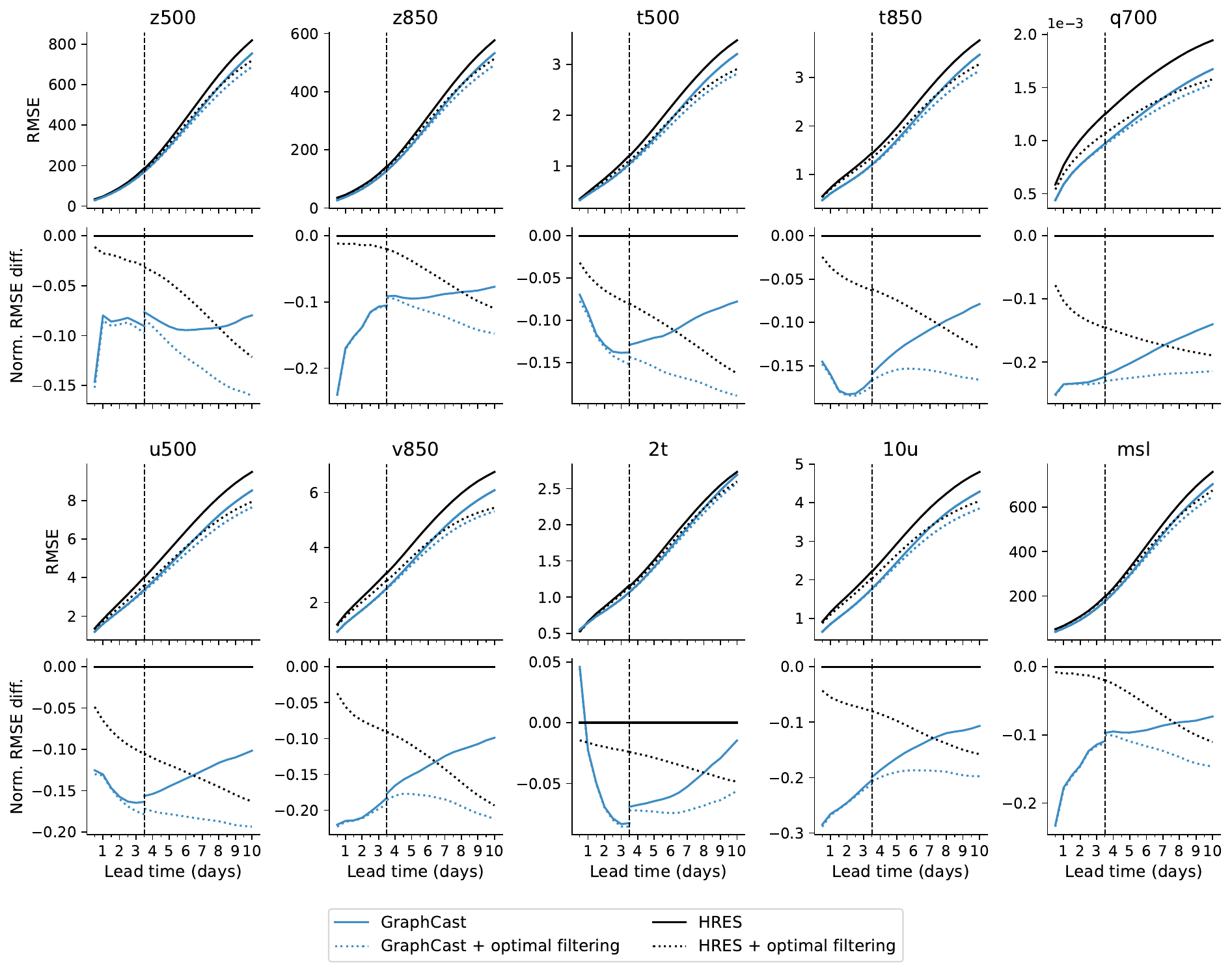}
  \caption{\small\textbf{Effect of optimal filtering on \ourmodel and HRES RMSE skill}. 
  We show RMSEs for unfiltered predictions (solid lines) and optimally filtered predictions (dotted lines) for both \ourmodel and HRES. Rows 1 and 3 show RMSEs, rows 2 and 4 show RMSE skill scores relative to unfiltered HRES.
  RMSEs are computed in the spherical harmonic domain (see \cref{eq:app:rmse_sh}).}
  \label{fig:app:optimalfiltering}
\end{figure}

\begin{figure}%
  \includegraphics[width=\textwidth]{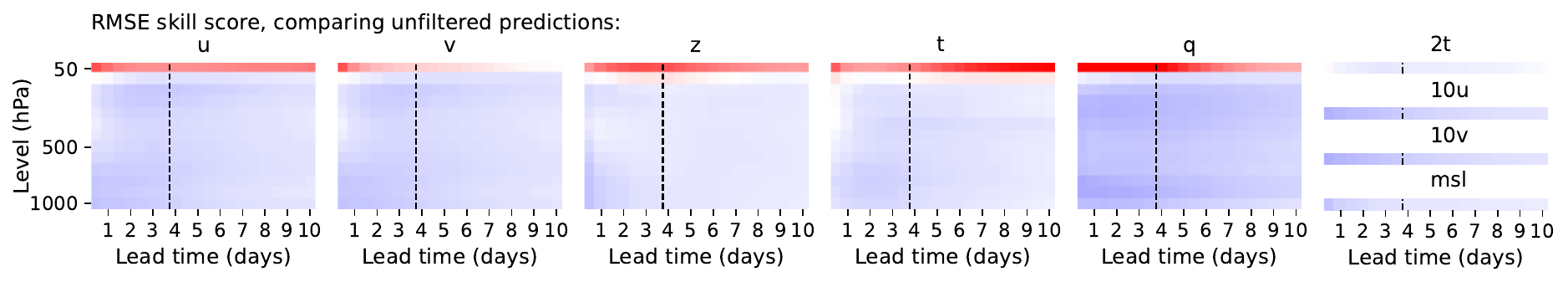}
  \includegraphics[width=\textwidth]{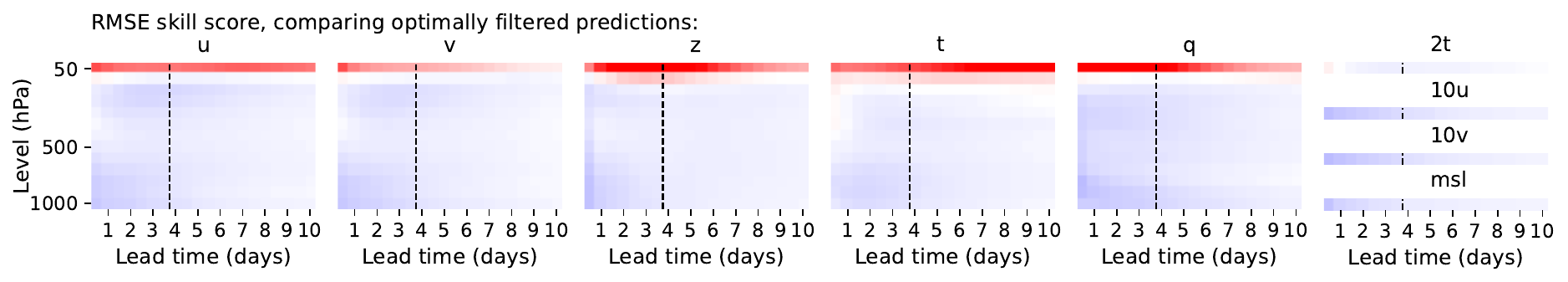}
  \caption{\small\textbf{Effect of optimal filtering on \ourmodel and HRES RMSE scorecards}. We show scorecards (as in \cref{fig:resultshres}) comparing unfiltered predictions, and scorecards comparing optimally filtered predictions. In these scorecards each cell's color represents the RMSE skill score, where blue represents negative values (\ourmodel has better skill) and red represents positive values (HRES has better skill).}
  \label{fig:app:optimalfiltering_scorecards}
\end{figure}

In \cref{fig:app:optimalfiltering,fig:app:optimalfiltering_scorecards} we compare the RMSE of HRES with \ourmodel before and after optimal blurring has been applied to both models. We can see that optimal blurring rarely changes the ranking of the two models, however it does generally narrow the gap between them.

\subsubsection{Filtering methodology}

We chose filters which minimize RMSE within the class of linear, homogeneous (location invariant), isotropic (direction invariant) filters on the sphere. These filters can be applied easily in the spherical harmonic domain, where they correspond to multiplicative filter weights that depend on the total wavenumber, but not the longitudinal wavenumber \cite{devaraju2015understanding}.

For each initialization $\dinit$, lead time $\lt$, variable and level $j$, we applied a discrete spherical harmonic transform \cite{driscoll1994-discrete-sht} to predictions $\xpred^{\dinit+\lt}_{j}$ and targets $\x^{\dinit+\lt}_{j}$, obtaining spherical harmonic coefficients $\hat{f}^{d_0 + \lt}_{j,l,m}$ and $f^{d_0 + \lt}_{j,l,m}$ for each pair of total wavenumber $l$ and longitudinal wavenumber $m$. To resolve the \quarterdegree{} (28km) resolution of our grid at the equator, we use a triangular truncation at total wavenumber 719, which means that $l$ ranges from 0 to $l_{max} = 719$, and for each $l$ the value of $m$ ranges from $-l$ to $l$.

We then multiplied each predicted coefficient $\hat{f}^{d_0 + \lt}_{j,l,m}$ by a filter weight $b^{\lt}_{j,l}$, which is independent of the longitudinal wavenumber $m$. The filter weights were fitted using least-squares to minimize mean squared error, as computed in the spherical harmonic domain:
\begin{align}
  \mathcal{L}_{\text{filters}}^{j,\lt} &=
  \frac{1}{|D_{\text{eval}}|} \sum_{\dinit \in D_{\text{eval}}}
  \frac{1}{4\pi}
  \sum_{l=0}^{l_\text{max}} \sum_{m=-l}^{l}
  {\left(b^{\lt}_{j,l} \hat{f}^{\dinit+\lt}_{j,l,m} - f^{\dinit+\lt}_{j,l,m}\right)}^2.
\label{eq:app:optimal_filter_objective}
\end{align}

We used data from 2017 to fit these weights, which does not overlap with the 2018 test set. When evaluating the filtered predictions, we computed MSE in the spherical harmonic domain, as detailed in \cref{eq:app:rmse_sh}.

By fitting different filters for each lead time, the degree of blurring was free to increase with increasing uncertainty at longer lead times.

While this method is fairly general, it also has limitations. Because the filters are homogeneous, they are unable to take into account location-specific features, such as orography or land-sea boundaries, and so they must choose between over-blurring predictable high-resolution details in these locations, or under-blurring unpredictable high-resolution details more generally. This makes them less effective for some surface variables like \varlevel{2t}, which contain many such predictable details. Future work may consider more complex post-processing schemes.

An alternative way to approximate a conditional expectation (and so improve RMSE) for our ECMWF forecast baseline would be to evaluate the ensemble mean of the ENS ensemble forecast system, instead of the deterministic HRES forecast. However the ENS ensemble is run at lower resolution than HRES, and because of this, it is unclear to us whether its ensemble mean will improve on the RMSE of a post-processed version of HRES. We leave an exploration of this for future work.

\subsubsection{Transfer functions of the optimal filters}

\begin{figure}[ht]
  \centering
  \includegraphics[width=\textwidth]{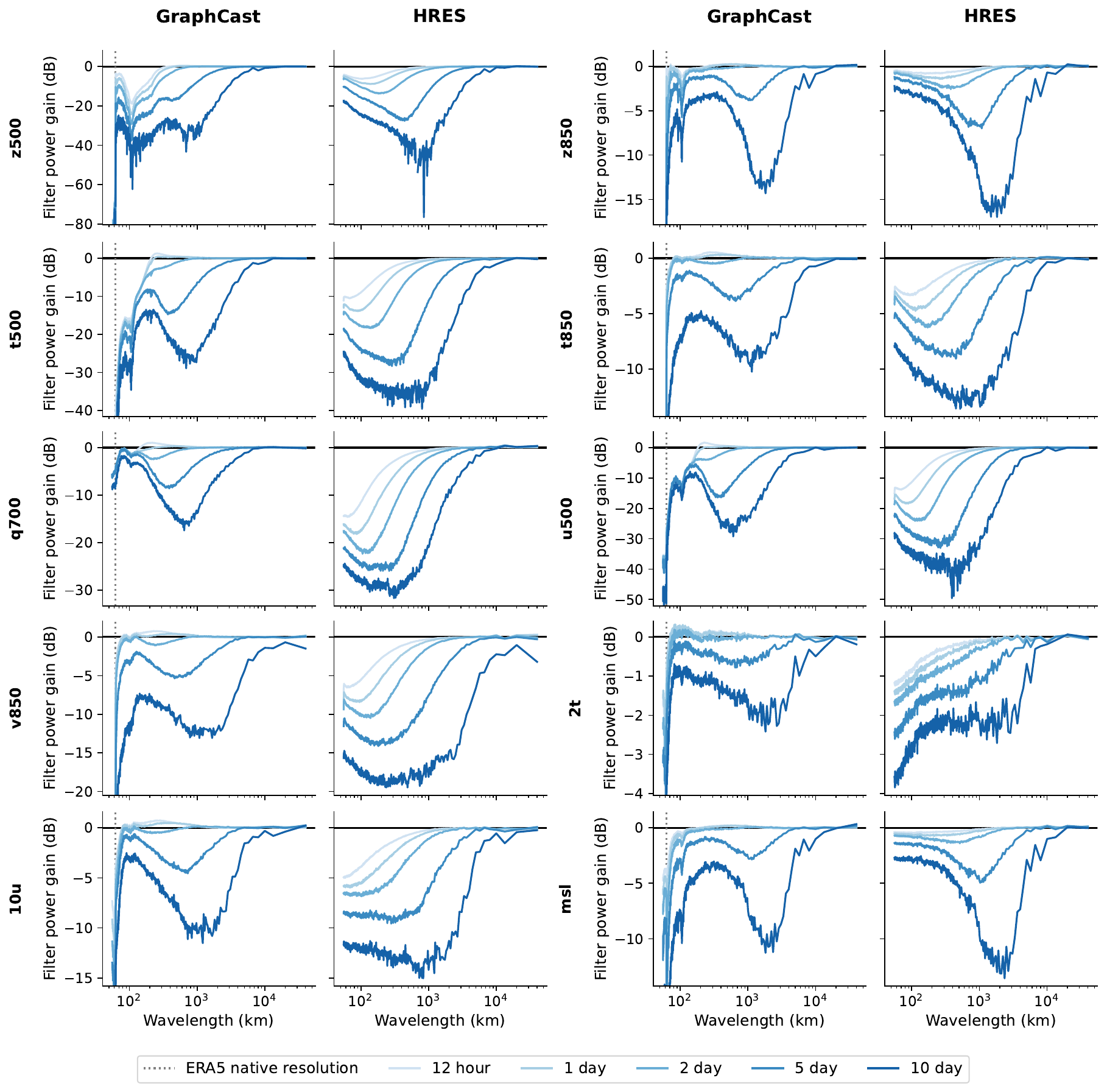}
  \caption{\small\textbf{Transfer functions of optimal filters for \ourmodel and HRES}. The y-axis shows the ratio of output power to input power for the filter, on the logarithmic decibel scale. This is plotted against wavelength on the x-axis. Blue lines correspond to filters fit for different lead times, and the horizontal black line at zero indicates an identity filter response.
  Vertical dotted lines on the \ourmodel plots show the shortest wavelength (62km) resolved at ERA5's native resolution (TL639).}
  \label{fig:app:optimal_filtering_weights}
\end{figure}

The filter weights are visualized in \cref{fig:app:optimal_filtering_weights}, which shows the ratio of output power to input power for the filter, on the logarithmic decibel scale, as a function of wavelength. (With reference to \cref{eq:app:optimal_filter_objective}, this is equal to $20 \log_{10}(b^{\lt}_{j,l})$ for the wavelength $C_e/l$ corresponding to total wavenumber $l$.)

For both HRES and \ourmodel{}, we see that it is optimal for MSE to attenuate power over some short-to-mid  wavelengths. As lead times increase, the amount of attenuation increases, as does the wavelength at which it is greatest. In optimizing for MSE, we seek to approximate a conditional expectation which averages over predictive uncertainty. Over longer lead times this predictive uncertainty increases, as does the spatial scale of uncertainty about the location of weather phenomena. We believe that this largely explains these changes in optimal filter response as a function of lead time.

We can see that HRES generally requires more blurring than \ourmodel, because \ourmodel's predictions already blur to some extent (see \cref{sec:app:spectra}), whereas HRES' do not.

The optimal filters are also able to compensate, to some extent, for spectral biases in the predictions of \ourmodel and HRES. For example, for many variables in our regridded ERA5 dataset, the spectrum cuts off abruptly for wavelengths below 62km that are unresolved at ERA5's native $0.28125^\circ$ resolution. \ourmodel has not learned to replicate this cutoff exactly, but the optimal filters are able to implement it.

We also note that there are noticeable peaks in the \ourmodel filter response around 100km wavelength for \varlevel{z}{500}, which are not present for HRES. We believe these are filtering out small, spurious artifacts which are introduced by \ourmodel around these wavelengths as a side-effect of the grid-to-mesh and mesh-to-grid transformations performed inside the model.

\FloatBarrier

\subsubsection{Relationship between autoregressive training horizon and blurring}
\label{sec:app:optimal_filtering_ar_training}

\begin{figure}[ht]
  \centering
  \includegraphics[width=.93\textwidth]{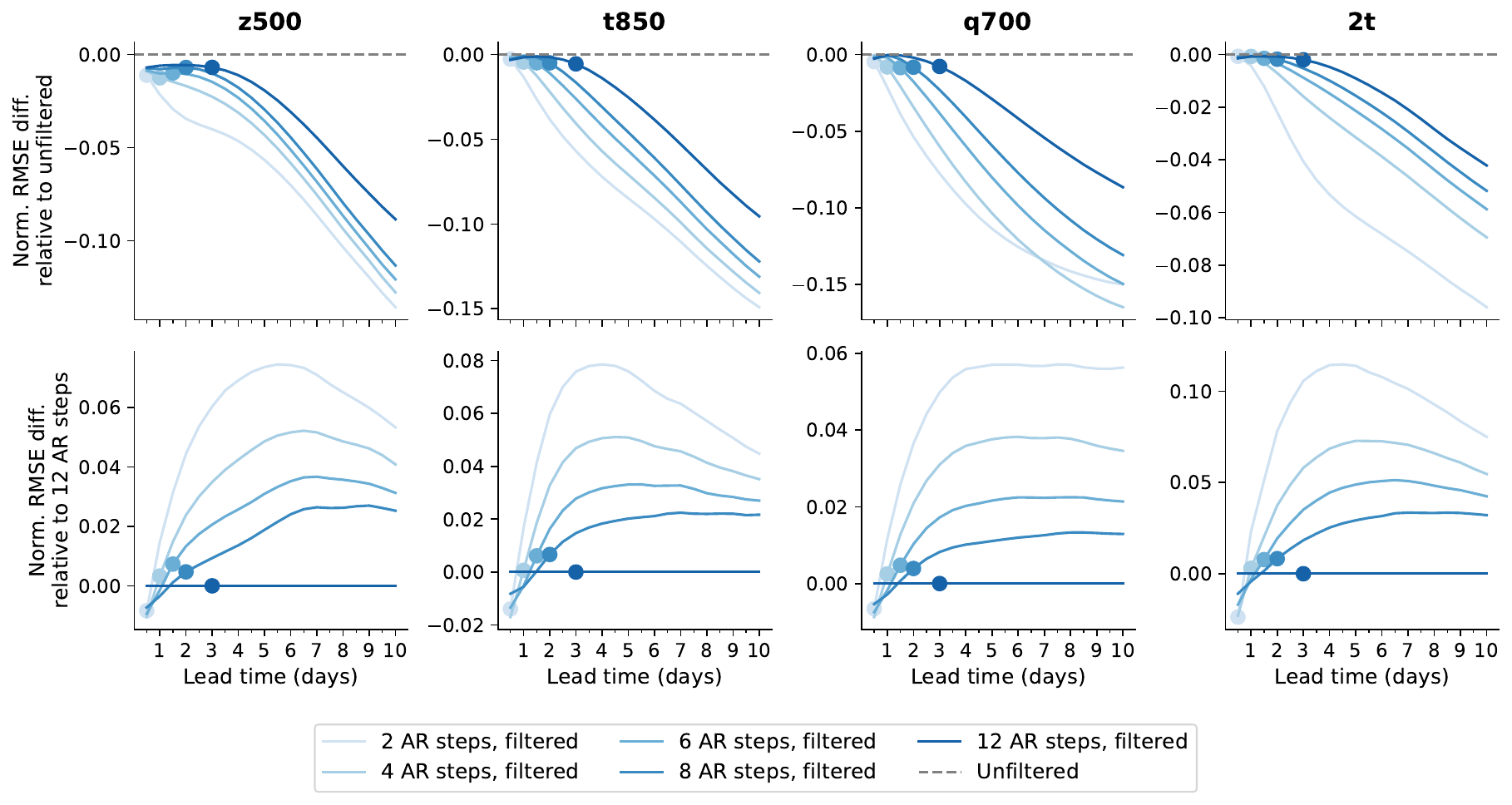}
  \caption{\small\textbf{Results of optimal filtering, for \ourmodel trained to different autoregressive training horizons}.
  In the first row we plot the RMSE of filtered predictions relative to corresponding unfiltered predictions.
  In the second row we plot the RMSE of filtered predictions relative to the filtered predictions trained to 12 autoregressive steps.
  Circles show the lead time equivalent to the autoregressive training horizon which each model was trained up to.} 
  \label{fig:app:optimal_filtering_ar_steps}
\end{figure}

In \cref{fig:app:optimal_filtering_ar_steps} we use the results of optimal blurring to investigate the connection between autoregressive training and the blurring of \ourmodel's predictions at longer lead times.

In the first row of \cref{fig:app:optimal_filtering_ar_steps}, we see that models trained with longer autoregressive training horizons benefit less from optimal blurring, and that the benefits of optimal blurring generally start to accrue only \emph{after} the lead time corresponding to the horizon they were trained up to. This suggests that autoregressive training is effective in teaching the model to blur optimally up to the training horizon, but beyond this further blurring is required to minimize RMSE.

It would be convenient if we could replace longer-horizon training with a simple post-processing strategy like optimal blurring, but this does not appear to be the case: in the second row of \cref{fig:app:optimal_filtering_ar_steps} we see that longer-horizon autoregressive training still results in lower RMSEs, even after optimal blurring has been applied.

If one desires predictions which are in some sense minimally blurry, one could use a model trained to a small number of autoregressive steps. This would of course result in higher RMSEs at longer lead times, and our results here suggest that these higher RMSEs would not \emph{only} be due to the lack of blurring; one would be compromising on other aspects of skill at longer lead times too. In some applications this may still be a worthwhile trade-off, however.

\FloatBarrier

\subsection{Spectral analysis}\label{sec:app:spectralanalysis}

\subsubsection{Spectral decomposition of mean squared error}
\label{sec:app:spectral_decomp_errors}

\begin{figure}[ht]
  \centering
    \includegraphics[width=0.92\textwidth]{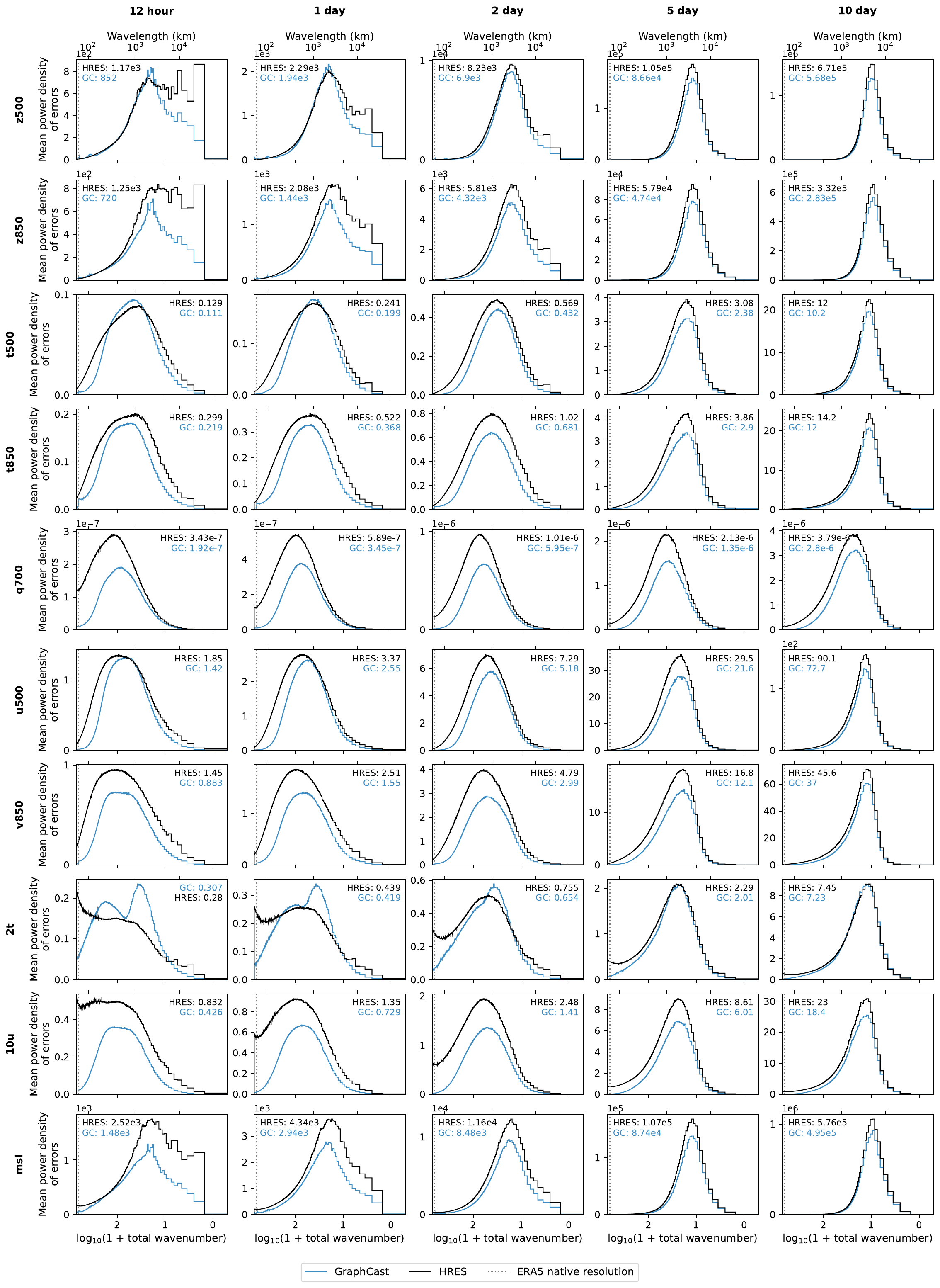}
  \caption{\small\textbf{Spectral decomposition of mean squared error for \ourmodel and HRES.} We plot histogram densities with respect to the $\log_{10}(1+\text{total wavenumber})$ x-axis, so the total area under the curve corresponds to the MSE, which is also indicated in the corner of each plot. Dotted vertical lines indicate the native resolution of \ourmodel's ERA5 training data.}
  \label{fig:app:spectral_decomp_errors}
\end{figure}

\begin{figure}[ht]
  \centering
    \includegraphics[width=0.92\textwidth]{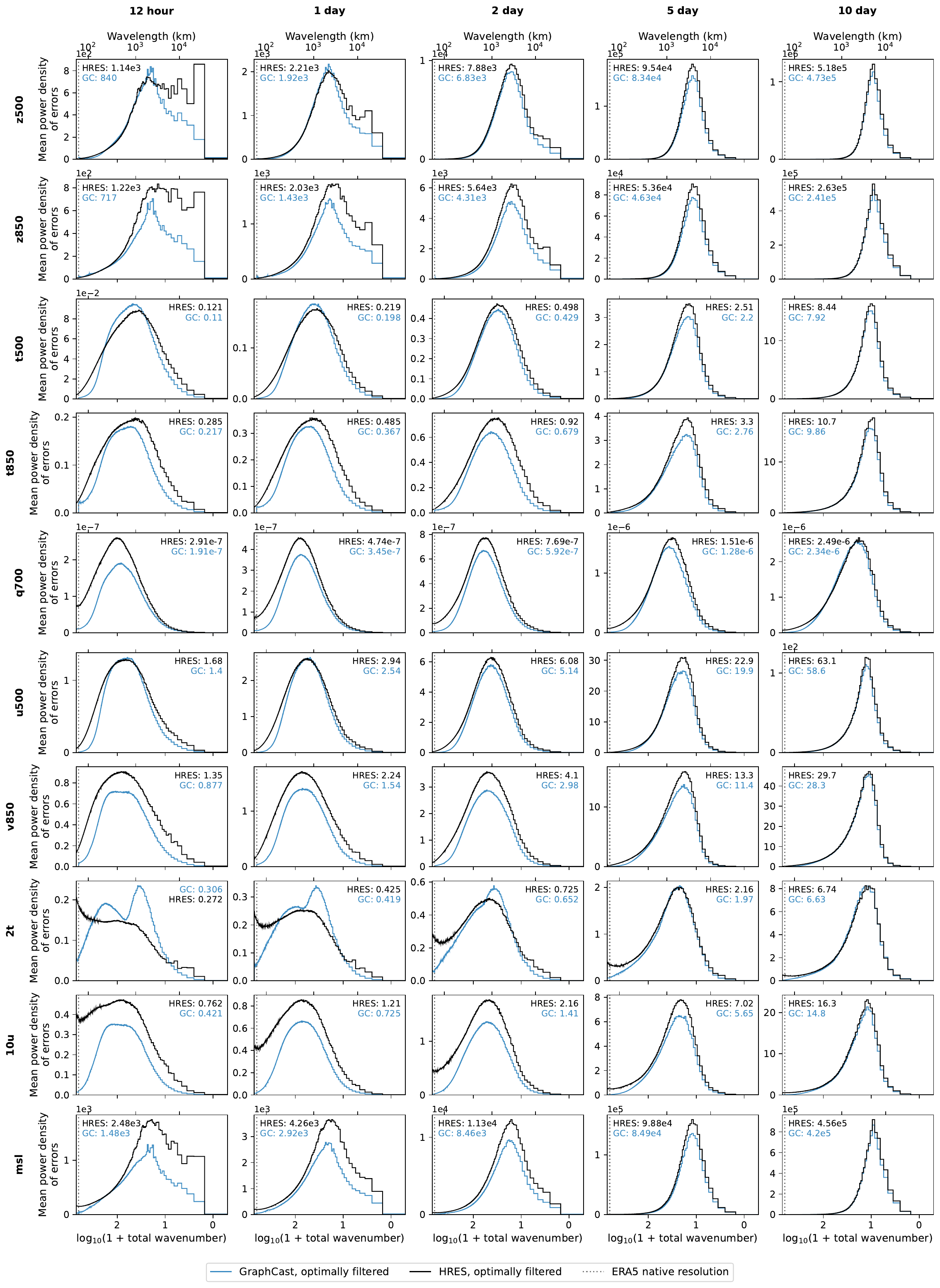}
  \caption{\small\textbf{Spectral decomposition of mean squared error for \ourmodel and HRES after optimal blurring.} We plot histogram densities with respect to the $\log_{10}(1+\text{total wavenumber})$ x-axis, so the total area under the curve corresponds to the MSE, which is also indicated in the corner of each plot. Dotted vertical lines indicate the native resolution of \ourmodel's ERA5 training data.}
  \label{fig:app:spectral_decomp_errors_optimal_filtering}
\end{figure}

In \cref{fig:app:spectral_decomp_errors,fig:app:spectral_decomp_errors_optimal_filtering} we compare the skill of \ourmodel with HRES over a range of spatial scales, before and after optimal filtering (see details in~\cref{sec:app:optimalfiltering}). The MSE, via its spectral formulation  (\cref{eq:app:rmse_sh}) can be decomposed as a sum of mean error powers at different total wavenumbers:
\begin{align}
  \text{MSE}_{sh}(j,\lt) &=
  \sum_{l=0}^{l_\text{max}} S^{j,\lt}(l)
  \\
  S^{j,\lt}(l) &=
  \frac{1}{|D_{\text{eval}}|} \sum_{\dinit \in D_{\text{eval}}}
  \frac{1}{4\pi}
   \sum_{m=-l}^{l}
  {\left(\hat{f}^{\dinit+\lt}_{j,l,m} - f^{\dinit+\lt}_{j,l,m}\right)}^2,
\label{eq:app:mse_decomp}
\end{align}
where $l_\text{max}=719$ as in \cref{eq:app:rmse_sh}.
Each total wavenumber $l$ corresponds approximately to a wavelength $C_e / l$, where $C_e$ is the earth's circumference.

We plot power density histograms, where the area of each bar corresponds to $S^{j,\lt}(l)$, and the bars center around $\log_{10}(1+l)$ (since a log frequency scale allows for easier visual inspection, but we must also include wavenumber $l=0$). In these plots, the total area under the curve is the MSE.

At lead times of 2 days or more, for the majority of variables \ourmodel improves on the skill of HRES uniformly over all wavelengths. (2m temperature is a notable exception).

At shorter lead times of 12 hours to 1 day, for a number of variables (including \varlevel{z}500, \varlevel{t}500, \varlevel{t}850 and \varlevel{u}500) HRES has greater skill than \ourmodel at scales in the approximate range of 200-2000km, with \ourmodel generally having greater skill outside this range.


\FloatBarrier

\subsubsection{RMSE as a function of horizontal resolution}

\begin{figure}[ht]
  \centering
  \includegraphics[width=.89\textwidth]{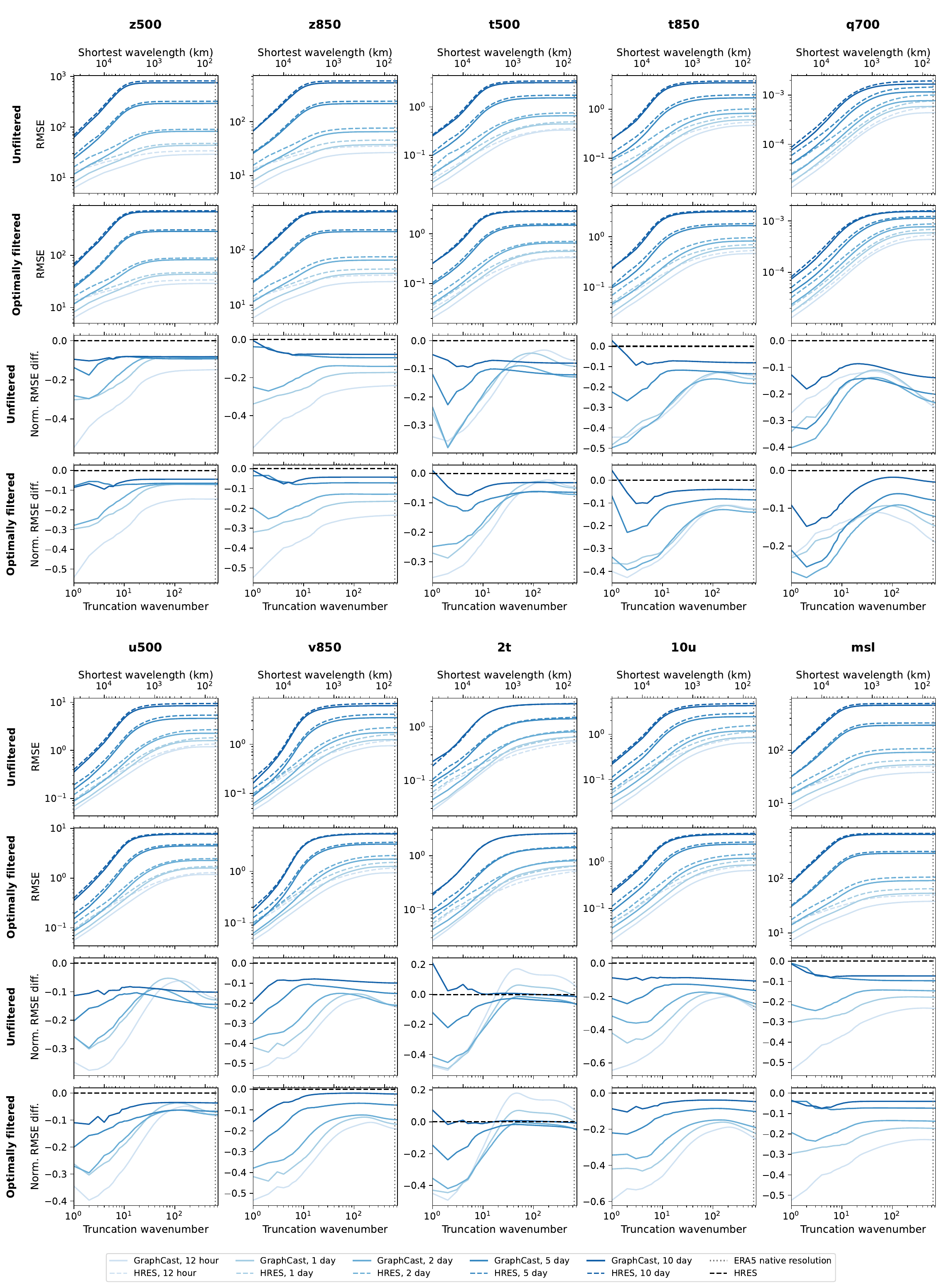}
  \caption{\small\textbf{RMSE as a function of horizontal resolution}. The x-axes show the total wavenumber and wavelength at which both predictions and targets were truncated. The y-axes show RMSEs (rows 1,2,5,6), or the ratio between \ourmodel and HRES RMSEs (rows 3,4,7,8). We give results before optimal blurring (rows 1,3,5,7) and after (rows 2,4,6,8).}
  \label{fig:app:rmse_by_resolution}
\end{figure}

In \cref{fig:app:rmse_by_resolution}, we compare the skill of \ourmodel with HRES when evaluated at a range of spatial resolutions. Specifically, at each total wavenumber $l_{\text{trunc}}$, we plot RMSEs between predictions and targets which are both truncated at that total wavenumber. This is approximately equivalent to a wavelength $C_e / l_{\text{trunc}}$ where $C_e$ is the earth's circumference.

The RMSEs between truncated predictions and targets can be obtained via cumulative sums of the mean error powers $S^{j,\lt}(l)$ defined in~\cref{eq:app:mse_decomp}, according to
\begin{align}
  \text{RMSE}_{\text{trunc}}(j,\lt,l_{\text{trunc}}) &=
  \sqrt{\sum_{l=0}^{l_{\text{trunc}}} S^{j,\lt}(l)}.
\label{eq:app:mse_trunc}
\end{align}

\cref{fig:app:rmse_by_resolution} shows that in most cases \ourmodel has lower RMSE than HRES at all resolutions typically used for forecast verification. This applies before and after optimal filtering (see \cref{sec:app:optimalfiltering}). Exceptions include 2 meter temperature at a number of lead times and resolutions, \varlevel{t500} at 12 hour lead times, and \varlevel{u500} at 12 hour lead times, where \ourmodel does better at \quarterdegree resolution but HRES does better at resolutions around $0.5^\circ$ to $2.5^\circ$ (corresponding to shortest wavelengths of around 100 to 500 km).

In particular we note that the native resolution of ERA5 is $0.28125^\circ$ corresponding to a shortest wavelength of 62km, indicated by a vertical line in the plots. \hresfczero targets contain some signal at wavelengths shorter than 62km, but the ERA5 targets used to evaluate \ourmodel do not, natively at least (see \cref{sec:app:spectra}). In \cref{fig:app:rmse_by_resolution} we can see that evaluating at $0.28125^\circ$ resolution instead of \quarterdegree does not significantly affect the comparison of skill between \ourmodel and HRES.

\FloatBarrier

\subsubsection{Spectra of predictions and targets}
\label{sec:app:spectra}

\begin{figure}[ht]
  \centering
  \includegraphics[width=\textwidth]{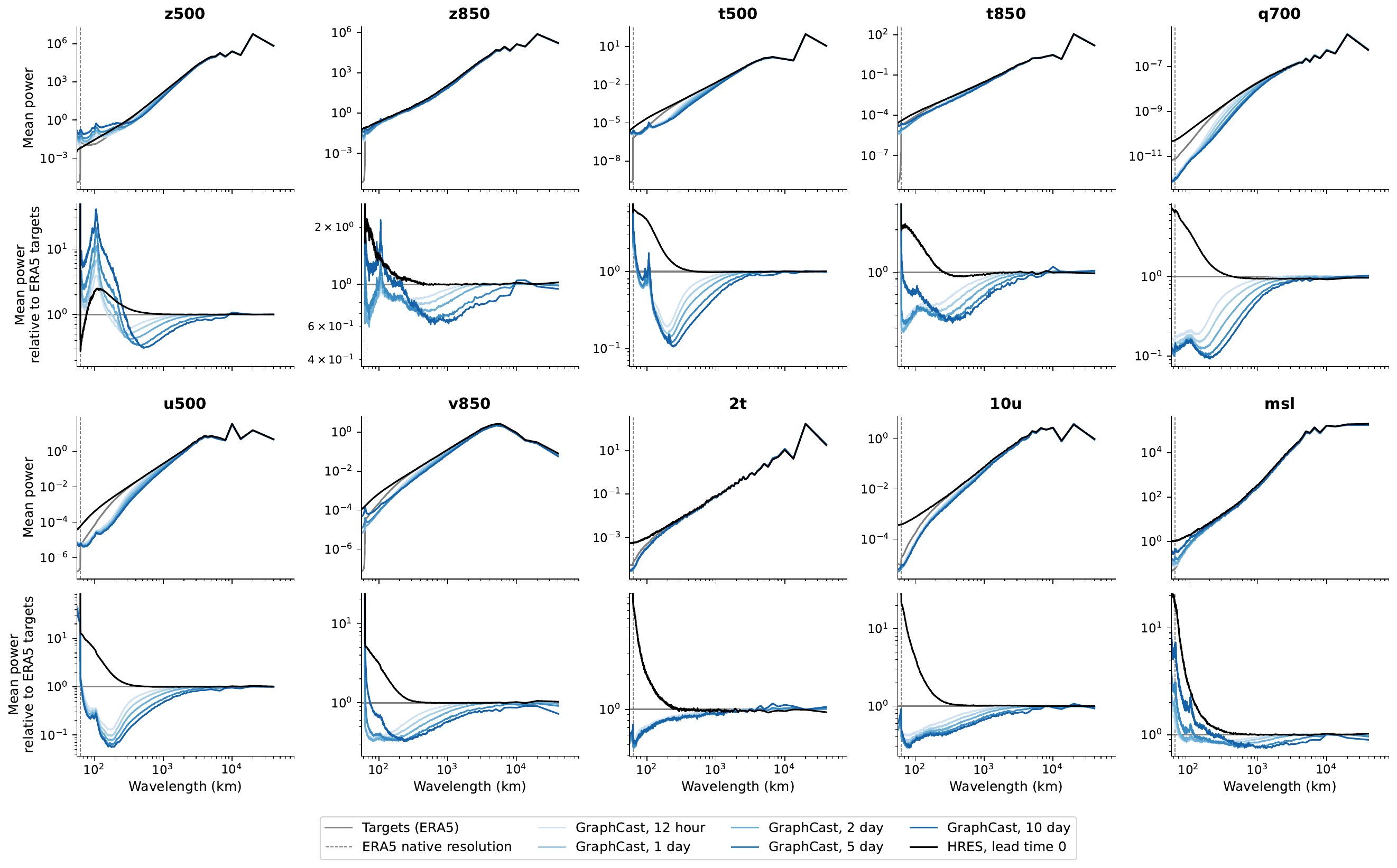}
  \caption{\small\textbf{Power spectra of predictions and targets for \ourmodel}.
  For each variable, the first row plots power at each total wavenumber on a log-log scale; the second row plots power relative to the ERA5 targets used for \ourmodel. We also show the spectrum of HRES in black.
  }
  \label{fig:app:spectra}
\end{figure}

\cref{fig:app:spectra} compares the power spectra of \ourmodel's predictions, the ERA5 targets they were trained against, and \hresfczero. A few phenomena are notable:
\paragraph{Differences between HRES and ERA5}

There are noticeable differences in the spectra of ERA5 and \hresfczero, especially at short wavelengths. These differences may in part be caused by the methods used to regrid them from their respective native IFS resolutions of TL639 ($0.28125^\circ$) and TCo1279 (approx. $0.1^\circ$, \cite{a-new-grid-for-the-ifs}) to a \quarterdegree equiangular grid. However even before this regridding is done there are differences in IFS versions, settings, resolution and data assimilation methodology used for HRES and ERA5, and these differences may also affect the spectra. Since we evaluate \ourmodel against ERA5 and HRES against \hresfczero, this domain gap remains an important caveat to attach to our conclusions.

\paragraph{Blurring in \ourmodel}

We see reduced power at short-to-mid wavelengths in \ourmodel's predictions which reduces further with lead time. We believe this corresponds to blurring which \ourmodel has learned to perform in optimizing for MSE. We discussed this further in \cref{sec:app:optimalfiltering,sec:app:optimal_filtering_ar_training}.

\paragraph{Peaks for \ourmodel around 100km wavelengths}

These peaks are particularly visible for \varlevel{z}{500}; they appear to increase with lead time. We believe they correspond to small, spurious artifacts introduced by the internal grid-to-mesh and mesh-to-grid transformations performed by \ourmodel at each autoregressive step. In future work we hope to eliminate or reduce the effect of these artifacts, which were also observed by \cite{keisler2022forecasting}.

Finally we note that, while these differences in power at short wavelengths are very noticeable in log scale and relative plots, these short wavelengths contribute little to the total power of the signal.

\newpage
\section{Additional severe event forecasting results} \label{app:severe_event_forecasting}

In this section, we provide additional details about our severe event forecasting analysis.
We note that \ourmodel is not specifically trained for those downstream tasks, which demonstrates that, beyond improved skills, \ourmodel provides useful forecast for tasks with real-world impact such as tracking cyclones (\cref{app:cyclones}), characterizing atmospheric rivers (\cref{sec:app:atmosphericriver}), and classifying extreme temperature (\cref{sec:app:extremetemperature}).
Each task can also be seen as evaluating the value of \ourmodel on a different axis: spatial and temporal structure of high-resolution prediction (cyclone tracking task), ability to non-linearly combine \ourmodel predictions to derive quantities of interest (atmospheric rivers task), and ability to characterize extreme and rare events (extreme temperatures). 

\subsection{Tropical cyclone track forecasting}\label{app:cyclones}

In this section, we detail the evaluation protocols we used for cyclone tracking (Supplements~\cref{sec:app:cyclone:eval_protocol}) and analyzing statistical significance (Supplements~\cref{sec:app:cyclone:statistical_methodology}), provide additional results (Supplements~\cref{sec:app:cyclone:complementary_results}), and describe our tracker and its differences with the one from ECMWF (Supplements~\cref{sec:app:cyclone:tracker}).

\subsubsection{Evaluation protocol}\label{sec:app:cyclone:eval_protocol}

The standard way of comparing two tropical cyclone prediction systems is to restrict the comparison to events where both models predict the existence of a cyclone.
As detailed in Supplements~\cref{sec:app:ensuring_equal_lookahead}, \ourmodel is initialized from 06z and 18z, rather than 00z and 12z, to avoid giving it a lookahead advantage over HRES.
However, the HRES cyclone tracks in the TIGGE archive~\cite{bougeault2010thorpex} are only initialized at 00z and 12z.
This discrepancy prevents us from selecting events where the initialization and lead time map to the same validity time for both methods, as there is always a 6h mismatch.
Instead, to compare HRES and \ourmodel on a set of similar events, we proceed as follows.
We consider all the dates and times for which our ground truth dataset IBTrACS~\cite{knapp2010international,ibtracs-dataset-2018} identified the presence of a cyclone. For each cyclone, if its time is 06z or 18z, we make a prediction with \ourmodel starting from that date, apply our tracker and keep all the lead times for which our tracker detects a cyclone.
Then, for each initialization time/lead time pairs kept for \ourmodel, we consider the two valid times at +/-6\unit{h} around the initialization time of \ourmodel, and use those as initialization time to pick the corresponding HRES track from the TIGGE archive. If, for the same lead time as \ourmodel, HRES detects a cyclone, we include both \ourmodel and HRES initialization time/lead time pairs into the final set of events we use to compare them.
For both methods, we only consider predictions up to 120 hours.

Because we compute error with respect to the same ground truth (i.e., IBTrACS), the evaluation is not subject to the same restrictions described in Supplements~\cref{sec:app:ensuring_equal_lookahead}, i.e., the targets for both models incorporate the same amount of lookahead. This is in contrast with most our evaluations in this paper, where the targets for HRES (i.e., \hresfczero) incorporates +3h lookahead, and the ones for \ourmodel (from ERA5) incorporate +3h or +9h, leading us to only report results for the lead times with a matching lookahead (multiples of 12h). Here, since the IBTrACS targets are the same for both models, we can report performance as a function of lead time by increments of 6\unit{h}.

For a given forecast, the error between the predicted center of the cyclone and the true center is computed using the geodesic distance.

\subsubsection{Statistical methodology}\label{sec:app:cyclone:statistical_methodology}

Computing statistical confidence in cyclone tracking requires particular attention in two aspects:
\begin{enumerate}
    \item There are two ways to define the number of samples. The first one is the number of tropical cyclone events, which can be assumed to be mostly independent events. The second one is the number of per-lead time data points used, which is larger, but accounts for correlated points (for each tropical cyclone event multiple predictions are made at 6\unit{h} interval). We chose to use the first definition which provides more conservative estimates of statistical significance. Both numbers are shown for lead times 1 to 5 days on the x-axis of Supplements~\cref{fig:app:cyclone_all_cat_track_error_and_paired_mean}.
    \item The per-example tracking errors of HRES and \ourmodel are correlated. Therefore statistical variance in their difference is much smaller than their joint variance. Thus, we report the confidence that \ourmodel is better than HRES (see Supplements~\cref{fig:app:cyclone_all_cat_track_error_and_paired_mean}b) in addition to the per-model confidence (see Supplements~\cref{fig:app:cyclone_all_cat_track_error_and_paired_mean}a).
\end{enumerate}

Given the two considerations above, we do bootstrapping with $95\%$ confidence intervals at the level of cyclones.
For a given lead time, we consider all the corresponding initialization time/lead time pairs and keep a list of which cyclone they come from (without duplication).
For the bootstrap estimate, we draw samples from this cyclone list (with replacement) and apply the median (or the mean) to the corresponding initialization time/lead time pairs.
Note that this gives us much more conservative confidence bounds than doing bootstrapping at the level of initialization time/lead time pairs, as it is equivalent to assuming all bootstrap samples coming from the sample cyclone (usually in the order of tens) are perfectly correlated.

For instance, assume for a given lead time we have errors of (50, 100, 150) for cyclone A, (300, 200) for cyclone B and (100, 100) for cyclone C, with A having more samples. A bootstrapping sample at the level of cyclones first samples uniformly at random 3 cyclones with replacement (for instance A,A,B) and then computes the mean on top of the corresponding samples with multiplicity: mean(50,100,150,50,100,150,200,300)=137.5.

\subsubsection{Results}\label{sec:app:cyclone:complementary_results}

In Supplements~\cref{fig:resultsextremes}a-b, we chose to show the median error rather than the mean.
This decision was made before computing the results on the test set, based on the performance on the validation set.
On the years 2016--2017, using the version of \ourmodel trained on 1979--2015, we observed that, using early versions of our tracker, the mean track error was dominated by very few outliers and was not representative of the overall population. Furthermore, a sizable fraction of these outliers were due to errors in the tracking algorithm rather than the predictions themselves, suggesting that the tracker was suboptimal for use with \ourmodel. Because our goal is to assess the value of \ourmodel forecast, rather than a specific tracker, we show median values, which are also affected by tracking errors, but to a lesser extent. In figure \cref{fig:app:cyclone_histograms} we show how that the distribution of both HRES and GraphCast track errors for the test years 2018--2021 are non-gaussian with many outliers. This suggests the median is a better summary statistic than the mean.

\begin{figure}%
  \centering
  \includegraphics[width=\textwidth]{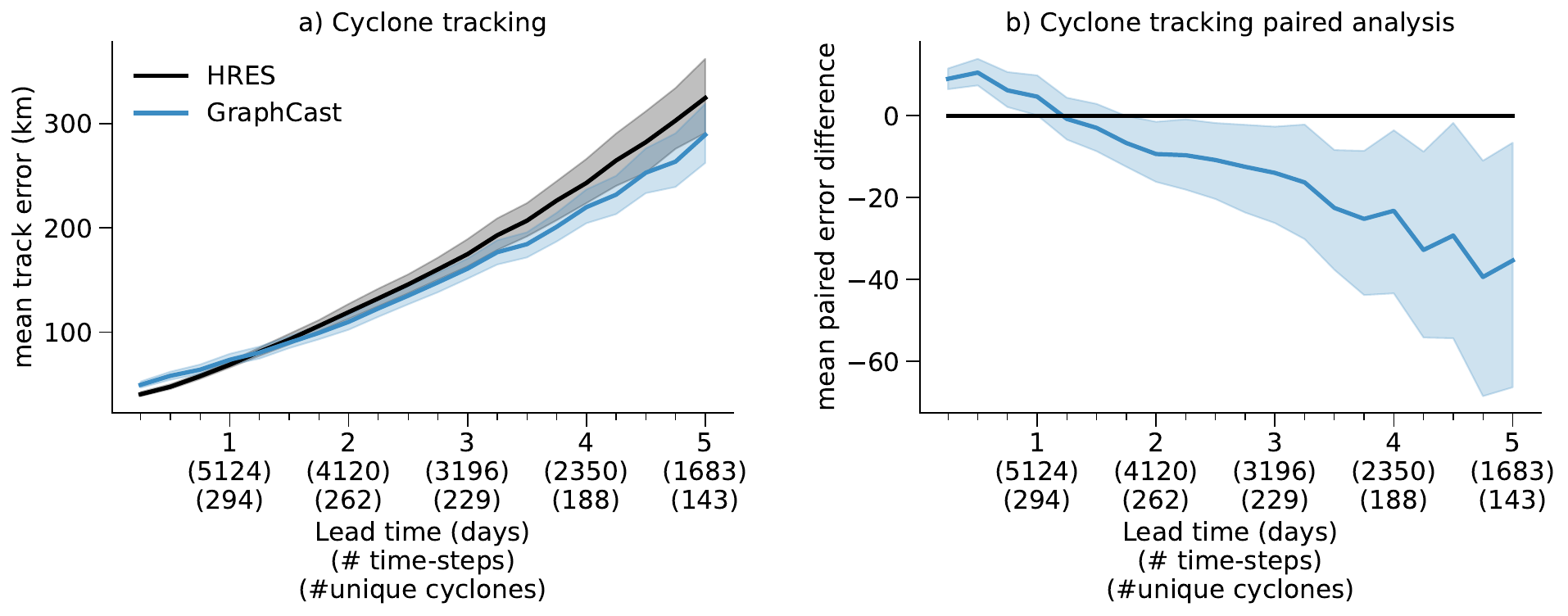}
\caption{\small\textbf{Mean performance on cyclone tracking \textit{(lower is better)}} a) Cyclone forecasting performances for \ourmodel{} and HRES. The x-axis represents lead times (in days). The y-axis represents mean track error (in km). The error bars represent the bootstrapped error of the mean. b) Paired analysis of cyclone forecasting. The x-axis represents lead times (in days). The y-axis represents mean per-track error difference between HRES and \ourmodel. The error bars represent the bootstrapped error of the mean.}
  \label{fig:app:cyclone_all_cat_track_error_and_paired_mean}
\end{figure}
Supplements~\cref{fig:app:cyclone_all_cat_track_error_and_paired_mean} complements \cref{fig:resultsextremes}a-b by showing the mean track error and the corresponding paired analysis.
We note that using the final version of our tracker (Supplements~\cref{sec:app:cyclone:tracker}), \ourmodel mean results are similar to the median one, with \ourmodel significantly outperforming HRES for lead time between 2 and 5 days.

\begin{figure}%
  \centering
  \includegraphics[width=\textwidth]{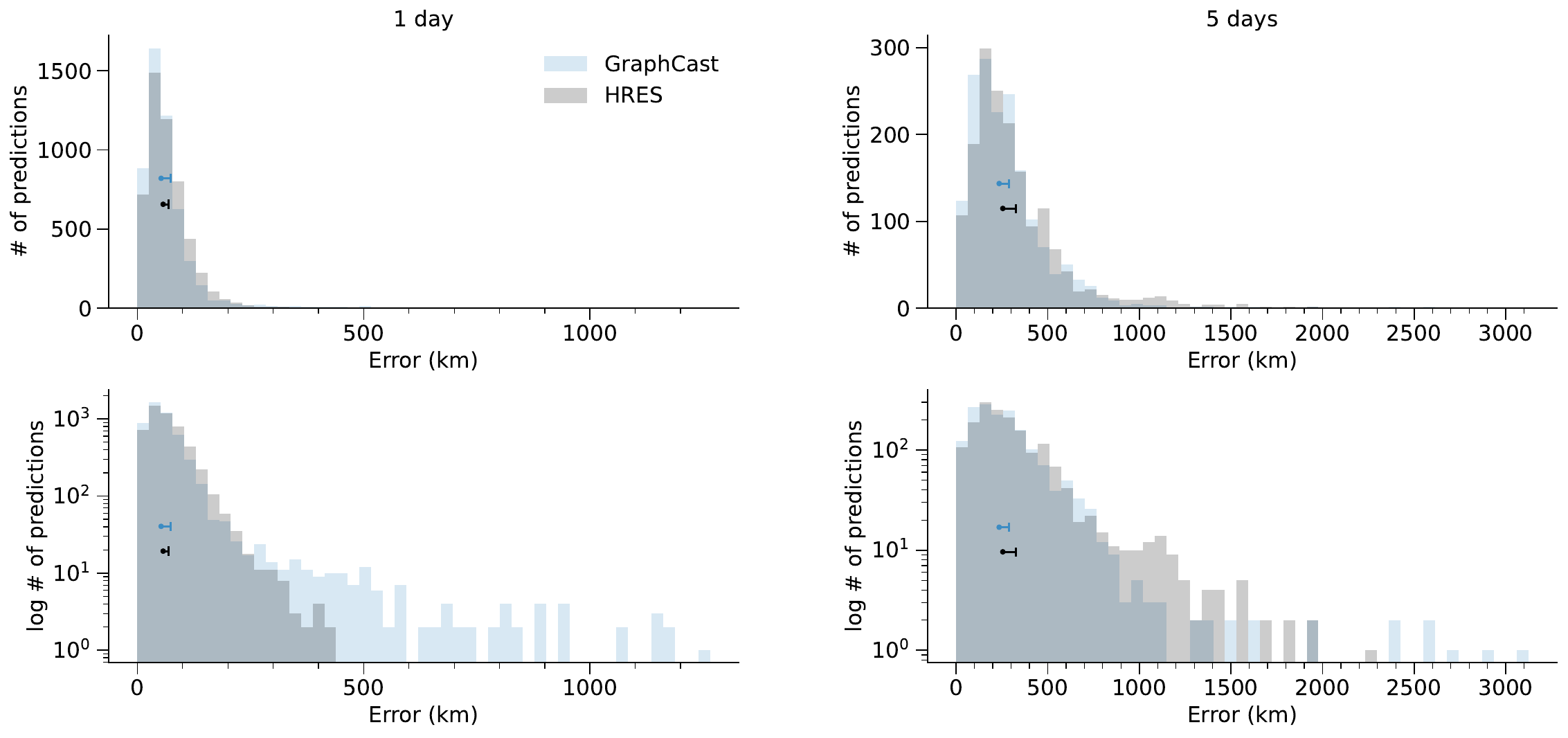}
\caption{\small\textbf{Histograms of cyclone track errors} with linear and logarithmic y-axis. The horizontal lines connect the median error (circle) to the mean error (vertical tick) for each model. We observe that the distribution of errors of HRES, and particularly GraphCast, are not gaussian and instead have some very big outliers.}
  \label{fig:app:cyclone_histograms}
\end{figure}

Because of well-known blurring effects, which tend to smooth the extrema used by a tracker to detect the presence of a cyclone, ML methods can drop existing cyclones more often than NWPs. 
Dropping a cyclone is very correlated with having a large positional error.
Therefore, removing from the evaluation such predictions, where a ML model would have performed particularly poorly, could give it an unfair advantage.

\begin{figure}%
  \centering
  \includegraphics[width=.7\textwidth]{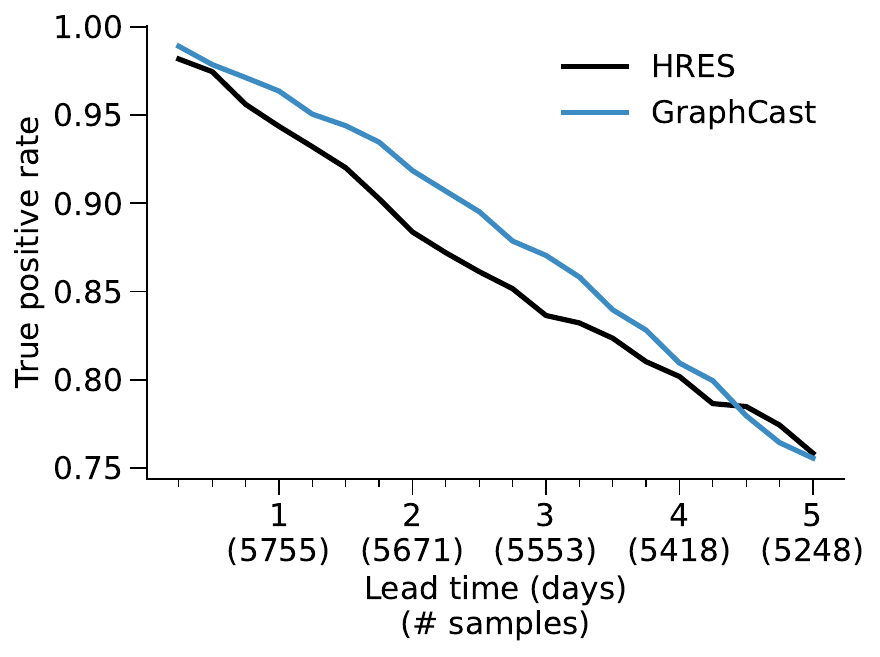}
\caption{\small\textbf{True positive rate detection of cyclones \textit{(higher is better)}} \ourmodel{} and HRES detect a comparable number of cyclones, decreasing as a function of lead time.}
  \label{fig:app:cyclone_truepos}
\end{figure}

To avoid this issue, we verify that our hyper-parameter-searched tracker (see Supplements~\cref{sec:app:cyclone:tracker}) misses a similar number of cyclones as HRES. Supplements~\cref{fig:app:cyclone_truepos} shows that on the test set (2018--2021), \ourmodel and HRES drop a similar number of cyclones, ensuring our comparisons are as fair as possible.

\begin{figure}%
  \centering
  \includegraphics[width=\textwidth]{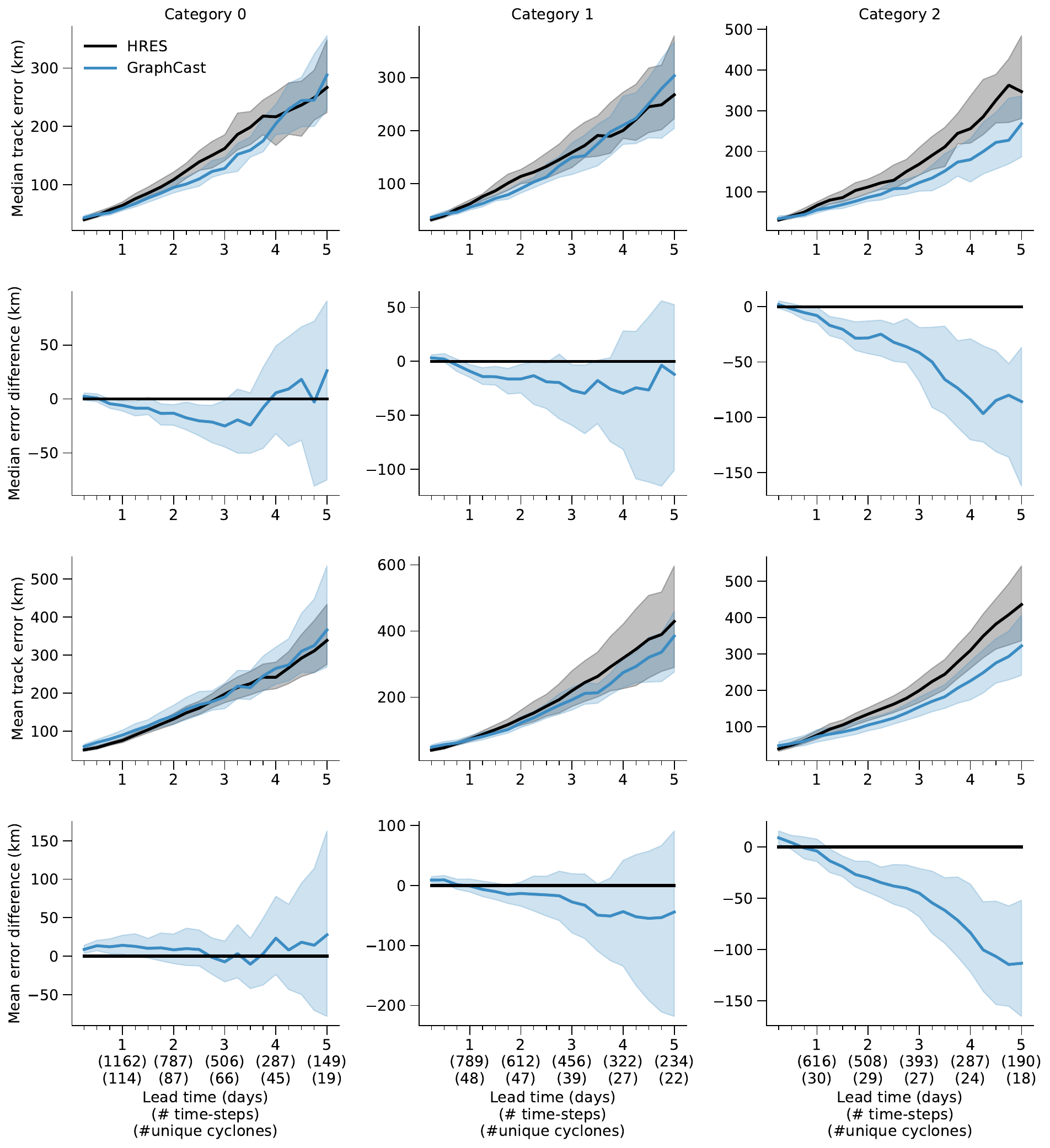}
\caption{\small\textbf{Per-cyclone-category median and mean performance (category 0 to 2)} Each column corresponds to a cyclone category from 0 to 2 on the Saffir-Simpson Hurricane Wind Scale.}
  \label{fig:app:categories_0_1_2}
\end{figure}

\begin{figure}%
  \centering
  \includegraphics[width=\textwidth]{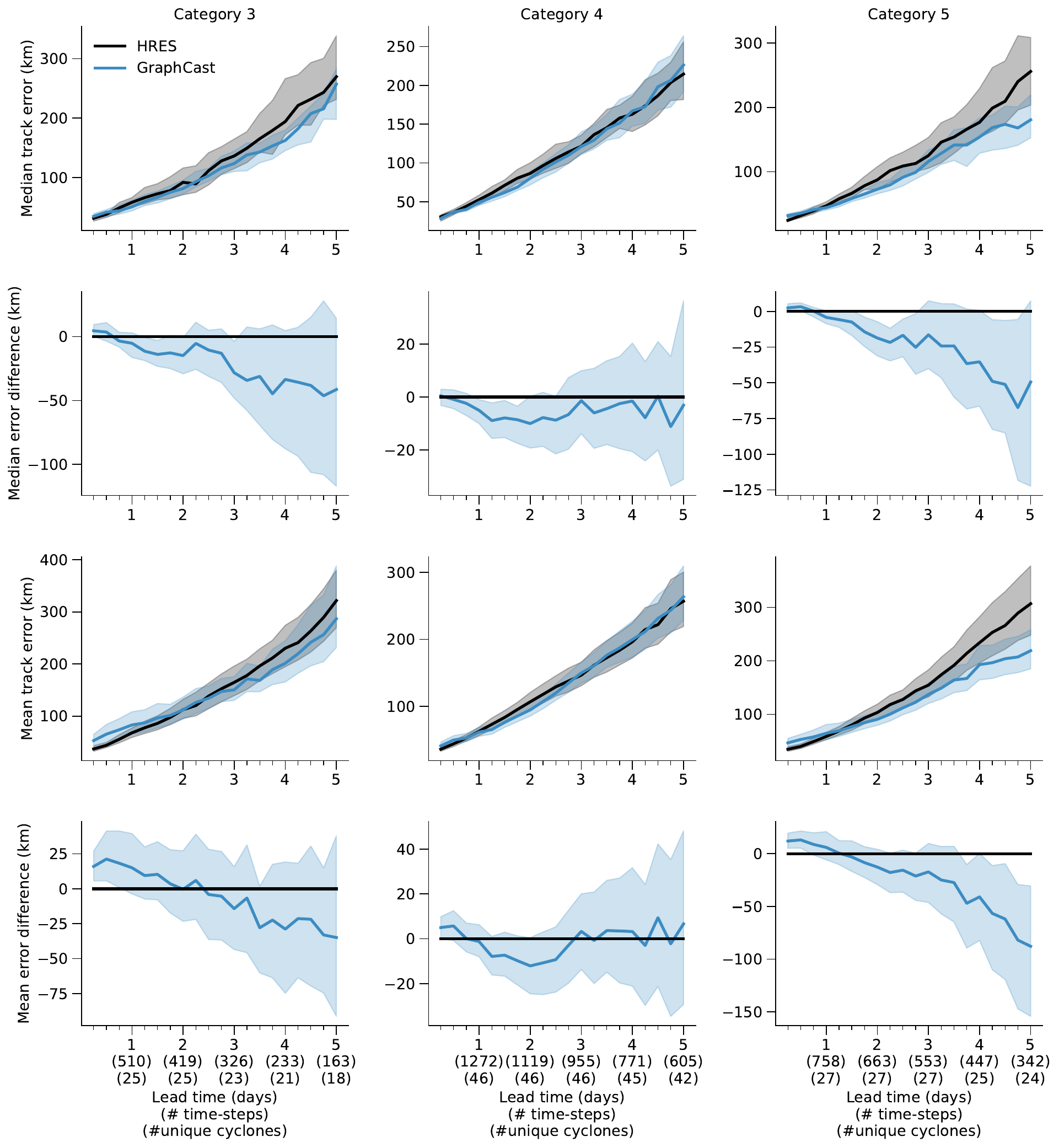}
\caption{\small\textbf{Per-cyclone-category median and mean performance (category 3 to 5)} Each column corresponds to a cyclone category from 3 to 5 on the Saffir-Simpson Hurricane Wind Scale.}
  \label{fig:app:categories_3_4_5}
\end{figure}

Supplements~\cref{fig:app:categories_0_1_2,fig:app:categories_3_4_5} show the median error and paired analysis as a function of lead time, broken down by cyclone category, where category is defined on the Saffir-Simpson Hurricane Wind Scale~\cite{taylor2010saffir}, with category 5 representing the strongest and most damaging storms (note, we use category 0 to represent tropical storms).
We found that \ourmodel has equal or better performance than HRES across all categories. For category 2, and especially for category 5 (the most intense events), \ourmodel is significantly better that HRES, as demonstrated by the per-track paired analysis. We also obtain similar results when measuring mean performance instead of median.

\subsubsection{Tracker details}\label{sec:app:cyclone:tracker}
The tracker we used for \ourmodel is based on our reimplementation of ECMWF's tracker \cite{magnusson2014verification}. Because it is designed for \pointonedegree HRES, we found it helpful to add several modifications to reduce the amount of mistracked cyclones when applied to \ourmodel predictions.
However, tracking errors still occur, which is expected from tracking cyclone from \quarterdegree predictions instead of \pointonedegree.
We note that we do not use our tracker for the HRES baseline, as its tracks are directly recovered from the TIGGE archives~\cite{bougeault2010thorpex}.

We first give a high-level summary of the default tracker from ECMWF, before explaining the modifications we made and our decision process. 

\paragraph{ECMWF tracker}

Given a model's predictions of the variables \varlevel{10u}, \varlevel{10v}, \varlevel{msl} as well as \varlevel{u}, \varlevel{v} and \varlevel{z} at pressure levels \varlevel{200}, \varlevel{500}, \varlevel{700}, \varlevel{850} and \varlevel{1000} \unit{hPa} over multiple time steps,
the ECMWF tracker \cite{magnusson2014verification} sequentially processes each time step to iteratively predict the location of a cyclone over an entire trajectory.
Each 6h prediction of the tracker has two main steps.
In the first step, based on the current location of the cyclone, the tracker computes an estimate of the next location, 6h ahead.
The second step consists in looking in the vicinity of that new estimate for locations that satisfy several conditions that are characteristic of cyclone centers.

To compute the estimate of the next cyclone location, the tracker moves the current estimate using a displacement computed as the average of two vectors:
1) the displacement between the last two track locations (i.e., linear extrapolation) and 2) an estimate of the wind steering, averaging the wind speed \varlevel{u} and \varlevel{v} at the previous track position at pressure levels 200, 500, 700 and 850 \unit{hPa}.

Once the estimate of the next cyclone location is computed, the tracker looks at all local minima of mean sea-level pressure (\varlevel{msl}) within 445 \unit{km} of this estimate. It then searches for the candidate minima closest to the current estimate that satisfies the following three conditions:
\begin{enumerate}
    \item Vorticity check: the maximum vorticity at 850 \unit{hPa} within 278 \unit{km} of the local minima is larger than $5\cdot 10^{-5}$ \unit{s^{-1}}
    for the Northern Hemisphere, or is smaller than $-5 \cdot 10^{−5}$\unit{s^{-1}} for the Southern Hemisphere.
    Vorticity can be derived from horizontal wind (\varlevel{U} and \varlevel{V}).
    \item Wind speed check: if the candidate is on land, the maximum 10\unit{m} wind speed within 278~\unit{km} is larger than 8 \unit{m/s}.
    \item Thickness check: if the cyclone is extratropical, there is a maximum of thickness between 850 \unit{hPa} and 200 \unit{hPa} within a radius of 278 \unit{km}, where the thickness is defined as \varlevel{z}{850}-\varlevel{z}{200}.
\end{enumerate}
If no minima satisfies all those conditions, the tracker considers that there is no cyclone.
ECMWF's tracker allows cyclones to briefly disappear under some corner-case conditions before reappearing. In our experiment with \ourmodel, however, when a cyclone disappear, we stop the tracking.

\paragraph{Our modified tracker}
We analysed the mistracks on cyclones from our validation set years (2016--2017), using a version of \ourmodel trained on 1979--2015, and modified the default re-implementation of the ECMWF tracker as described below. When we conducted a hyper-parameter search over the value of a parameter, we marked in bold the values we selected.
\begin{enumerate}
    \item The current step vicinity radius determines how far away from the estimate a new center candidate can be. We found this parameter to be critical and searched a better value among the following options:   $445\times f$ for f in 0.25, 0.375, \textbf{0.5}, 0.625, 0.75, 1.0 (original value). 
    \item The next step vicinity radius determines how strict multiple checks are. We also found this parameter to be critical and searched a better value among the following options: $278\times f$ for f in 0.25, 0.375, 0.5, 0.625, \textbf{0.75}, 1.0 (original value). 
    \item The next-step estimate of ECMWF uses a 50-50 weighting between linear extrapolation and wind steering vectors. In our case where wind is predicted at \quarterdegree resolution, we found wind steering to sometimes hinder estimates. This is not surprising because the wind is not a spatially smooth field, and the tracker is likely tailored to leverage \pointonedegree resolution predictions. Thus, we hyper-parameter searched the weighting among the following options: 0.0, 0.1, 0.33, \textbf{0.5} (original value).
    \item We noticed multiple misstracks happened when the track sharply reversed course, going against its previous direction. Thus, we only consider candidates that creates an angle between the previous and new direction below $d$ degrees, where $d$ was searched among these values: \textbf{90}, 135, 150, 165, 175, 180 (i.e. no filter, original value). 
    \item We noticed multiple misstracks made large jumps, due to a combination of noisy wind steering and features being hard to discern for weak cyclones. Thus, we explored clipping the estimate from moving beyond $x$ kilometers (by resizing the delta with the last center), searching over the following values for x: $445\times f$ for f in 0.25, 0.5, 1.0, 2.0, 4.0, $\bm{\infty}$ (i.e. no clipping, original value).
\end{enumerate}
During the hyper-parameter search, we also verified on validation data that the tracker applied to \ourmodel dropped a similar number of cyclones as HRES.

\FloatBarrier

\subsection{Atmospheric rivers} \label{sec:app:atmosphericriver}

The vertically integrated water vapor transport (\varlevel{ivt}) is commonly used to characterize the intensity of atmospheric rivers \cite{neiman2008meteorological,moore2012physical}.
Although \ourmodel does not directly predict \varlevel{ivt} and is not specifically trained to predict atmospheric rivers, we can derive this quantity from the predicted atmospheric variables specific humidity, \varlevel{q}, and horizontal wind, (\varlevel{u}, \varlevel{v}), via the relation \cite{neiman2008meteorological}:
\begin{equation}
    IVT = \frac{1}{g} \sqrt{\left(\int_{p_{b}}^{p_{t}} \varlevel{q}(p) \varlevel{u}(p) dp \right)^{2} + \left(\int_{p_{b}}^{p_{t}} \varlevel{q}(p) \varlevel{v}(p) dp \right)^{2}},
\end{equation}
where $g=9.80665 ~\unit{m/s^2}$ is the acceleration due to gravity at the surface of the Earth, $p_{b} = 1000~\unit{hPa}$ is the bottom pressure, and $p_{t} = 300~\unit{hPa}$ is the top pressure.

Evaluation of \varlevel{ivt} using the above relation requires numerical integration and the result therefore depends on the vertical resolution of the prediction.
\ourmodel has a vertical resolution of 37 pressure levels which is higher than the resolution of the available HRES trajectories with only 25 pressure levels. For a consistent and fair comparison of both models, we therefore only use a common subset of pressure levels, which are also included in the WeatherBench benchmark, when evaluating \varlevel{ivt}~\footnote{As suggested by ECMWF during personal conversation.}, namely $[300, 400, 500, 600, 700, 850, 925, 1000]~\unit{hPa}$.

Consistently with the rest of our evaluation protocol, each model is evaluated against its own ``analysis''.
For \ourmodel, we compute the \varlevel{ivt} based on its predictions and we compare it to the \varlevel{ivt} computed analogously from ERA5. Similarly,  we use HRES predictions to compute the \varlevel{ivt} for HRES and and compare it to the \varlevel{ivt} computed from \hresfczero.

\begin{figure}%
  \centering
  \includegraphics[width=\textwidth]{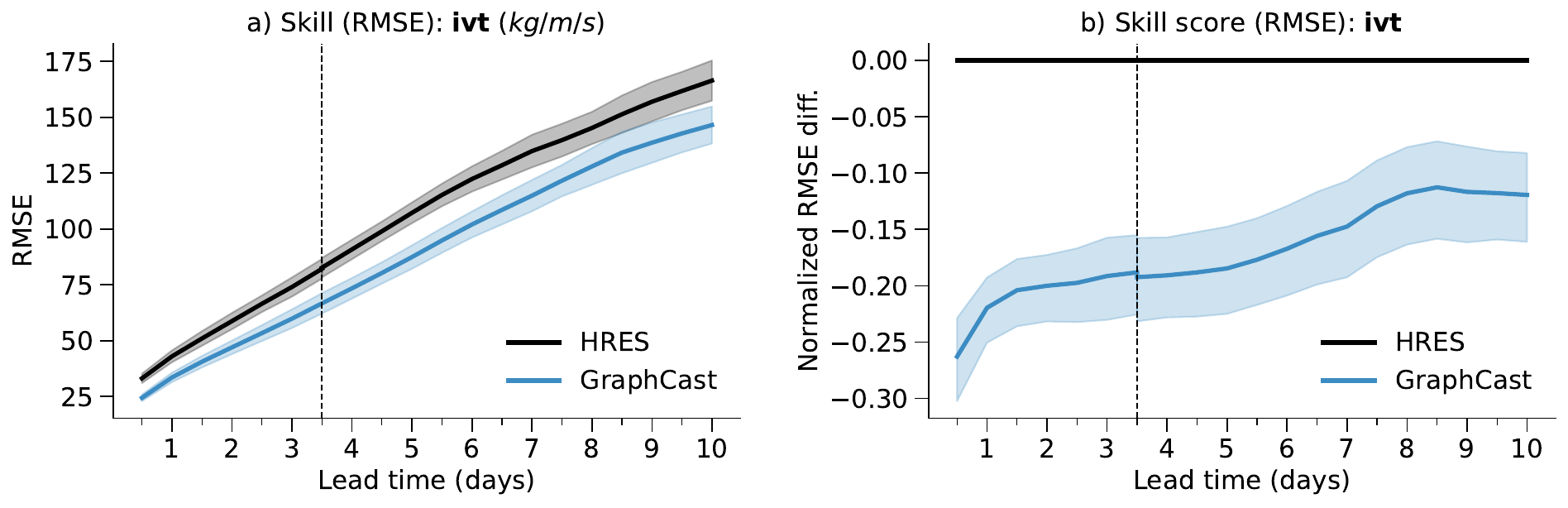}
\caption{\small\textbf{Skill and skill score for \ourmodel and HRES on vertically integrated water vapor transport (ivt) \textit{(lower is better)}.} (a) RMSE skill (y-axis) for \ourmodel (blue line) and HRES (black line) on \varlevel{ivt} as a function of lead time (x-axis), with 95\% confidence interval error bars (see \cref{sec:app:conf_int_rmse}). (b) RMSE skill score (y-axis) for \ourmodel and HRES with respect to HRES on \varlevel{ivt} as a function of lead time (x-axis), with 95\% confidence interval error bars (see \cref{sec:app:conf_int_rmse_skill_score}). \ourmodel{} improves the prediction of \varlevel{ivt} compared to HRES, from 25\% at short lead time, to 10\% at longer horizon.}
  \label{fig:app_atmospheric_river}
\end{figure}

Similarly to previous work \cite{chapman2019improving}, \cref{fig:app_atmospheric_river} reports RMSE skill and skill score averaged over coastal North America and the Eastern Pacific (from \ang{180}W to \ang{110}W longitude, and \ang{10}N to \ang{60}N latitude) during the cold season (Jan-April and Oct-Dec 2018), which corresponds to a region and a period with frequent atmospheric rivers.

\FloatBarrier

\subsection{Extreme heat and cold} \label{sec:app:extremetemperature}

We study extreme heat and cold forecasting as a binary classification problem \cite{magnusson2014verification,lopez2022global} by comparing whether a given forecasting model can correctly predict whether the value for a certain variable will be above (or below) a certain percentile of the distribution of a reference historical climatology (for example above 98\% percentile for extreme heat, and below 2\% percentile for extreme cold). Following previous work  \cite{magnusson2014verification}, the reference climatology is obtained separately for (1) each variable, (2) each month of the year, (3) each time of the day, (4) each latitude/longitude coordinate, and (5) each pressure level (if applicable). This makes the detection of extremes more contrasted by removing the effect of the diurnal and seasonal cycles in each spatial location.
To keep the comparison as fair as possible between HRES and \ourmodel{}, we compute this climatology from HRES-fc0 and ERA5 respectively, for years 2016-2021. We experimented with other ways to compute climatology (2016-2017 as well as using ERA5 climatology 1993-2016 for both models), and found that results hold generally.

Because extreme prediction is by definition an imbalanced classification problem, we base our analysis on precision-recall plots which are well-suited for this case \cite{saito2015precision}. The precision-recall curve is obtained by varying a free parameter ``gain'' consisting of a scaling factor with respect to the median value of the climatology, i.e. $\text{scaled forecast} = \text{gain} \times (\text{forecast} - \text{median climatology}) + \text{median climatology}$.  This has the effect of shifting the decision boundary and allows to study different trade offs between false negatives and false positives. Intuitively, a 0 gain will produce zero forecast positives (e.g. zero false positives), and an infinite gain will produce amplify every value above the median to be a positive (so potentially up to 50\% false positive rate). The ``gain'' is varied smoothly from 0.8 to 4.5. Similar to the rest of the results in the paper we also use labels from HRES-fc0 and ERA5 when evaluating  HRES and \ourmodel{}, respectively.

\begin{figure}[ht]
  \centering
  \vspace{-1.5cm}
  \includegraphics[width=0.95\textwidth]{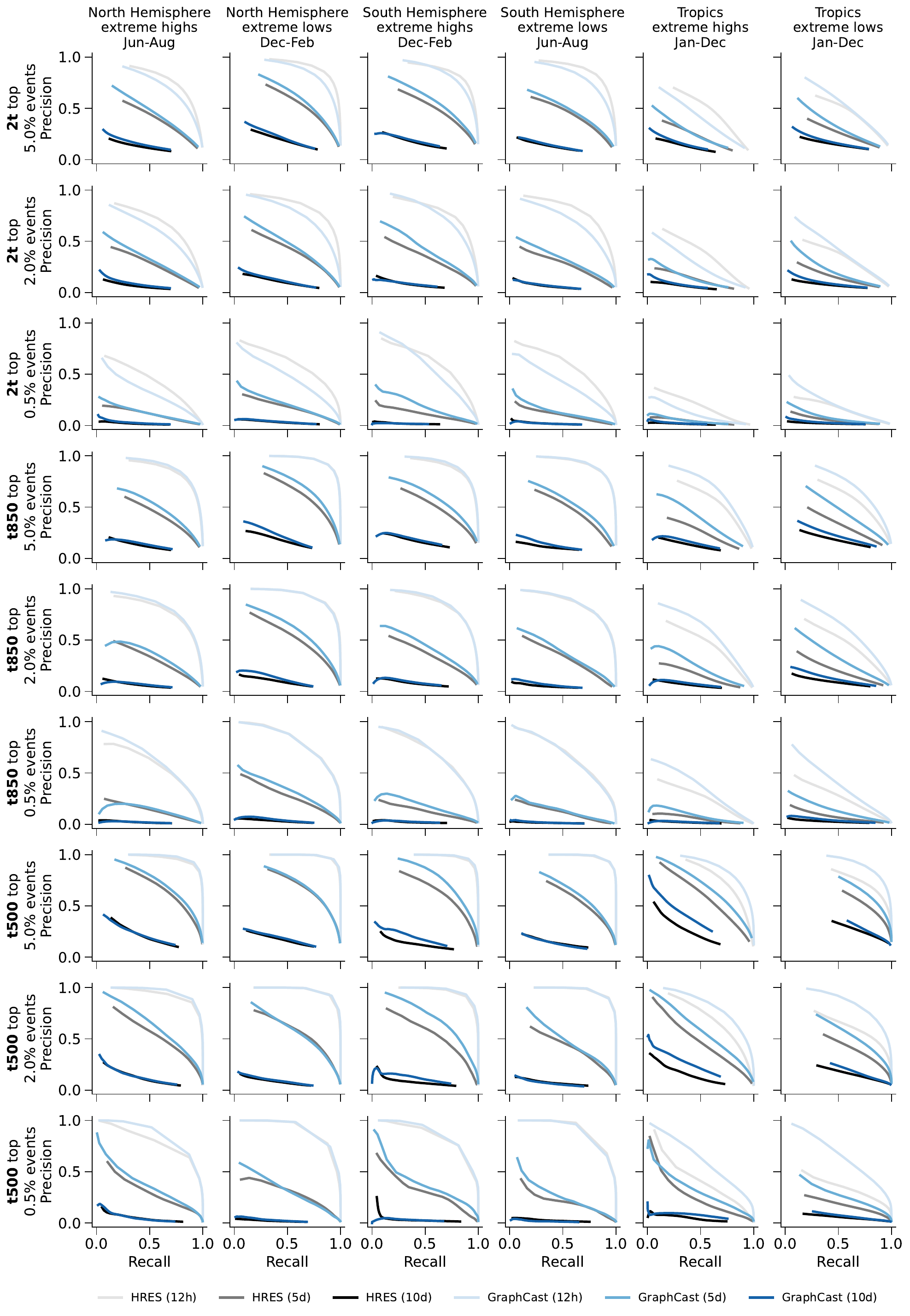}
  \caption{\small\textbf{Detailed extremes evaluation.} Higher precision and recall is better.
  }
  \label{fig:app:extremes}
\end{figure}

\begin{figure}[ht]
  \centering
  \vspace{-1.5cm}
  \includegraphics[width=0.95\textwidth]{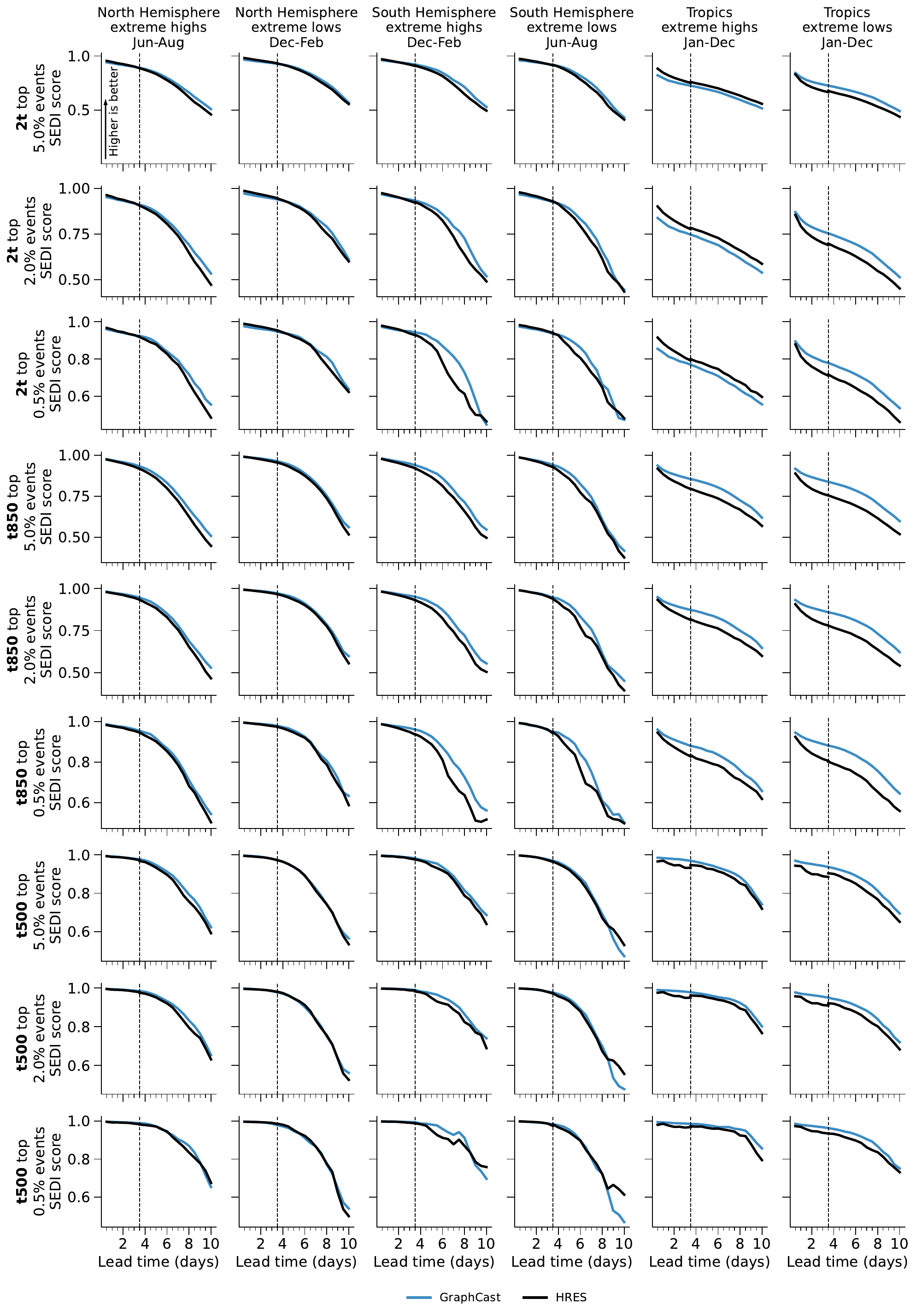}
  \caption{\small\textbf{Extremes SEDI scores.} Maximum SEDI scores across the extreme prediction precision-recall curves (\cref{fig:app:extremes}) as function of lead time.
  }
  \label{fig:app:extremes_sedi}
\end{figure}

We focus our analysis on variables that are relevant for extreme temperature conditions, specifically \varlevel{2t} \cite{magnusson2014verification, lopez2022global}, and also \varlevel{t850}, \varlevel{z500} which are often used by ECMWF to characterize heatwaves \cite{ecmwf2022heatwaveuk}. Following previous work\cite{lopez2022global}, for extreme heat we average across June, July, and August over land in the northern hemisphere (latitude \textgreater\, 20$^{\circ}$) and across December, January, and February over land in the southern hemisphere (latitude \textless\, -20$^{\circ}$). For extreme cold, we swapped the months for the northern and southern hemispheres. See full results in \cref{fig:app:extremes}. We also provide a more fine-grained lead-time comparison, by summarizing the precision-recall curves by selecting the point with the highest SEDI score\cite{magnusson2014verification} and showing this as function of lead time (\cref{fig:app:extremes_sedi}).

\FloatBarrier

\newpage
\section{Forecast visualizations}

\begin{figure}%
  \centering
  \includegraphics[width=0.88\textwidth]{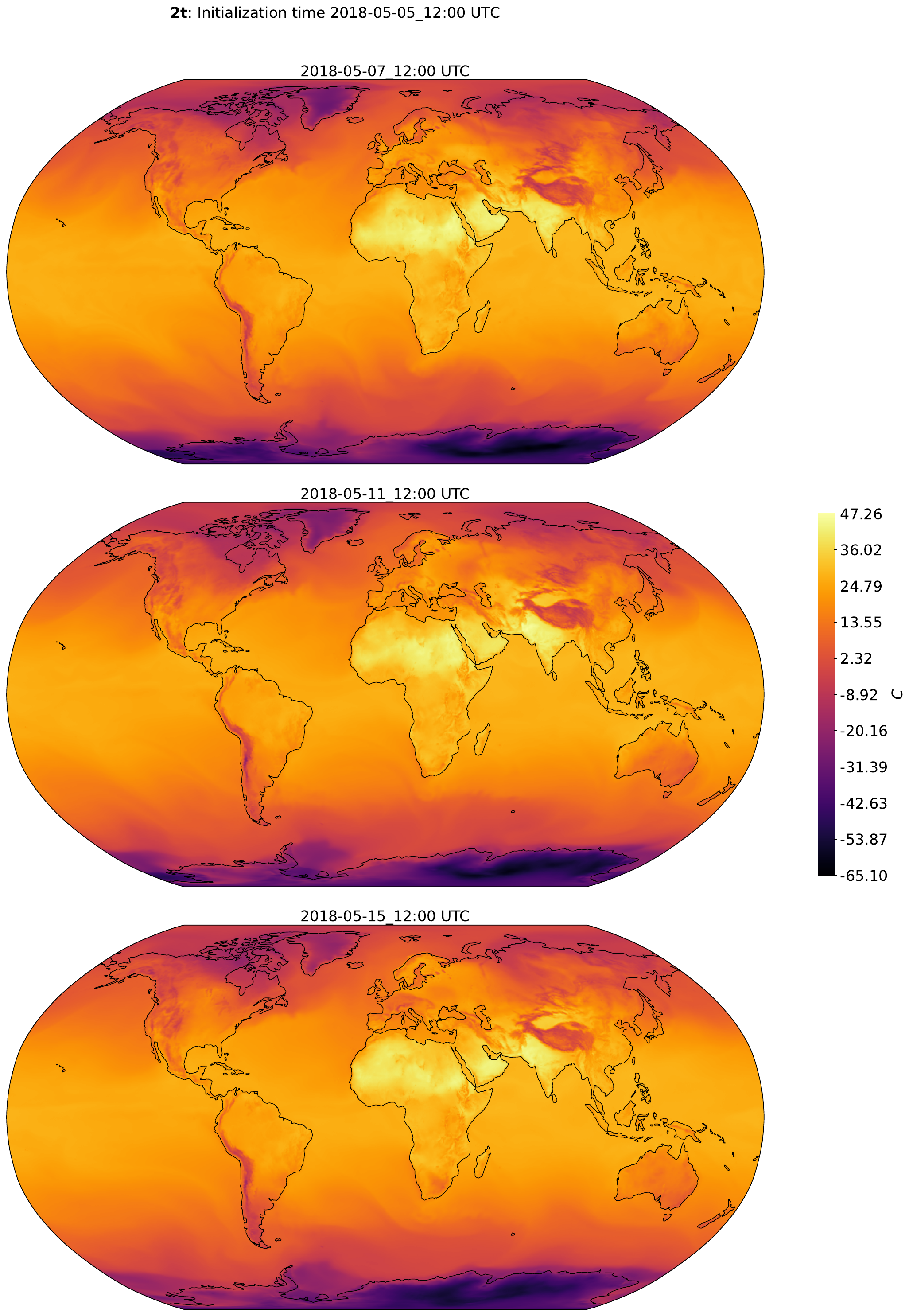}
\caption{\small\textbf{Forecast visualization: \varlevel{2t}.} Forecast initialized at 2018-05-05 12:00 UTC, with plots corresponding to 2, 6, and 10 day lead times.}
  \label{fig:vis1}
\end{figure}

\begin{figure}%
  \centering
  \includegraphics[width=0.88\textwidth]{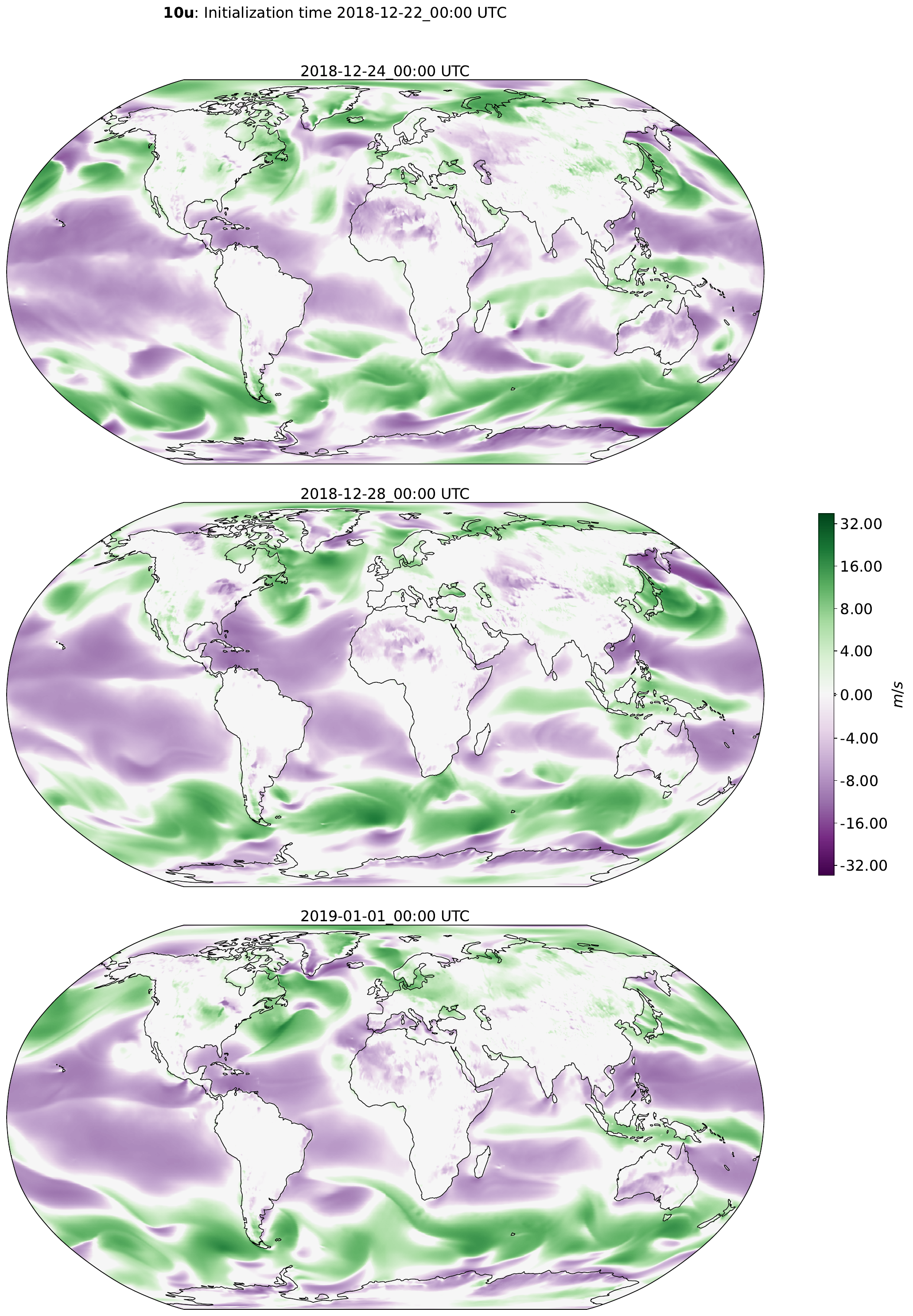}
\caption{\small\textbf{Forecast visualization: \varlevel{10u}.} Forecast initialized at 2018-12-22 00:00 UTC, with plots corresponding to 2, 6, and 10 day lead times.}
  \label{fig:vis2}
\end{figure}

\begin{figure}%
  \centering
  \includegraphics[width=0.88\textwidth]{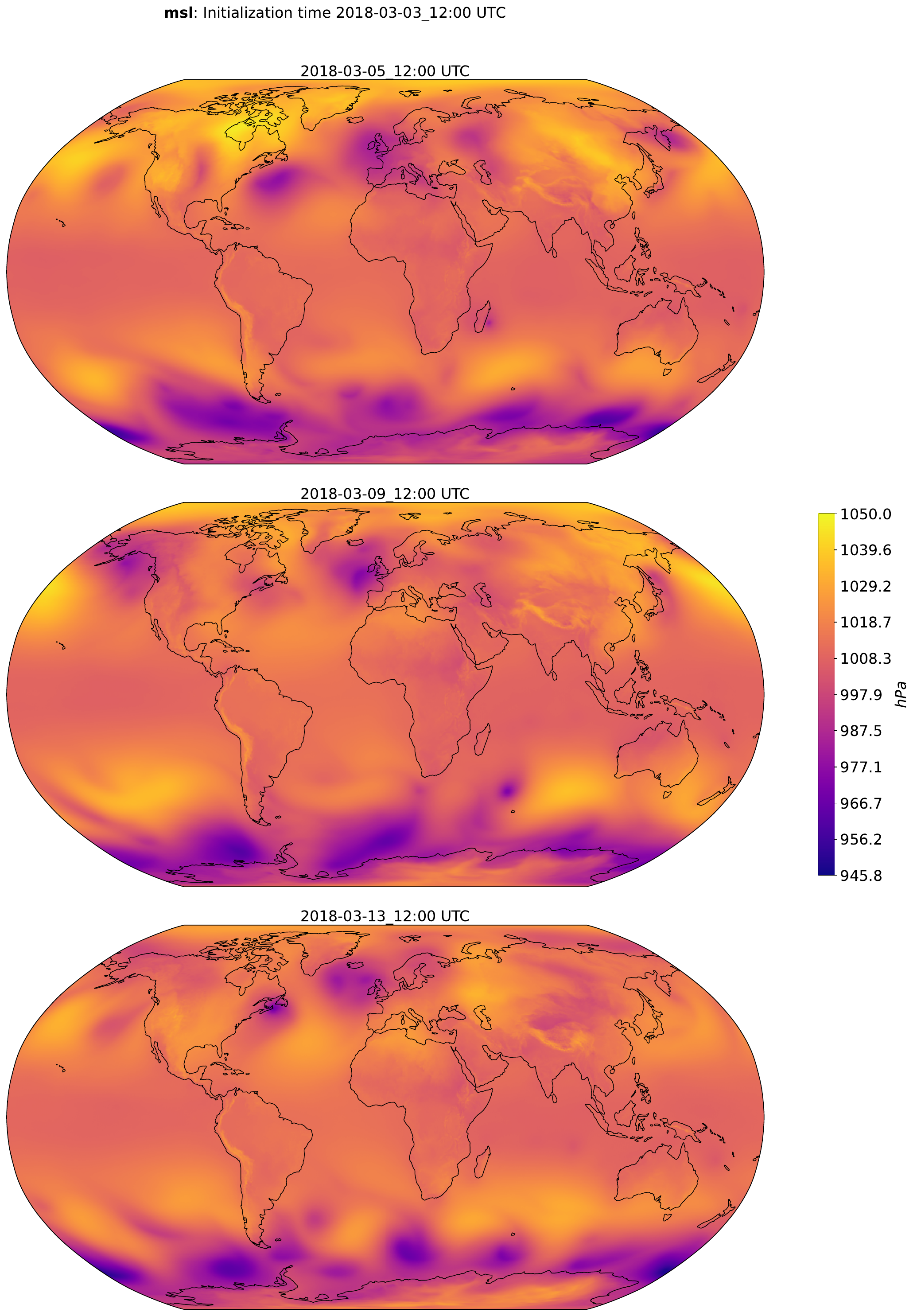}
\caption{\small\textbf{Forecast visualization: \varlevel{msl}.} Forecast initialized at 2018-03-03 12:00 UTC, with plots corresponding to 2, 6, and 10 day lead times.}
  \label{fig:vis3}
\end{figure}

\begin{figure}%
  \centering
  \includegraphics[width=0.88\textwidth]{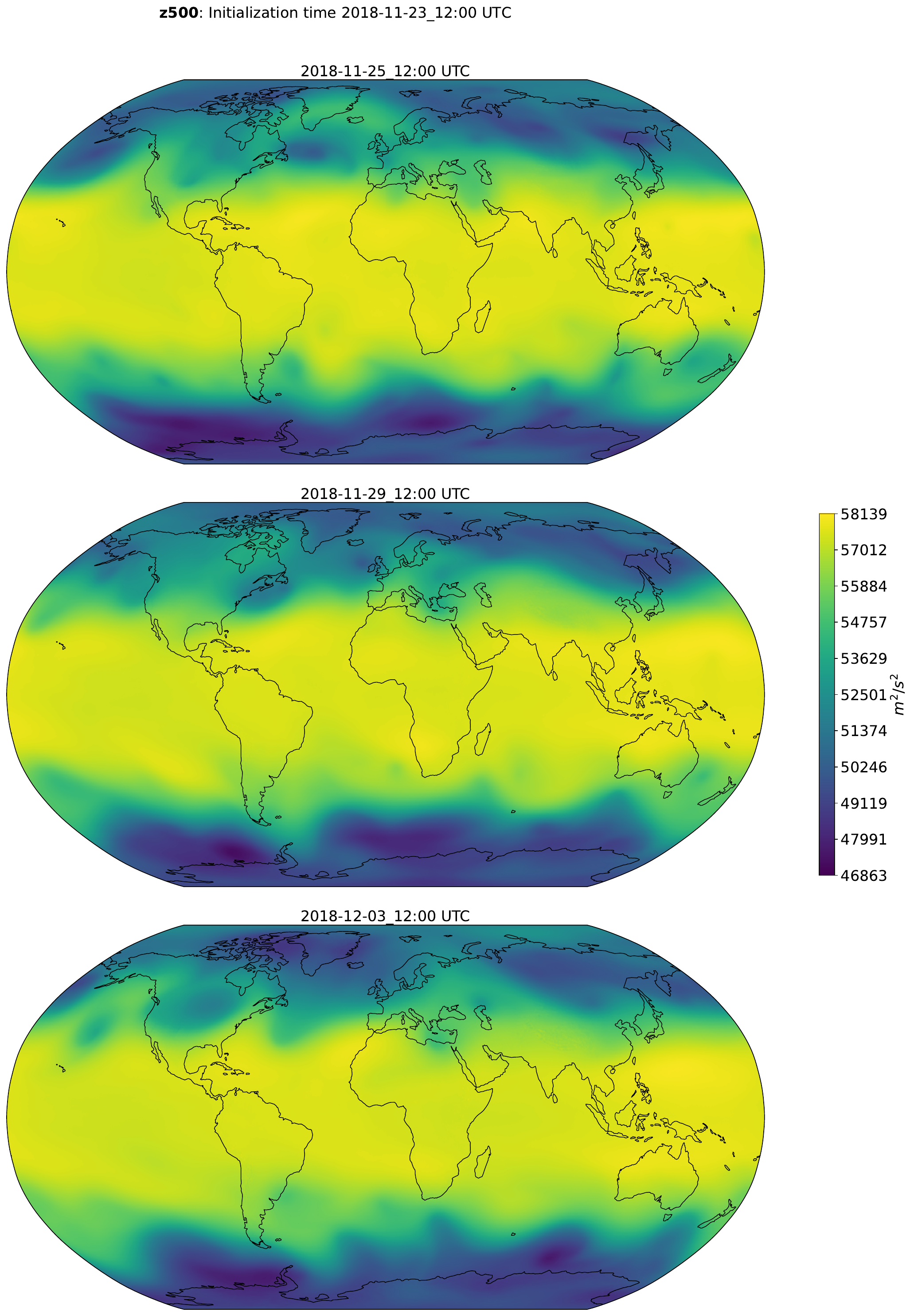}
\caption{\small\textbf{Forecast visualization: \varlevel{z}{500}.} Forecast initialized at 2018-11-23 12:00 UTC, with plots corresponding to 2, 6, and 10 day lead times.}
  \label{fig:vis4}
\end{figure}

\begin{figure}%
  \centering
  \includegraphics[width=0.88\textwidth]{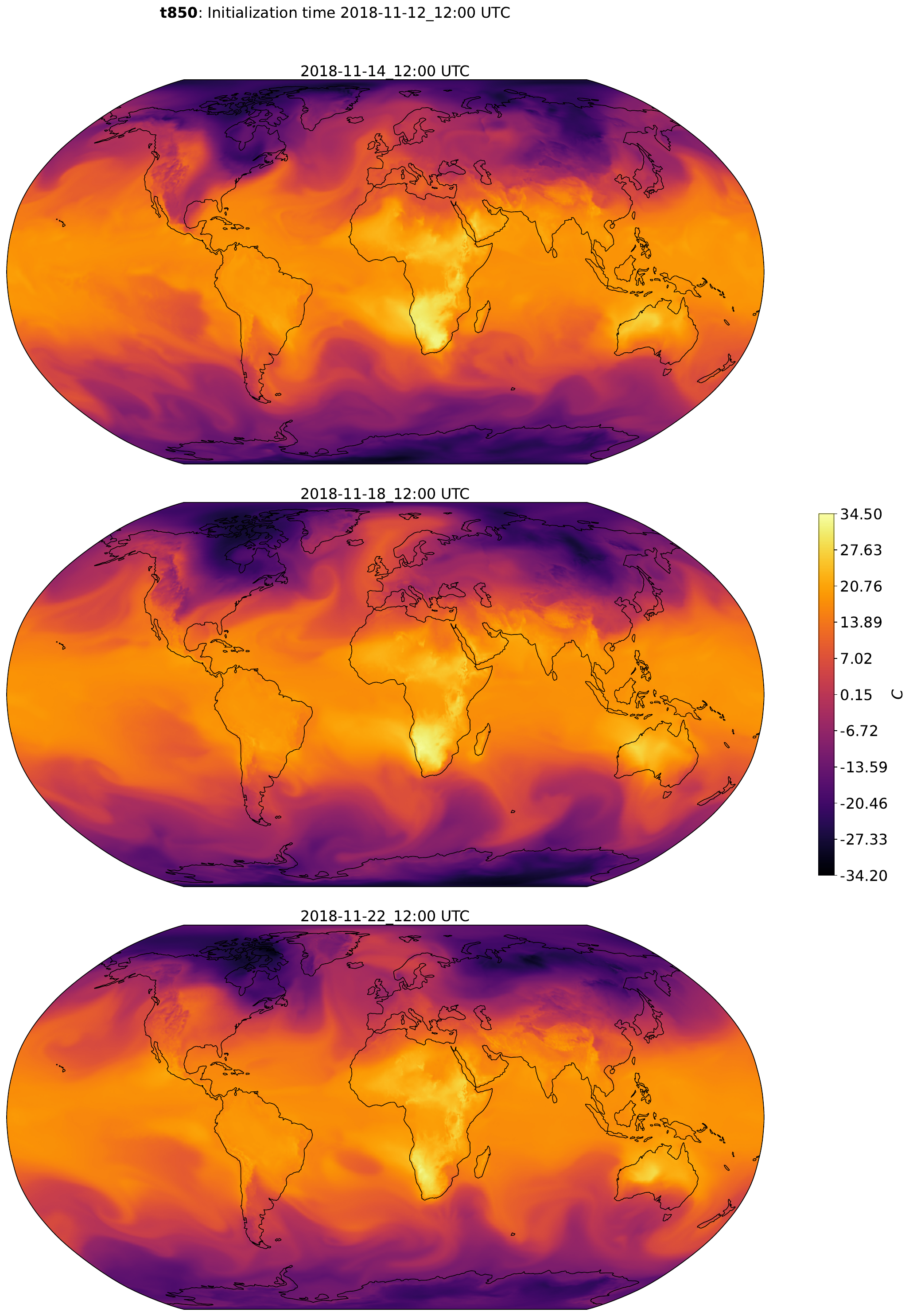}
\caption{\small\textbf{Forecast visualization: \varlevel{t}{850}.} Forecast initialized at 2018-11-12 12:00 UTC, with plots corresponding to 2, 6, and 10 day lead times.}
  \label{fig:vis5}
\end{figure}

\begin{figure}%
  \centering
  \includegraphics[width=0.88\textwidth]{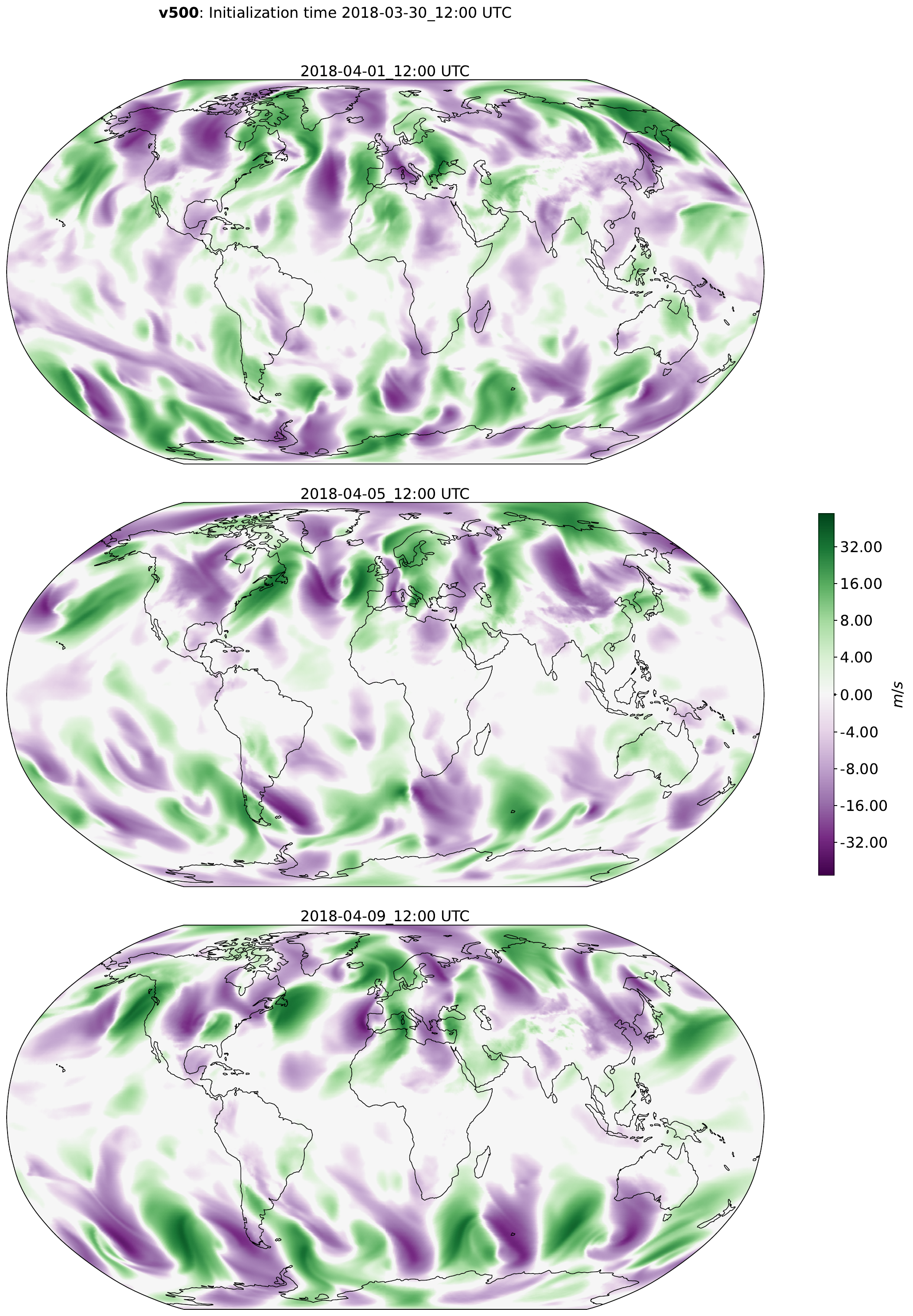}
\caption{\small\textbf{Forecast visualization: \varlevel{v}{500}.} Forecast initialized at 2018-03-30 12:00 UTC, with plots corresponding to 2, 6, and 10 day lead times.}
  \label{fig:vis6}
\end{figure}

\begin{figure}%
  \centering
  \includegraphics[width=0.88\textwidth]{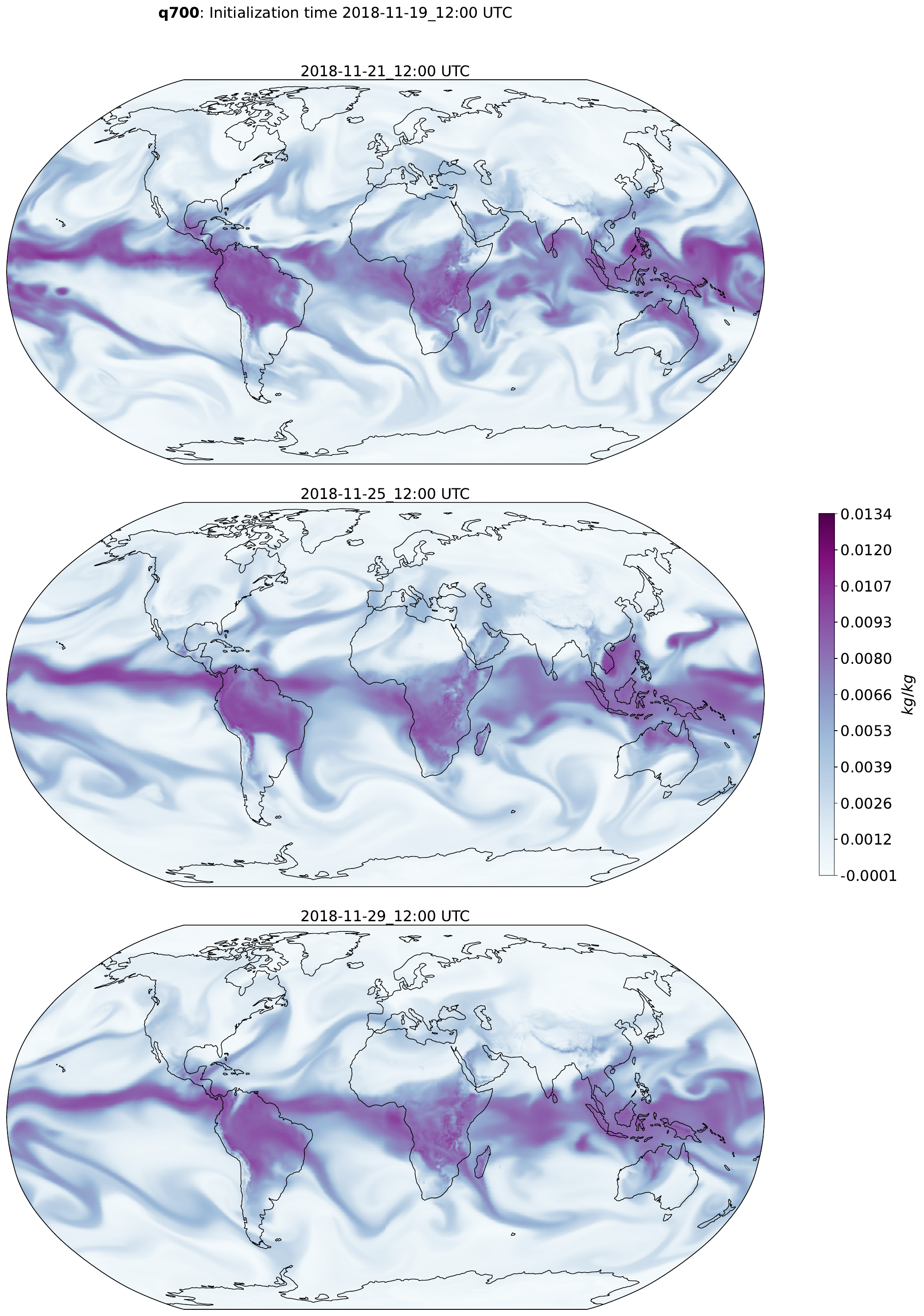}
\caption{\small\textbf{Forecast visualization: \varlevel{q}{700}.} Forecast initialized at 2018-11-19 12:00 UTC, with plots corresponding to 2, 6, and 10 day lead times.}
  \label{fig:vis7}
\end{figure}

In this final section, we provide a few visualization examples of the predictions made by \ourmodel for variables \varlevel{2t} (\cref{fig:vis1}), \varlevel{10u} (\cref{fig:vis2}), \varlevel{msl} (\cref{fig:vis3}), \varlevel{z}{500} (\cref{fig:vis4}), \varlevel{t}{850} (\cref{fig:vis5}), \varlevel{v}{500} (\cref{fig:vis6}), \varlevel{q}{700} (\cref{fig:vis7}). For each variable, we show a representative prediction from \ourmodel by choosing the example with the median performance on 2018.


\FloatBarrier

\putbib[references]
\end{bibunit}

\end{document}